\documentclass[10pt,journal,compsoc,peerreview]{IEEEtran}
\usepackage{color}

\newcommand{\vect}[1]{\boldsymbol{#1}}
\newcommand{\tbf}[1]{\textbf{#1}}
\usepackage{bbm}
\usepackage{bbold}
\usepackage{mathtools}

\ifCLASSOPTIONcompsoc
  \usepackage[nocompress]{cite}
\else
  \usepackage{cite}
\fi
\ifCLASSINFOpdf
  \usepackage[pdftex]{graphicx}
  \usepackage{tikz}
  \graphicspath{{./images/}{./images/stereo_dataset/}{./images/stereo_dataset/quarter_res_undistort/}}
  \DeclareGraphicsExtensions{.pdf,.jpg,.jpeg,.png}
\else
\fi
\usepackage[export]{adjustbox}  
\usepackage{rotating}

\usepackage{amsmath}
\usepackage{color}

\usepackage{algorithmic}

\usepackage[ruled,norelsize]{algorithm2e}
\makeatletter
\newcommand{\removelatexerror}{\let\@latex@error\@gobble}
\DeclareRobustCommand\onedot{\futurelet\@let@token\@onedot}
\def\@onedot{\ifx\@let@token.\else.\null\fi\xspace}

\def\eg{e.g\onedot} 
\def\ie{i.e\onedot} 
\def\insitu{\textit{in situ }}
\def\etal{\emph{et al}\onedot}

\makeatother
\usepackage{placeins}  
\usepackage{float}

\usepackage{stfloats}
\usepackage{url}

\usepackage{cleveref}

\usepackage{multirow}

\begin{document}
%
\title{Underwater Single Image Color Restoration Using Haze-Lines and a New Quantitative Dataset}
%
%
%
%

\author{Dana~Berman,
        Deborah~Levy,
        Shai~Avidan,
        and~Tali~Treibitz
\IEEEcompsocitemizethanks{\IEEEcompsocthanksitem D. Berman and S. Avidan are with the School of
of Electrical Engineering, Tel Aviv University, Israel.\protect\\
E-mail: danamena@post.tau.ac.il
\IEEEcompsocthanksitem D. Levy and T. Treibitz are with the School of Marine Sciences, University of Haifa, Israel.}
}

%
%

\markboth{IEEE TRANSACTIONS ON PATTERN ANALYSIS AND MACHINE INTELLIGENCE}
{Shell \MakeLowercase{\textit{et al.}}: Bare Demo of IEEEtran.cls for Computer Society Journals}
%


\IEEEtitleabstractindextext{%
\begin{abstract}
Underwater images suffer from color distortion and low contrast, because light is attenuated while it propagates through water. Attenuation under water varies with wavelength, unlike terrestrial images where attenuation is assumed to be spectrally uniform. The attenuation depends both on the water body and the 3D structure of the scene, making color restoration difficult.

Unlike existing single underwater image enhancement techniques, our method takes into account multiple spectral profiles of different water types.
By estimating just two additional global parameters: the attenuation ratios of the blue-red and blue-green color channels, the problem is reduced to single image dehazing, where all color channels have the same attenuation coefficients.
Since the water type is unknown, we evaluate different parameters out of an existing library of water types. Each type leads to a different restored image and the best result is automatically chosen based on color distribution.

We collected a dataset of images taken in different locations with varying water properties, showing color charts in the scenes. Moreover, to obtain ground truth, the 3D structure of the scene was calculated based on stereo imaging. This dataset enables a quantitative evaluation of restoration algorithms on natural images and shows the advantage of our method.

\end{abstract}

}

\maketitle

\IEEEdisplaynontitleabstractindextext

%
\IEEEpeerreviewmaketitle

\IEEEraisesectionheading{\section{Introduction}\label{sec:introduction}}
\IEEEPARstart{U}{nderwater} images often lack contrast and depict inaccurate colors due to the scattering and absorption of light as it propagates through the water. Yet color and contrast are extremely important for visual surveys in the ocean. For example, enhanced images can improve automatic segmentation, increase the accuracy of feature matching between images taken from multiple viewpoints, and aid in navigation.

The attenuation of light depends both on the light's wavelength and the distance it travels~\cite{mobley1994light}.
The wavelength-dependent attenuation causes color distortions that increase with an object's distance. In addition, the scattering induces a distance-dependent additive component on the scene which reduces contrast. These phenomena cannot be globally corrected since the color degradation depends on the distance of the object from the camera.
Moreover, the attenuation parameters are affected by seasonal, geographic, and climate variations. These variations were categorized into different optical water types by Jerlov \cite{jerlov1976marine}.

\sloppy Unfortunately, existing single underwater image enhancement techniques under-perform as they do not take into account the diverse spectral properties of water. In addition, their evaluation is generally based on a handful of images, and is mostly qualitative. A few methods have been evaluated using no-reference image quality metrics, which operate only on the luminance and cannot measure the color correction.

In this paper we suggest a method to recover the distance maps and object colors in scenes photographed under water and under ambient illumination, using just a single image as input. Our recovery takes into account the different optical water types and is based on a more comprehensive physical image formation model than the one previously used.
Restoration from a single image is desirable because water properties temporally change, sometimes within minutes. In addition, no additional equipment, such as a tripod or filters, is required.

A variety of methods have been developed for the closely related single image dehazing problem, in which images are degraded by weather conditions such as haze or fog.
Under the assumption of wavelength-independent attenuation, single image dehazing is an ill-posed problem with three measurements per pixel (the R,G,B values of the input image) and four unknowns (the R,G,B values of the object and its distance from the camera). The transmission is the fraction of the scene's radiance that reaches the camera, and is related to the distance via the attenuation coefficient.

\begin{figure*}[t]
\centering
     \includegraphics[width=0.98\linewidth]{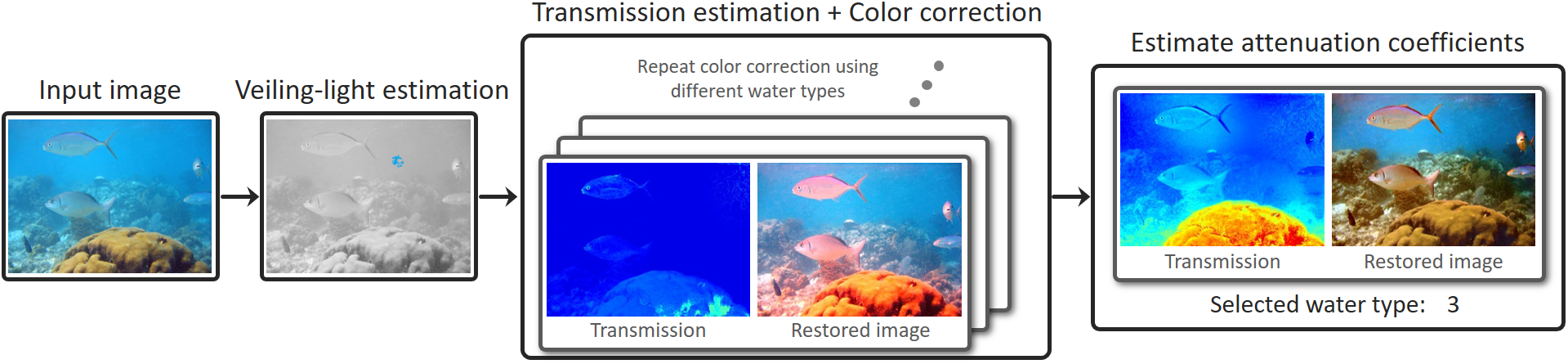}\vspace{-0.3cm}
    \caption[Overview of the proposed color restoration algortihm]{The proposed color restoration and transmission estimation method: First, the veiling-light (the ambient light that is scattered into the line of sight) is estimated. The pixels whose color is the veiling light are shown in color. Then, the transmission estimation and color restoration are repeated for multiple water types that have different optical characteristics, as described in Section~\ref{Jerlov}. Finally, the best result is selected automatically based on the gray-world assumption. Photo from~\cite{carlevaris2010initial}.}\vspace{-0.2cm}
    \label{fig:Pipeline}
\end{figure*}

Under water, where the assumption of wavelength-independent attenuation does not hold, there are theoretically three unknown transmission values per pixel, one per channel, yielding six unknowns with only three measurements. However, the color-dependent transmission is related to the distance via the attenuation coefficients. Based on this relation we reduce the problem to estimation of four unknowns per pixel as before, with two new \emph{global} parameters, the ratios between the attenuation coefficients of the color channels.
We estimate these parameters using an existing library of water types, and based on the color distribution of the image after correction. We utilize the fact that using a wrong water type leads to distorted colors. Our results demonstrate a successful single image restoration of underwater scenes using a comprehensive physical image formation model. Thus, we are able to recover more complex 3D scenes than previous methods and, in addition, estimate the optical water type. Fig.~\ref{fig:Pipeline} depicts the proposed method.

Since public ground truth data is not available, we took multiple stereo images that contain color charts. We used the stereo images to recover the true distance from the camera. We then conducted a thorough quantitative analysis, comparing the results of multiple algorithms to the ground truth data. Our algorithm is competitive with other state-of-the-art methods.

\section{Related Work}\label{sec:related}

The participating medium causes color shifts and reduces contrast in images. The degradation depends both on the water properties and the 3D structure, which can be estimated by analyzing \emph{multiple} images that satisfy certain conditions. E.g., Schechner and  Karpel~\cite{ClearUnderwaterVisionCVPR2004,schechner2005recovery} take two images with orthogonal polarizer angles and utilize the partial polarization of light to restore the visibility.
Multiple images of the same object taken from different \emph{known} viewpoints~\cite{bryson2012colour,bryson2015true,yamashita2007color} are used to estimate attenuation coefficients and recover the scene.
These methods have limited applicability since they impose constraints on the imaging conditions.

Scattering media limits visibility in terrestrial images as well, when imaging in fog, haze, or turbulence~\cite{TurbulenceInduced}.
\emph{Single} image dehazing methods, \eg~\cite{HazeLines,DarkChannelCVPR2009,Tan2008}, take as input a single hazy image and estimate the unknown distance map and the clear image of the scene. Different priors are used in order to solve the ill-posed problem, under the assumption of color-independent transmission, which is reasonable for terrestrial images yet is violated under water. The dark channel prior (DCP)~\cite{DarkChannelCVPR2009} assumes that within small image patches, at least one pixel has a low value in some color channel, and uses the minimal value to estimate the distance.
The haze-lines prior~\cite{HazeLines} is based on the observation that the number of colors in natural images is small, and that similar colors appear all over the image plane.

Some variations of DCP were proposed for the underwater domain~\cite{carlevaris2010initial,chiang2012underwater,drews2013transmission,Galdran2015RedChannel,lu2015contrast}. Perez \etal~\cite{BenchmarkDehazingAUVs} recently performed an evaluation of DCP variations for underwater autonomous vehicles.
Carlevaris-Bianco et al.~\cite{carlevaris2010initial} assume a color-independent transmission, and propose a variant of DCP based on the difference between the maximum of the red channel and the maximum of the blue and green channels in each patch. They claim this value is inversely related to transmission, since red is attenuated at a higher rate than blue or green.
Drews et al.~\cite{drews2013transmission} apply DCP to the blue and green channels only, since the red channel is often attenuated rapidly and cannot be used to estimate the transmission. They get improved transmission maps, but still assume a uniform transmission across channels for recovery.
Chiang and Chen~\cite{chiang2012underwater} recover the transmission using standard DCP. They assume the recovered transmission is the transmission of the red channel, as under water this channel has the lowest transmission. They use fixed attenuation coefficients measured for open ocean waters to recover the image based on the estimated red transmission.
Lu et al.~\cite{lu2015contrast} estimate the transmission using the darker of the blue and red channels.
They use the same fixed water attenuation as in~\cite{chiang2012underwater} to recover the scene. 
Galdran \etal~\cite{Galdran2015RedChannel} suggest the Red-Saturation prior, incorporating information from the inverse of the red channel and the saturation in order to estimate the transmission. They use a spectrally uniform transmission since the water type is not easy to determine. Instead, they add an additive veiling-light term to the color restoration. 

Emberton \etal~\cite{emberton2015hierarchical} detect and segment regions of pure veiling-light and deliberately set their transmission value to be high, while using the estimation of Drews et al.~\cite{drews2013transmission} for the rest of the scene. This is done to avoid enhancing artifacts in the veiling-light regions. To handle the spectral dependency of the attenuation, they classify the water type to blue, turquoise, or green-dominated, and apply white-balance to the image before estimating the spectrally-uniform transmission. Since this step is global, it cannot compensate for the distance-dependent attenuation. In addition, the classification to three different water types is too coarse compared to common optical classification schemes.

Despite the abundance of DCP based methods, the underlying assumption does not hold in many underwater scenarios: bright foreground sand has high values in all color channels and might be mistaken to have a low transmission despite being close to the camera. Moreover, the background water has a dominant color (hence at least one color channel is low), and many of the mentioned methods inaccurately estimate the transmission there to be high.

Peng et al.~\cite{blurrinessICIP2015} estimate the scene distance via image blurriness, which grows with the distance due to scattering. They disregard the spectral dependence of the transmission.
Peng and Cosman~\cite{blurriness2017} combine the blurriness prior with~\cite{carlevaris2010initial} and assume open ocean waters. While this prior is physically valid, it has limited efficiency in textureless areas.

The above mentioned restoration methods aim for a physics-based recovery of a scene's colors, while estimating its 3D structure in the process. Other methods aim for a visually pleasing result, \eg~\cite{ancutiICPR2016, ancuti2012enhancing, ancutiTIP2018, ancutiICIP2017, GuidedTrigonometricBilateral}, but have not shown color consistency that is required for scientific measurements.

Convolutional networks have been recently used for dehazing \cite{AOD_Net_ICCV17, MSCNN_ECCV2016}, and also for underwater restoration~\cite{shin2016estimation}. However, their training relies on purely synthetic data and thus highly depends on the quality of the simulation models, which wrongly neglected the spectral dependency of the attenuation, leading to poor results.
We believe that the severe lack of data and the challenge in conducting \insitu experiments are withholding progress in this direction. Two recent attempts to cope with this problem include synthesizing data and acquiring images in test tanks.
Blasinski \etal~\cite{blasinski2017uiss} released tools to realistically simulate the appearance of underwater scenes in a variety of water conditions. Unfortunately, the simulation accuracy decreases with increasing depth and chlorophyll concentration. To the best of our knowledge, it hasn't been utilized for learning-based underwater image enhancement methods yet.
Duarte \etal~\cite{duarte2016dataset} published a dataset for evaluating underwater image restoration methods. However, these images were acquired in a test tank using milk, and do not represent real-world conditions.
In addition to the theoretical contributions, a major goal of this research is to create and publish an extensive dataset of underwater images with
ground-truth of the 3D scene structure.

The wavelength dependent absorption properties of water are used in~\cite{ShapeFromWater} to recover depth. However this method requires an active illumination system and is valid only for short distances, since near-infrared light is rapidly attenuated in water. Known geometry is used in~\cite{sheinin2016next} to plan the imaging viewpoints under water that result in high contrast.

Morimoto et al.~\cite{SpiderModel2010} estimate the optical properties of layered surfaces based on non-linear curves in RGB space, which is a similar physical model. Nonetheless, user scribbles are required in order to detect the curves.

\section{Background} \label{sec:UW_background}

\subsection{Image Formation Model}

We follow the model developed in~\cite{schechner2005recovery}. In each color channel $c \in \{R,G,B\}$, the image intensity at each pixel is composed of two components, attenuated signal and veiling-light:
\begin{equation} \label{eq:BasicModelUW}
I_c(\vect{x}) = t_c(\vect{x})  J_c(\vect{x}) + (1-t_c(\vect{x})) \cdot A_c \;\;,
\end{equation}
where bold denotes vectors, $\vect{x}$ is the pixel coordinate, $I_c$ is the acquired image value in color channel $c$, $t_c$ is the transmission of that color channel, and $J_c$ is the object radiance that we wish to restore. The global veiling-light component $A_c$ is the scene value in areas with no objects ($t_c=0,\, \forall c \in \{R,G,B\}$). Eq.~(\ref{eq:BasicModelUW}) applies to linear captured data, prior to in-camera processing such as color-space conversion, tone mapping, and compression. Therefore, $\vect{I}$ refers to the image obtained from the raw file after minimal processing such as demosaicing and black current subtraction \cite{akkaynak2014use,RawGuide}.

The transmission depends on object's distance $z(\vect{x})$ and the water attenuation coefficient for each channel $\beta_c$:
\begin{equation} \label{eq:transmissionUW}
t_c(\vect{x}) = \exp(-\beta_c z(\vect{x})) \;\;.
\end{equation}
In the ocean, the attenuation of red colors can be an order of magnitude larger than the attenuation of blue and green~\cite{mobley1994light}. Therefore, as opposed to the common assumption in single image dehazing, the transmission $\vect{t} = (t_R,t_G,t_B)$ is wavelength-dependent.

\begin{figure*}[tb]
\centering
\includegraphics[width=0.98\linewidth]{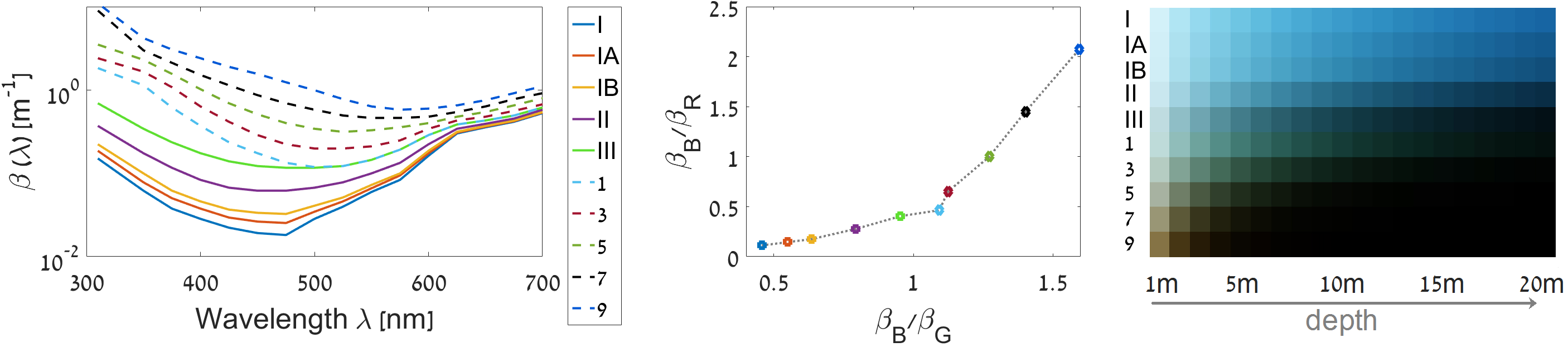} \vspace{-0.1cm}
\caption[Jerlov's classification of water types]{Left:~Approximate attenuation coefficients ($\beta$) of Jerlov water types. Data taken from ~\cite{mobley1994light}, based on measurements in~\cite{austin1986spectral}.
Solid lines mark open ocean water types while dashed lines mark coastal water types.
Middle:~$\beta$ ratios of $R,G,B$ channels for each water type, based the wavelengths: 475nm, 525nm, and 600nm, respectively.
Right:~simulation of the appearance of a perfect white surface viewed at depth of 1-20m in different water types (reproduced from \cite{DeryaCVPR17}).}
\label{fig:jerlov}
\end{figure*}

\vspace{-0.2cm}
\subsection{Water Attenuation}\vspace{-0.1cm}\label{Jerlov}

The attenuation of light under water is not constant and varies with geographical, seasonal, and climate related events. In clear open waters, visible light is absorbed at the longest wavelengths first, appearing deep-blue to the eye. In near-shore waters, sea water contains more suspended particles than the central ocean waters, which scatter light and make coastal waters appear less clear than open waters. In addition, the absorption of the shortest wavelengths is stronger, thus the green wavelength reaches deeper than other wavelengths.

Jerlov~\cite{jerlov1976marine} developed a frequently used classification scheme for oceanic waters, based on water clarity. The Jerlov water types are I, IA, IB, II and III for open ocean waters, and 1 through 9 for coastal waters. Type I is the clearest and type III is the most turbid open ocean water. Likewise, for coastal waters, type 1 is clearest and type 9 is most turbid. Fig.~\ref{fig:jerlov}(Left) depicts the attenuation coefficients' dependence on wavelength, while Fig.~\ref{fig:jerlov}(Right) shows an RGB simulation of the appearance of a perfect white surface viewed at different depths in different water types. The common notion that red colors are attenuated faster than blue/green only holds for oceanic water types.

When capturing an image using a commercial camera, three color channels $R,G,B$ are obtained. Thus, we are interested in three attenuation coefficients: $(\beta_R,\beta_G,\beta_B)$ in order to  correct the image. We use the Jerlov water types to constrain the space of attenuation coefficients in the RGB domain. We show in Sec.~\ref{sec:UnderwtaerTransEstimation} that the three attenuation coefficient themselves are not required for transmission estimation, but rather their ratios (two variables). Fig.~\ref{fig:jerlov}(Middle) shows the ratios of the attenuation coefficients: $\beta_B /\beta_R$ vs. $\beta_B/\beta_G$ of Jerlov water types for wavelengths of peak camera sensitivity according to \cite{jiang2013space} ($475{\rm nm}, 525{\rm nm}$, and $600{\rm nm}$ for $B,G,R$, respectively). Essentially, we approximate the cameras' spectral response as a Dirac delta function, similarly to~\cite{SpiderModel2010,blurriness2017}. This approximation is camera-agnostic, which is advantageous since the true response functions are rarely published.

\vspace{-0.2cm}
\section{Color Restoration} \label{sec:UW_algo}
We show in the following that if we were given the two \emph{global} attenuation ratios, then the problem can be  reduced to single image dehazing (i.e., single attenuation coefficient across all channels) for which good algorithms exist. The question now becomes how to estimate these two parameters? Our solution is to evaluate every possible water type, as defined by Jerlov, and pick the best one. Each water type defines a known, and fixed, pair of attenuation ratios that we can use. Once we have evaluated all possible water types (10 in total) we use a standard natural image statistics prior to pick the best result. In particular, we found the Gray-World assumption prior to work well for our needs. The method is illustrated in Fig.~\ref{fig:Pipeline} and summarized in Alg.~\ref{alg:HLUW}.

\vspace{-0.2cm}
\subsection{Veiling-Light Estimation}\label{sec:VeilingLightEstimation}

First we describe how we estimate the veiling-light, which is required for de-scattering.
We assume an area without objects is visible in the image, in which the pixels' color is determined by the veiling-light alone. Such an area should be smooth and not have texture, which is an important feature for veiling light estimation, e.g.~\cite{emberton2015hierarchical, blurriness2017}.
This assumption often holds when the line of sight is horizontal. It does not hold when photographing a reef wall up close, or when the camera is pointed downwards. However, in these cases, the distance of objects from the camera usually varies less then in horizontal scenes, and a simple contrast stretch is likely to be sufficient.

In order to detect the pixels that belong to the veiling-light, we apply a linear contrast stretching and then generate an edge map of the scene using the Structured Edge Detection Toolbox~\cite{DollarICCV13edges} with pre-trained model and default parameters. We then threshold the edge map and look for the largest connected component. The pixels belonging to the largest component are classified as veiling-light pixels ($\vect{x} \in VL$). The veiling-light $\vect{A}$ is the average color of those pixels. This is demonstrated on the veiling-light estimation step of Fig.~\ref{fig:Pipeline}, where only the pixels $\vect{x} \in VL$ are shown in color.


\subsection{Transmission Estimation}\label{sec:UnderwtaerTransEstimation}

Combining and rearranging Eqs.~(\ref{eq:BasicModelUW},\ref{eq:transmissionUW}) yields for the blue channel:
\begin{equation} \label{eq:BasicBlue} 
A_B - I_B = e^{-{\beta_B}Z} \cdot \left( A_B - J_B\right)\;\;,
\end{equation}
and similarly for the red channel:
\begin{equation} \label{eq:BasicRed}
A_R - I_R = e^{-{\beta_R}Z} \cdot \left( A_R - J_R\right)\;\;.
\end{equation}

Raising Eq.~(\ref{eq:BasicRed}) to the power of $\frac{\beta_B}{\beta_R}$ yields:
\begin{equation} \label{eq:HazeLine1}
\left( A_R \!-\! I_R \right) ^{\frac{\beta_B}{\beta_R}} = e^{-{\beta_R}z \cdot {\frac{\beta_B}{\beta_R}}}\! \left( A_R \!-\! J_R\right)^{\frac{\beta_B}{\beta_R}} = t_B \! \left( A_R \!-\! J_R\right)^{\frac{\beta_B}{\beta_R}}\:.
\end{equation} 

Denote the ratios between the attenuation coefficients:
\begin{equation}\label{eq:def_beta_ratio}
\beta_{BR}=\beta_B/\beta_R~~,~~\beta_{BG}=\beta_B/\beta_G\;\;.
\end{equation}

Then, in this medium-compensated space we achieve a form similar to Eq.~(\ref{eq:BasicModelUW}), with one unknown transmission per-pixel, common to all color channels: \vspace{-0.05cm}
\begin{equation} \label{eq:CommonSpace}
\left[\begin{array}{l}
\!\!(I_R(\vect{x}) - A_R)^{\beta_{BR}\!}\!\!\\
\!\!(I_G(\vect{x}) - A_G)^{\beta_{BG}\!}\!\!\\
\!\!(I_B(\vect{x}) - A_B)
\end{array}\right]
 = t_B(\vect{x})\!
 \left[\begin{array}{l}
\!\!(J_R(\vect{x}) - A_R)^{\beta_{BR}\!}\!\\
\!\!(J_G(\vect{x}) - A_G)^{\beta_{BG}\!}\!\\
\!\!(J_B(\vect{x}) - A_B)
\end{array}\right] .
\end{equation}

This form is similar to the Haze-Lines~\cite{HazeLines} formulation. Therefore, we similarly cluster the pixels to Haze-Lines and obtain an initial estimation of the transmission of the blue channel $\tilde{t}_B$.
Since the value ($I_c -A_c$) might be negative, we avoid numerical issues when raising to the power $\beta_{Bc}$ by raising the absolute value and keeping the sign.

In~\cite{HazeLines} it was assumed that there is a haze-free pixel in each Haze-Line. However, the attenuation coefficients measures by Jerlov indicate that even scene points that are located at a distance of only one meter from the camera have a $t_B$ of about $0.9$, depending on water type. Thus, we multiply the initial transmission estimation by $0.9$ (similarly to~\cite{blurrinessICIP2015}).


A bound on the transmission arises from the fact that $J_c \ge 0, \forall c \in \{R,G,B\}$. We substitute this bound in Eq.~(\ref{eq:BasicModelUW}) and obtain a lower bound $t^{LB}$ on the transmission of the blue channel, $t_B$, taking into account the different attenuation coefficients of the different color channels: \vspace{-0.1cm}
\begin{equation} \label{eq:LowerBoundUW}
t^{LB} := max \left\lbrace 1 - \frac{I_B}{A_B},  \left(1 - \frac{I_G}{A_G} \right)^{\beta_{BG}}\!\!\!, \left(1 - \frac{I_R}{A_R} \right)^{\beta_{BR}} \! \right\rbrace. \vspace{-0.1cm}
\end{equation}

We detect pixels $\vect{x}$ with no scene objects based on their Mahalanobis distance from the distribution of the veiling-light pixels: $D_M \left( \vect{I} (\vect{x}) \right)$ (from here on, when referring to the Mahalanobis distance, it is with respect to the distribution of intensities of veiling-light pixels). We set the transmission of such pixels to be the lower bound $t^{LB}$ calculated in Eq.~(\ref{eq:LowerBoundUW}). However, a binary classification of the pixels to $VL, \overline{VL}$ often results in abrupt discontinuities in the transmission map, which are not necessarily distance discontinuities. Therefore, we use a soft-matting and calculate the transmission as follows:
\vspace{-0.1cm}\begin{equation} \label{eq:transBlend1}
t_B (\vect{x})\! =\!\! \begin{cases}
\!t^{L\!B}\!(\!\vect{x}\!)\,  &\!\! D\!_M\!\! \left( \vect{I}\! (\!\vect{x}\!) \right)\! \le\! \overline{D\!_M}\!\! +\!\! \sigma\!_M\\
\!\tilde{t}\!_B(\!\vect{x}\!)\,  &\!\! D\!_M\!\! \left( \vect{I}\! (\!\vect{x}\!) \right)\! \ge\! D\!_M^{max}\! +\!\!  \sigma\!_M\\
\!\alpha (\!\vect{x}\!)\! \cdot\! t^{L\!B}\!(\!\vect{x}\!)\!  +\! \left(1\!\!-\!\!\alpha (\!\vect{x}\!)\right)\! \cdot\! \tilde{t}(\!\vect{x}\!)  &\! otherwise\\
\end{cases} 
\end{equation}
\sloppy where $\;\overline{D_M}\;=\; \frac{1}{\left\vert VL \right\vert} \;\cdot\; \sum_{\vect{x} \in VL}{D_M \left(\; \vect{I} (\vect{x}) \;\right)}\;$ is the average \\ Mahalanobis distance of the veiling-light pixels,
\mbox{$D_M^{max} = \max_{\vect{x} \in VL} \left\{  D_M \left( \vect{I} (\vect{x}) \right) \right\}$} is their maximal Mahalanobis distance and $\sigma_M$ is the standard deviation. $\alpha (\vect{x})$ is the matting coefficient for pixels that cannot be classified to object/ water with high probability: \mbox{$\alpha (\vect{x}) = \frac{D_M \left( \vect{I} (\vect{x}) \right) - \overline{D_M} - \sigma_M}{D_M^{max} - \overline{D_M} }$}, yielding a relatively steep transition between $VL$ and $\overline{VL}$.

Finally, we regularize the transmission using Guided Image Filter~\cite{GuidedFilterJournal}, with a contrast-enhanced input image as guidance.

\vspace{-0.2cm}
\subsection{Scene Recovery}\label{sec:SceneRecovery}

Once $t_B$ is estimated, we can compensate for the color attenuation using the following: \vspace{-0.1cm}
\begin{equation} \label{eq:CalcJ} \vspace{-0.1cm}
J_c = A_c + \frac{I_c-A_c}{e^{-\beta_cZ}} = A_c + \frac{I_c-A_c}{t_B^{\beta_c / \beta_B}} , \;\; \forall c \in \{R,G,B\}.\;
\vspace{-0.05cm}\end{equation}

Eq.~(\ref{eq:CalcJ}) compensates for the intensity changes that happen in the path between the object and the camera.
In addition, the ambient illumination is attenuated by the water column from the surface to the imaging depth, resulting in a colored global illumination. We are interested in restoring the colors as if they were viewed under white light, without a color cast. Since this effect is global in the scene, we correct it by performing white balance on the result. This global operator works well only because the distance-dependent attenuation has already been compensated for.

Finally, since Eq.~(\ref{eq:BasicModelUW}) applies to the linear captured data, we convert the linear image to sRGB using a standard image processing pipeline, including color-space conversion and gamma curve for tone mapping~\cite{RawGuide}.

\begin{figure}[tb]
 \removelatexerror
\begin{algorithm}[H]
\caption{Underwater image restoration \label{alg:HLUW}}
\renewcommand{\algorithmicrequire}{\textbf{Input:}}
\renewcommand{\algorithmicensure}{\textbf{Output:}}
\begin{algorithmic}[1]
\REQUIRE $\vect{I}(\vect{x})$
\ENSURE $\vect{\hat{J}}(\vect{x}), \hat{\vect{t}}(\vect{x})$
\STATE Detect veiling light pixels using structured edges \\ (Sec. \ref{sec:VeilingLightEstimation}) and calculate $\vect{A}$
\FOR{each $(\beta_{BR},\beta_{BG})$ values of water types (Fig.~\ref{fig:jerlov} middle)}
\FOR{each $c \in \{R,G,B\}$ }
\STATE $\tilde{I}_c(\vect{x}) =  {\rm sign}\left( I_c(\vect{x}) - A_c \right) \cdot  {\rm abs} \left( I_c(\vect{x}) - A_c  \right)^{\beta_{Bc}}$  ($\beta_{Bc}$ is defined in Eq.~\ref{eq:def_beta_ratio})
\ENDFOR
\STATE Cluster pixels to 500 Haze-Lines as in~\cite{HazeLines} and estimate an initial transmission $\tilde{t}_B$
\STATE Apply soft matting to $\tilde{t}_B$ with lower bound (Eq.~\ref{eq:transBlend1})
\STATE Regularization using guided image filter, with a contrast-enhanced input as guidance
\STATE Calculate the restored image using Eq.~\ref{eq:CalcJ}
\STATE Perform a global WB on the restored image
\ENDFOR 
\STATE Return the image that best adheres to the \\ Gray-World assumption on pixels  $\vect{x} \notin VL$
\end{algorithmic}
\end{algorithm}\vspace{-0.5cm}
\setlength{\textfloatsep}{9pt}
\end{figure}

\vspace{-0.2cm}
\subsection{Estimation of Attenuation Coefficient Ratios}\label{sec:EstimateAttenuationRatio}

Using the wrong coefficients ($\beta_{BR},\beta_{BG}$) leads to restorations with skewed colors and wrong transmission maps.
We use this insight to determine the most appropriate water type. We perform the restoration multiple times using different ratios of attenuation coefficients, corresponding to different Jerlov water types, and choose the best result automatically based on the Gray-World assumption. The attenuation coefficients' ratios are shown in Fig.~\ref{fig:jerlov}(middle).

According to the Gray-World assumption~\cite{GreyWorldAssumption}, the average reflectance of surfaces in the world is achromatic. It has been used in the past for estimating attenuation coefficients underwater using known distances~\cite{bryson2012colour}. However, a significant portion of images taken under water often contain water without any objects. The Gray-World assumption obviously does not hold there. Therefore, we apply the Gray-World assumption only at image regions that contain objects, i.e., those that were not identified as veiling-light pixels. Thus, among all results using different water types, we choose the image with the smallest difference between the average values of the red, green and blue channels.

We considered several other measures (e.g. maximal contrast~\cite{Tan2008}), but found that the simple Gray-World assumption gave the best results and therefore we focus on this measure.

\vspace{-0.2cm}
\section{Experiments}\label{sec:UW_experiments}

\subsection{Experimental Set-up and Method}\label{sec:UW_quantitative}

We wish to quantitatively evaluate the proposed method and compare it to other methods. Since it is impossible to obtain medium-free \textit{in situ} images, comparisons in the existing literature are either based on photos of a color card taken from a short range, e.g.~\cite{chiang2012underwater, lu2015contrast, ancutiTIP2018}, or based on non-reference image quality metrics~\cite{panetta2016human, UCIQE}.
Photos of color cards, even if they are taken at different depths, e.g.~\cite{chiang2012underwater}, are usually taken from a short range and do not contain objects at different distances from the camera as is often the case in natural scenes. Nevertheless, restoring colors of objects that are at a single distance from the camera is a much easier problem than the challenging color reconstruction required for a complex 3D scene.
Similarly, images taken in water tanks~\cite{lu2015contrast} or swimming pools~\cite{ancutiTIP2018} depict objects close to the camera, and exhibit different absorption and scattering properties than natural bodies of water.
Alternatively, several no-reference image quality metrics have been developed for comparing natural underwater scenes. UCIQE~\cite{UCIQE} metric is designed to quantify the nonuniform color cast and low-contrast that characterize underwater images, while UIQM~\cite{panetta2016human} addresses three underwater image quality criterions: colorfulness, sharpness and contrast. These metrics are somewhat heuristic and have limited applicability.

\begin{figure}[tb]
\centering
\frame{\includegraphics[width=0.98\linewidth]{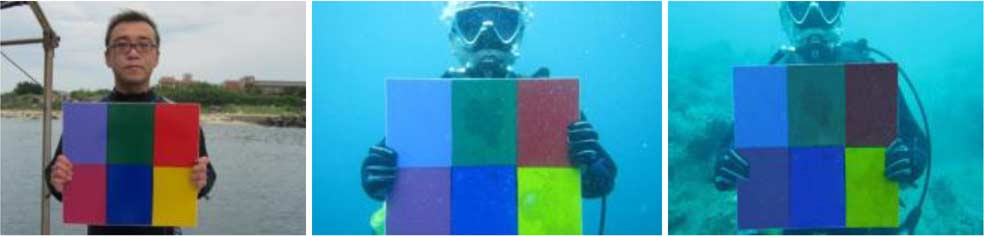}} \\
(a) Source: \cite{chiang2012underwater}\vspace{0.05cm} \\
\addtolength{\tabcolsep}{-4pt}
\begin{tabular}{ccc}
\includegraphics[height=2.05cm]{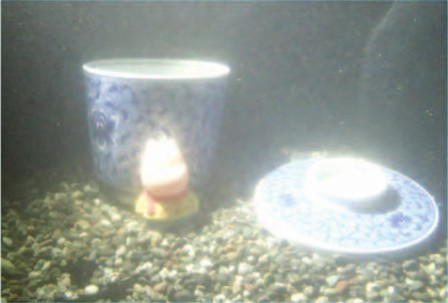} &
\includegraphics[height=2.05cm]{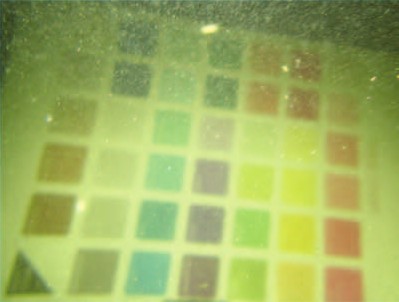} &
\includegraphics[height=2.05cm]{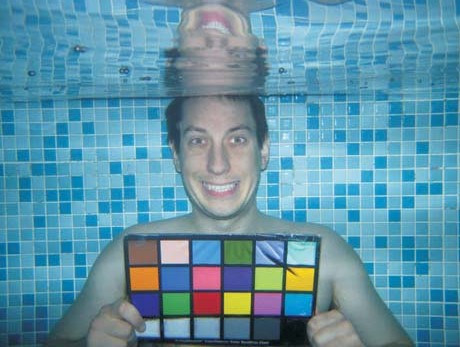} \\
(b) Source: \cite{lu2015contrast} & (c) Source: \cite{lu2015contrast} & (d) Source: \cite{ancutiTIP2018}
\end{tabular}\vspace{-0.1cm}
\caption[Quantitative evaluation examples from the literature]{\textbf{Quantitative evaluations in the literature}. (a) Diver holding a board with six colored patches, from left to right: before
diving, diving at a depth of $5 \text{m}$, and diving at a depth of $15 \text{m}$. (b-c) Water tank experiments: due to the tank's size ($90 \text{cm} \times 45 \text{cm} \times
45 \text{cm}$), the objects are 30 cm deep and the distance between the objects and the camera is approximately $60 \text{cm}$. Deep sea soil is added to the seawater to increase scattering, at a concentration of $20\text{mg/liter}$ (b) and $100\text{mg/liter}$ (c). (d) Example of an underwater taken in a swimming pool~\cite{ancutiTIP2018}, where similar photos were taken with different cameras to evaluate white balancing techniques.\vspace{-0.05cm}}
\label{fig:ColorCardLiterature}
\end{figure}

\begin{figure}[tb]
\centering \addtolength{\tabcolsep}{-4pt}
\begin{tabular}{cc}
\includegraphics[width=0.48\linewidth]{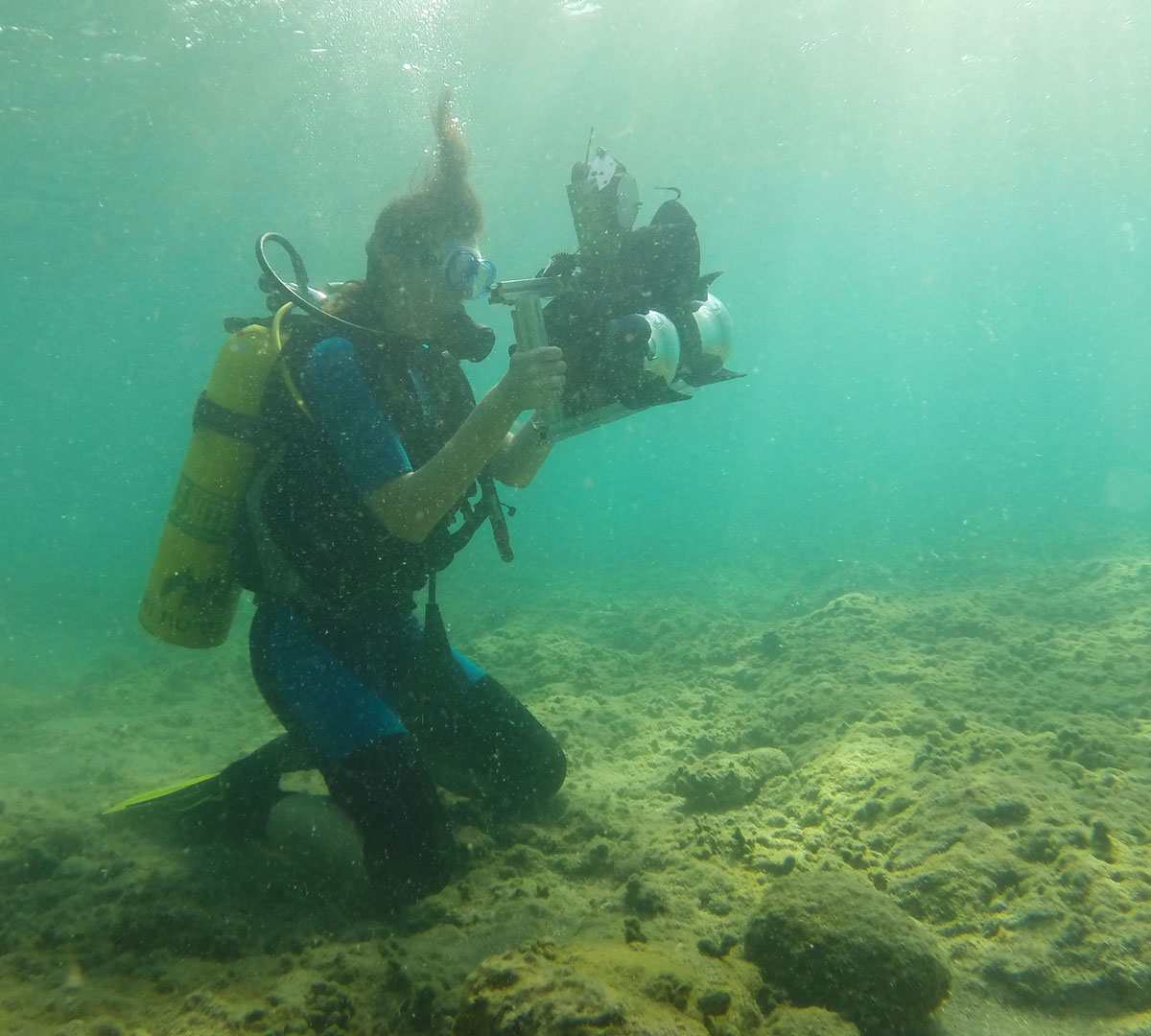} &
\includegraphics[width=0.48\linewidth]{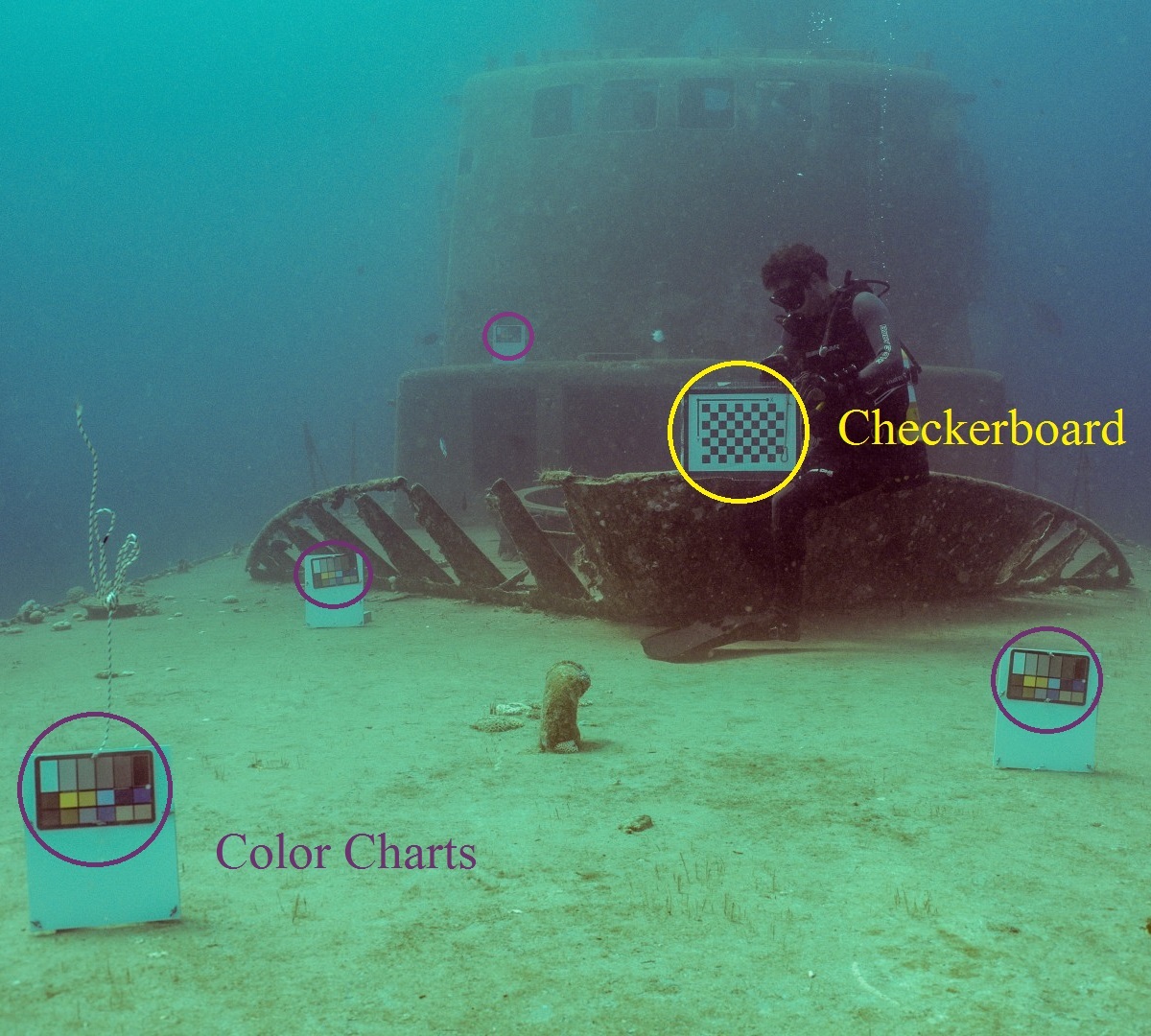}
\end{tabular}\vspace{-0.1cm}
\caption[Creating the underwater dataset]{\textbf{Creating the dataset.} The stereo rig is shown on the left, and the color charts and calibration checkerboard are visible on the right.}
\label{fig:DataGathering} \vspace{-0.3cm}
\end{figure}

We, on the other hand, gathered \insitu data at different seasons, depths, and water types (both in tropical waters and in murkier coastal waters). All of the scenes were illuminated by natural light only.
The quantitative evaluation is two-fold, and relies both on color and distances.
First, we placed identical waterproof color charts (from DGK Color Tools) at different distances from the camera (circled in purple in Fig.~\ref{fig:DataGathering}right), as opposed to previous methods that only show one card at a fixed distance. This enables us to make sure that the color restoration is consistent across different distances from the camera.
In addition, all of the images were taken using a pair of DSLR cameras (Nikon D810 with an AF-S NIKKOR 35mm f/1.8G ED lens, encased in a Hugyfot housing with a dome port) on a rigid rig, as shown in Fig.~\ref{fig:DataGathering}(left). Based on stereo imaging we recover the distances of objects from the camera (except for occlusions) and we can quantitatively evaluate the transmission maps, which are estimated from a single image. Using raw images is crucial in order to process the linear intensity, since the signal of highly attenuated areas is stored in the least significant bits of each pixel.
High resolution TIF images, raw images, camera calibration files, and the reconstructed scene distance maps can be downloaded from the project's webpage: \url{http://csms.haifa.ac.il/profiles/tTreibitz/datasets/ambient_forwardlooking/index.html}.

\begin{figure*}[tb]
\centering \addtolength{\tabcolsep}{-4pt}
\begin{tabular}{lr}
\includegraphics[height=3.4cm]{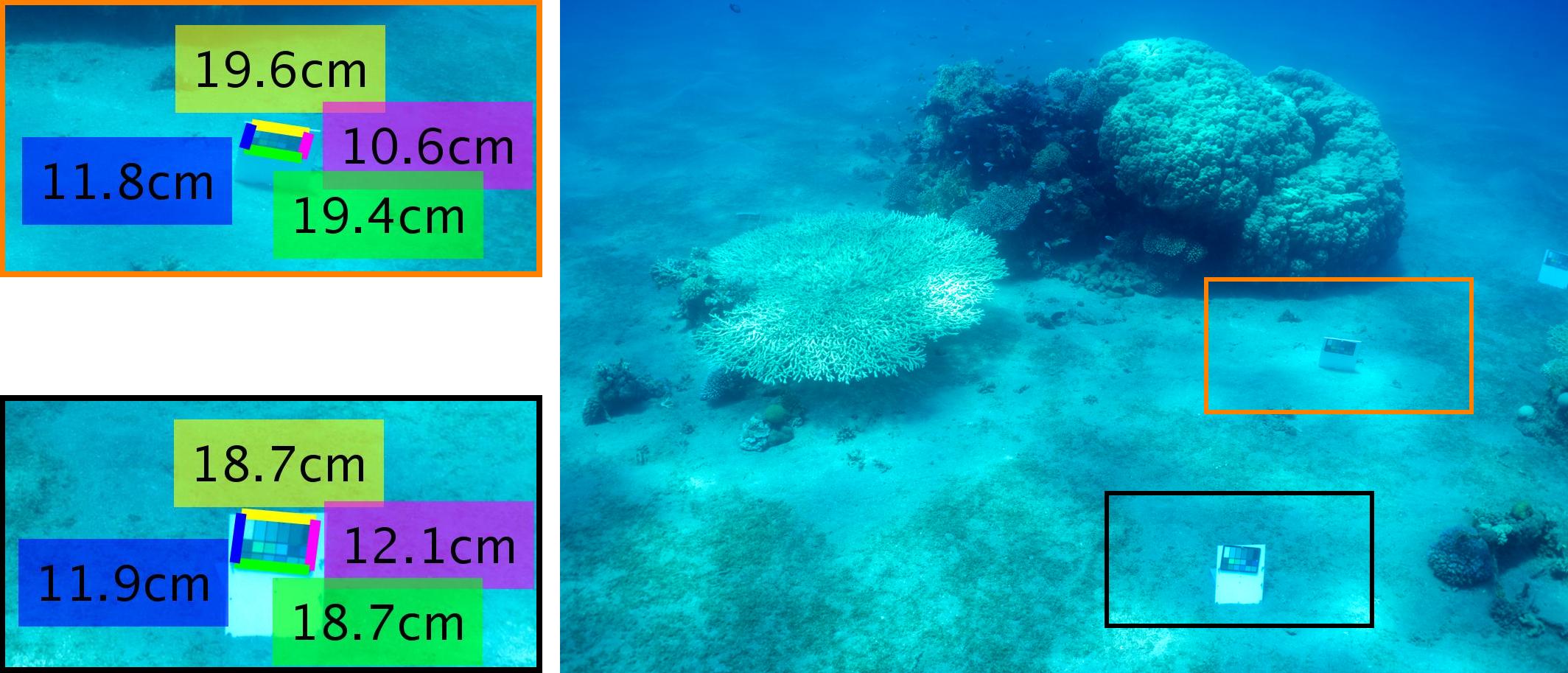} &
\includegraphics[height=3.4cm]{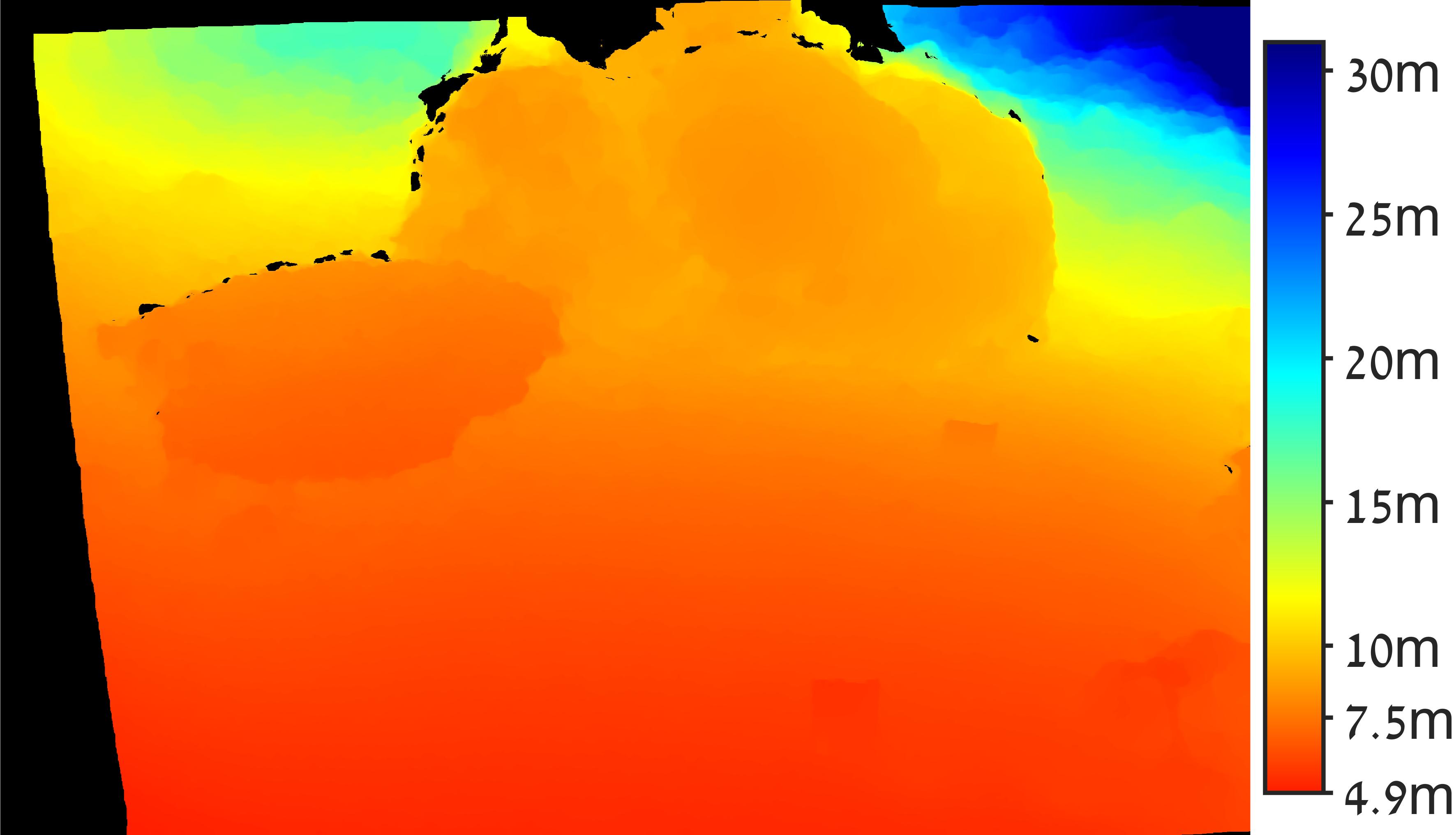}
\end{tabular}\vspace{-0.2cm}
\caption[Example scene from the underwater dataset we gathered.]{\textbf{Example scene from the dataset.} Middle: The image from the left camera. The sizes of two identical color chart were measured based on the 3D reconstruction, and are marked on the enlarged areas on the left. The true size of the chart is: $12\text{cm} \times 18.5 \text{cm}$. At a distance of $6\text{m}$ from the camera, the reconstruction accuracy is about $\sim 1.5\text{cm}$, which is quite accurate considering the precision of existing datasets~\cite{depthSingleImagePrecision}. Right: The distances of the objects from the left camera.}
\label{fig:UnderwaterSceneValidate} \vspace{-0.5cm}
\end{figure*}

After arriving in each dive location and focusing the cameras, we took 20-30 different images of a checkerboard. See example in Fig.~\ref{fig:DataGathering}(right). Using those images and the MATLAB stereo calibration toolbox, we extracted both the intrinsic and extrinsic parameters of the stereo cameras.
Given the intrinsic parameters, we corrected the lens distortions and got pairs of images for which the pin-hole camera model holds. The pin-hole model holds since we use a dome port and a wide-angle lens, such that the lens entrance pupil is at the center of the dome, and there is no refraction at the interface between the water and the housing.

The next step was to find matching points in both images in order to triangulate the real-world data points. This task is quite difficult in the areas further from the camera, due to the low signal-to-noise-ratio. In order to find dense matching we used EpicFlow~\cite{epicflow}, and applied it in a bi-directional manner: from the left image to the right and vice-versa. We considered a match to be valid only for pixels that had an end-point-error of 5 pixels or less for images with resolution of $1827 \times 2737$ pixels. A resolution of over 5MPixels is good for practical use cases, and was chosen due to memory constraints using EpicFlow.
The distance map was validated by measuring the size of the color charts in the scene, which is known to be $12.5\text{cm} \times 18\text{cm}$. Fig.~\ref{fig:UnderwaterSceneValidate} shows an example of an image from the dataset, with the dimensions of the color charts marked on the left, while the distance map of the scene is shown on the right.

\vspace{-0.2cm}
\subsection{Color Restoration - Qualitative comparison}

The proposed color restoration method in Sec.~\ref{sec:UW_algo} is based on a single image, and is evaluated as such. The color charts are used only for validation and are masked when fed to our algorithm. The transmission values of the masked pixels are determined based on neighboring values at the bottom of the chart.
The stereo pair was \emph{not} used for the color restoration, it was only used to generate the ground-truth distance map. This type of data enables us to quantitatively evaluate the accuracy of the estimated transmission.
The code is available at: \url{https://github.com/danaberman/underwater-hl}.

Figs.~\ref{fig:OurDataset1} and~\ref{fig:OurDataset2} show the images we acquired on the top row, followed by the output of different underwater image enhancement methods. A na\"{\i}ve contrast adjustment is also included as a baseline. The results of Drews \etal~\cite{drews2013transmission} and Peng \etal~\cite{blurrinessICIP2015} were generated using code released by the authors, while the results of Ancuti \etal~\cite{ancutiICPR2016, ancutiTIP2018, ancutiICIP2017} and Emberton \etal~\cite{Emberton2017} were provided by the authors.
A qualitative comparison shows that methods \cite{drews2013transmission,blurrinessICIP2015} are not able to consistently correct the colors in the images. These methods estimate the transmission and rely to some extent of the dark channel prior (DCP) which was proven empirically on outdoor scenes, but its assumptions do not always hold underwater~\cite{UnderwaterHazeLines}.
The methods by Ancuti \etal~\cite{ancutiICPR2016, ancutiTIP2018, ancutiICIP2017} focus on enhancing the image itself, rather than estimating an accurate transmission. Specifically, texture is extremely enhanced by~\cite{ancutiTIP2018}, as evident by the details on the distant area of the shipwreck in Fig.~\ref{fig:OurDataset1}. On the other hand, it also creates false textures on the foreground of the ship and in the sand in Fig.~\ref{fig:OurDataset1}.
In~\cite{ancutiICIP2017}, color transfer is used in conjunction with DCP-based transmission estimation to restore the color compensated image. The reference image for the color transfer in shown in the white inset on the left column of Fig.~\ref{fig:OurDataset1}, and is identical for all of the images shown in Figs.~\ref{fig:OurDataset1} and~\ref{fig:OurDataset2}. This method is rather effective in restoring the colors across different distances, as demonstrated for example by the color charts on the left column of Fig.~\ref{fig:OurDataset2}.

\vspace{-0.12cm}
\subsection{Transmission Estimation - Quantitative Evaluation}
\vspace{-0.08cm}

\begin{figure*}[b]   
\small\addtolength{\tabcolsep}{-3pt}
\begin{tabular}{lllll}
\rotatebox{90}{\hspace{0.8cm} Input} &
\begin{tikzpicture}
\draw (0, 0) node[inner sep=0] {\includegraphics[width=0.22\linewidth,trim=0 10 0 70, clip]{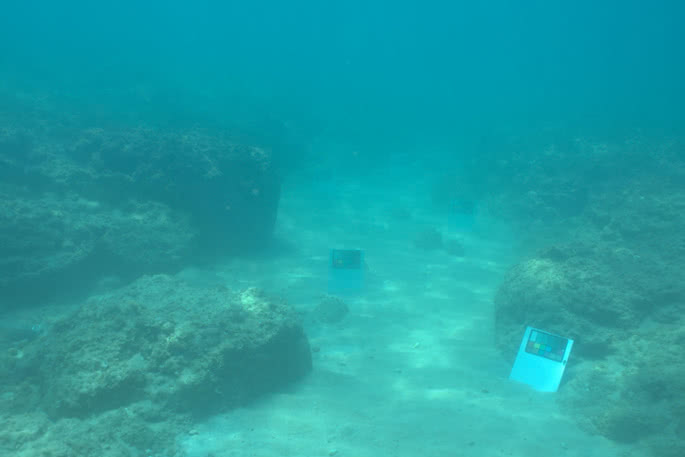}};
\draw (-0.103\linewidth, +0.95cm) node {\makebox[0pt][l]{\scriptsize\textbf{\color{white} R3008}}};
\end{tikzpicture} &
\begin{tikzpicture}
\draw (0, 0) node[inner sep=0] {\includegraphics[width=0.22\linewidth,trim=0 20 0 60, clip]{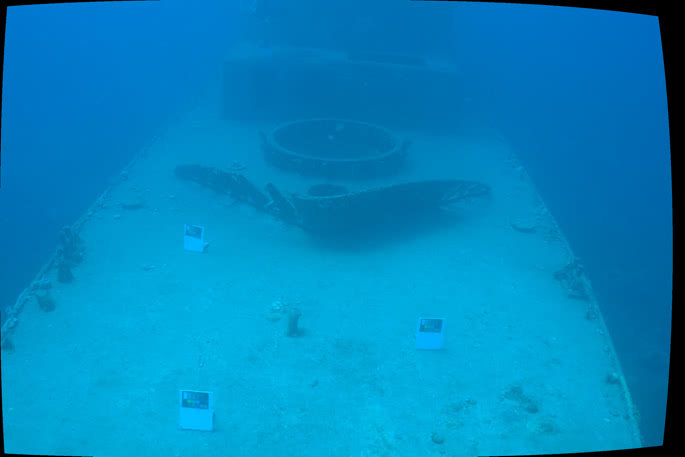}};
\draw (-0.103\linewidth, +0.95cm) node {\makebox[0pt][l]{\scriptsize\textbf{\color{white} R4376}}};
\end{tikzpicture} &
\begin{tikzpicture}
\draw (0, 0) node[inner sep=0] {\includegraphics[width=0.22\linewidth,trim=0 20 0 60, clip]{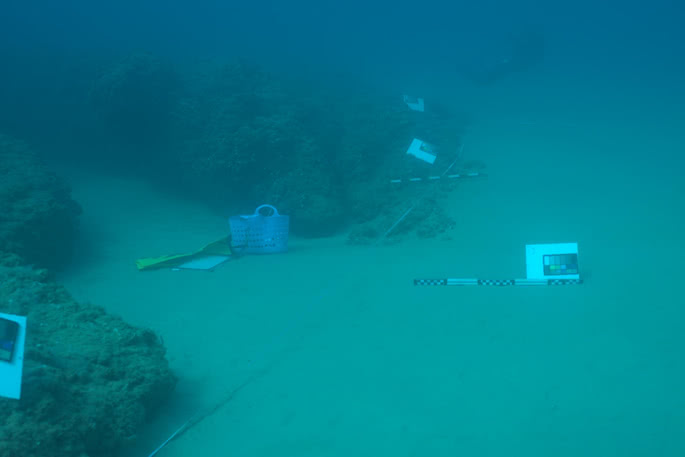}};
\draw (-0.103\linewidth, +0.95cm) node {\makebox[0pt][l]{\scriptsize\textbf{\color{white} R5478}}};
\end{tikzpicture} &
\begin{tikzpicture}
\draw (0, 0) node[inner sep=0] {\includegraphics[width=0.22\linewidth,trim=0 20 0 60, clip]{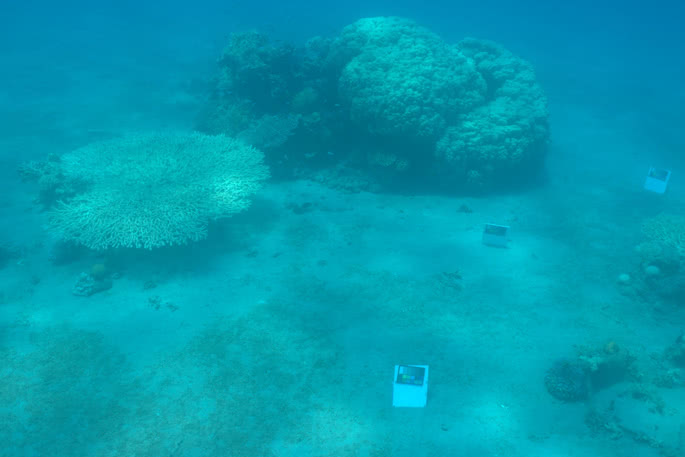}};
\draw (-0.103\linewidth, +0.95cm) node {\makebox[0pt][l]{\scriptsize\textbf{\color{white} R4491}}};
\end{tikzpicture} \\ 
\rotatebox{90}{\hspace{0.6cm} Contrast} &
\includegraphics[width=0.22\linewidth,trim=0 10 0 70, clip]{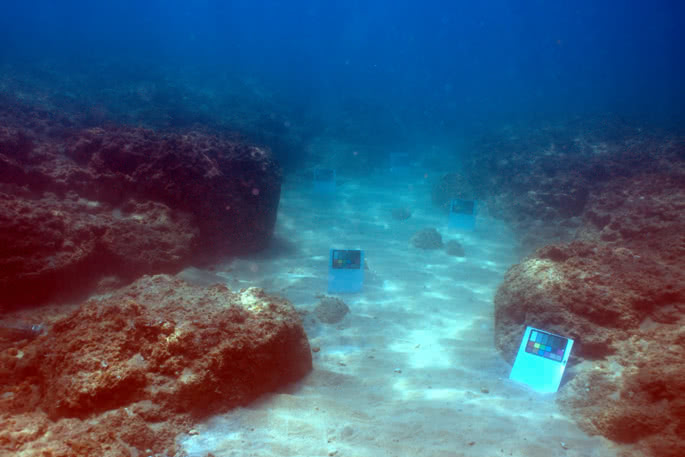} &
\includegraphics[width=0.22\linewidth,trim=0 20 0 60, clip]{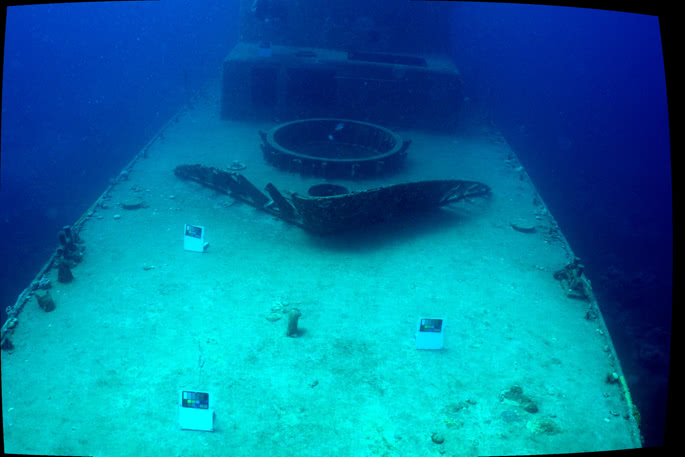} &
\includegraphics[width=0.22\linewidth,trim=0 20 0 60, clip]{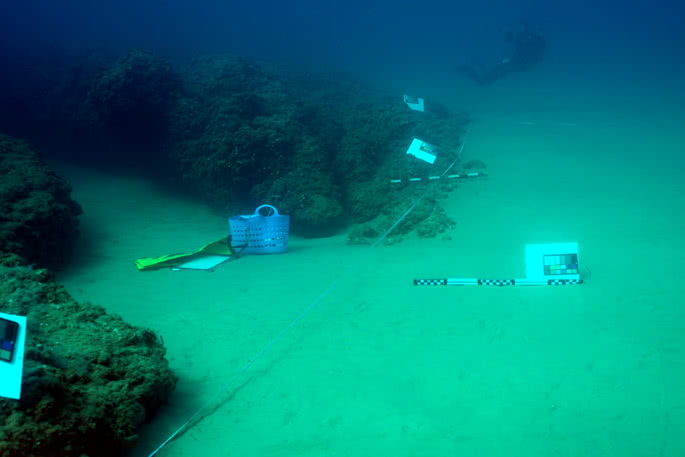} &
\includegraphics[width=0.22\linewidth,trim=0 20 0 60, clip]{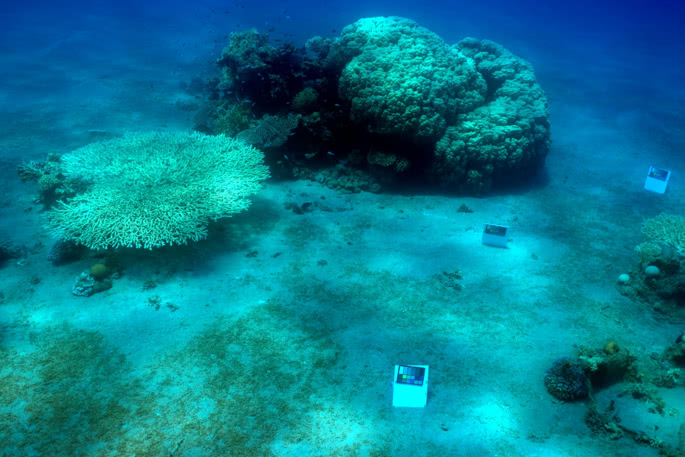} \\
\rotatebox{90}{\hspace{+0.1cm} Drews \etal\cite{drews2013transmission}} &
\includegraphics[width=0.22\linewidth]{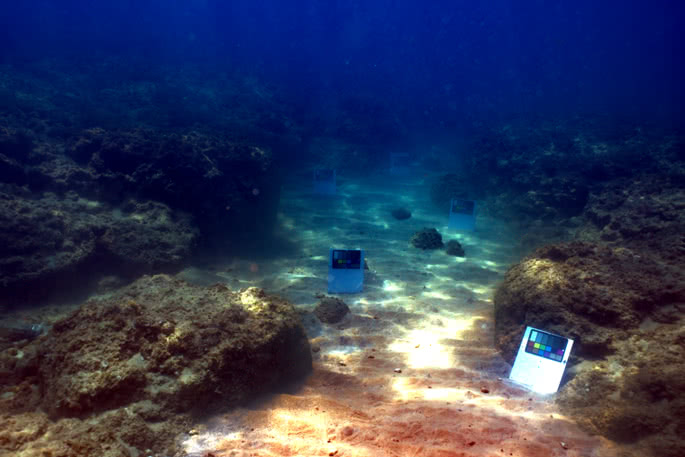} &
\includegraphics[width=0.22\linewidth]{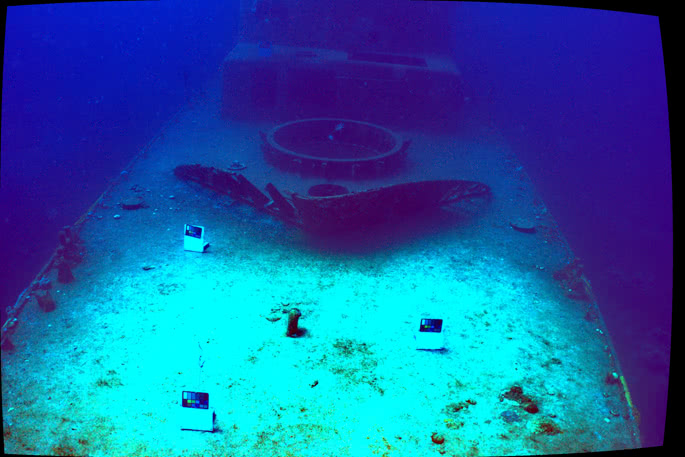} &
\includegraphics[width=0.22\linewidth]{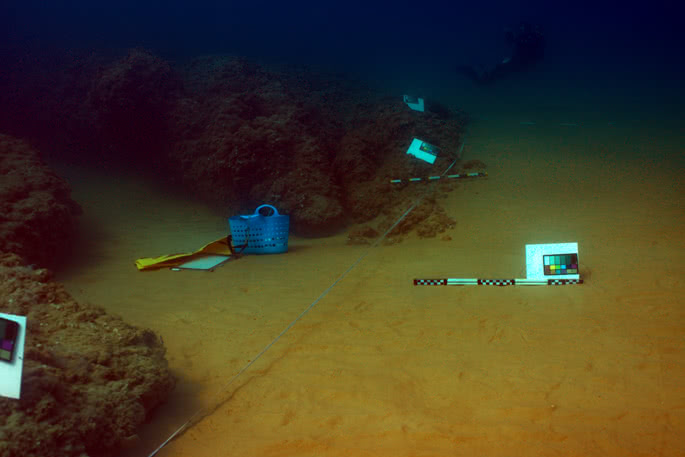} &
\includegraphics[width=0.22\linewidth]{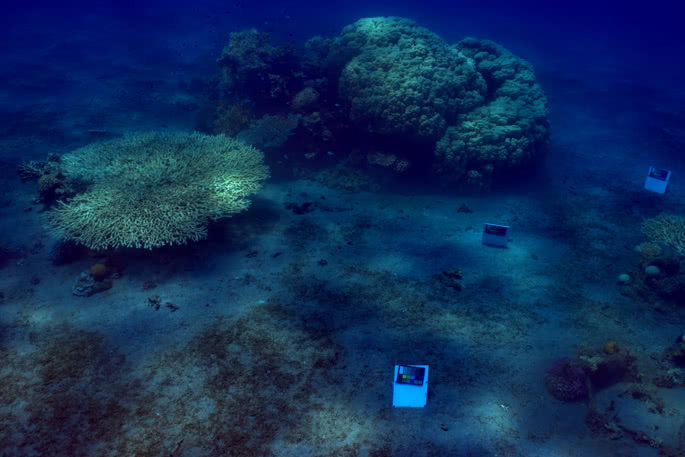} \\
\rotatebox{90}{\hspace{+0.2cm} Peng \etal\cite{blurrinessICIP2015}} &
\includegraphics[width=0.22\linewidth]{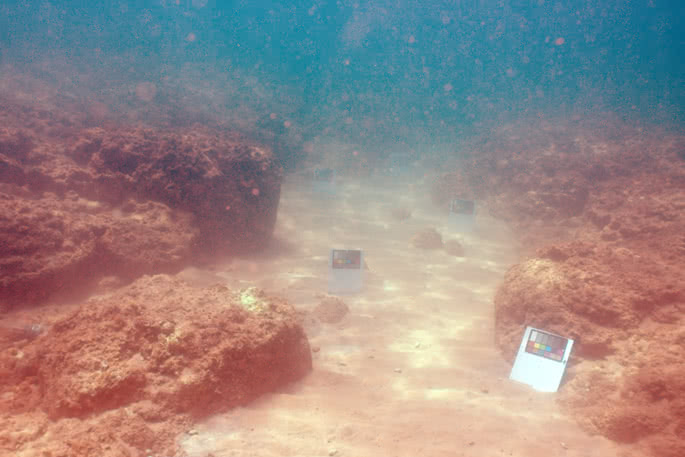} &
\includegraphics[width=0.22\linewidth]{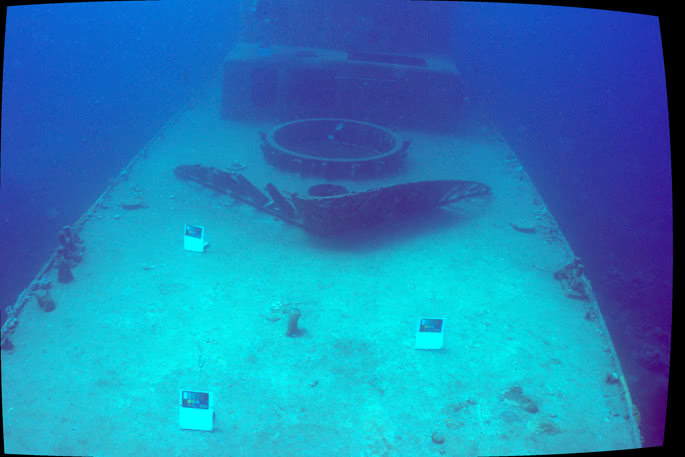} &
\includegraphics[width=0.22\linewidth]{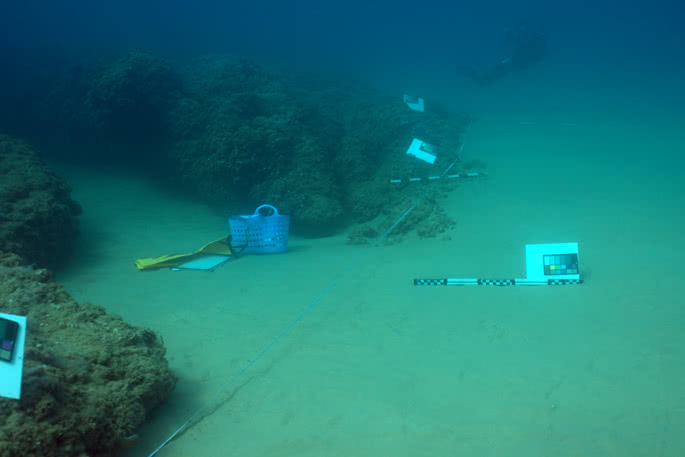} &
\includegraphics[width=0.22\linewidth]{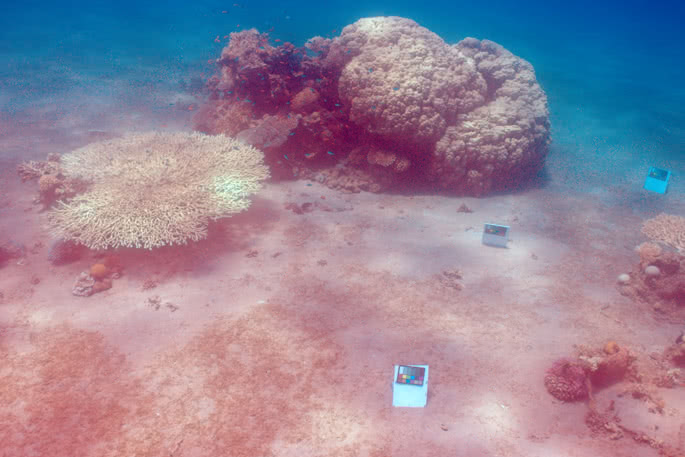} \\
\rotatebox{90}{\hspace{+0.1cm} Ancuti \etal\cite{ancutiICPR2016}} &
\includegraphics[width=0.22\linewidth]{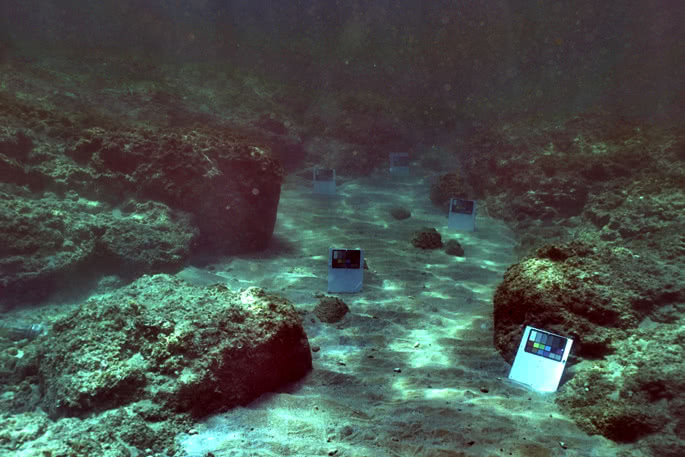} &
\includegraphics[width=0.22\linewidth]{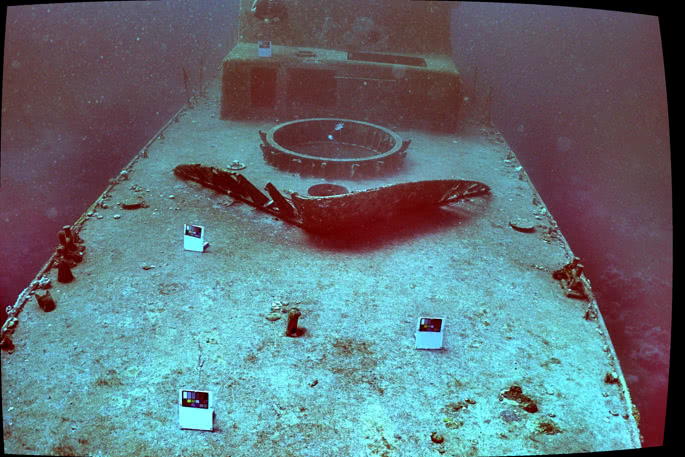} &
\includegraphics[width=0.22\linewidth]{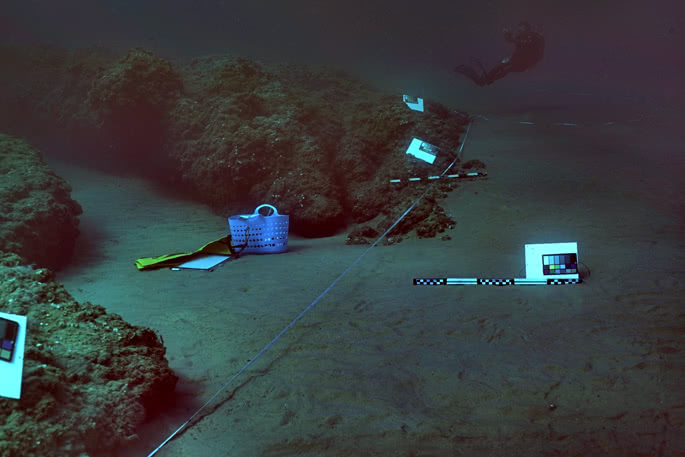} &
\includegraphics[width=0.22\linewidth]{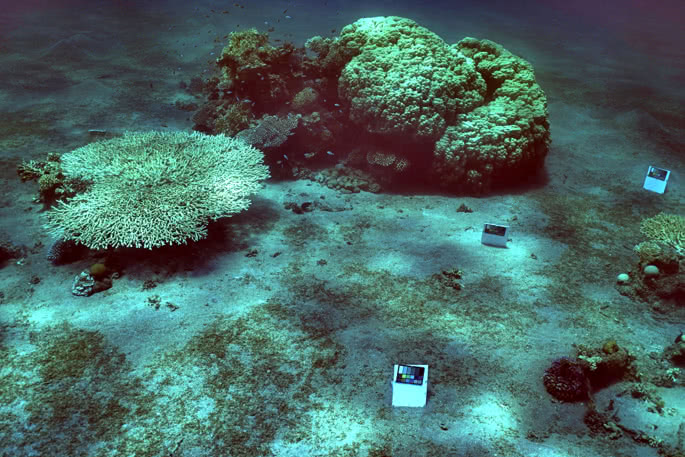} \\
\rotatebox{90}{\hspace{+0.1cm} Ancuti \etal\cite{ancutiICIP2017}} &
\includegraphics[width=0.22\linewidth]{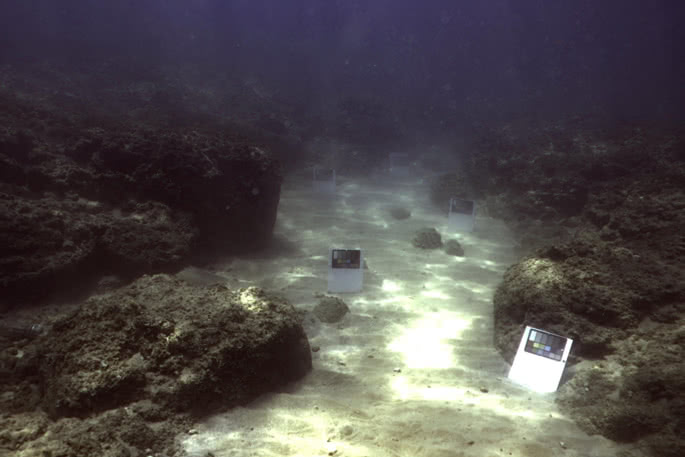} &
\includegraphics[width=0.22\linewidth]{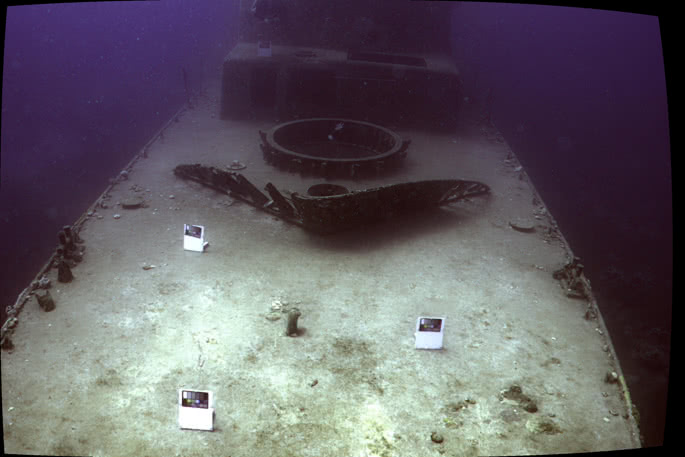} &
\includegraphics[width=0.22\linewidth]{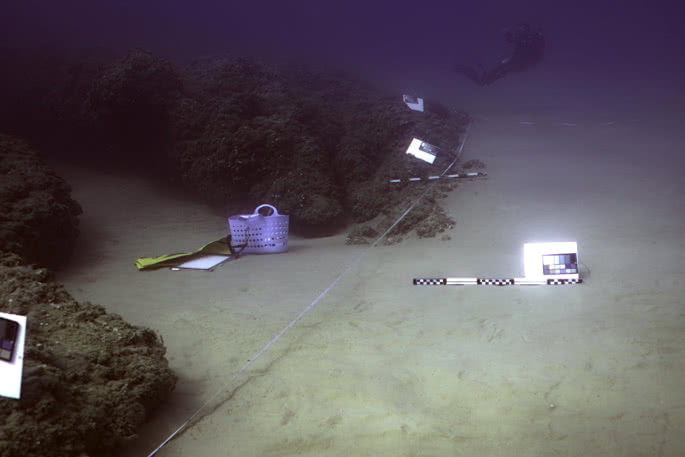} &
\includegraphics[width=0.22\linewidth]{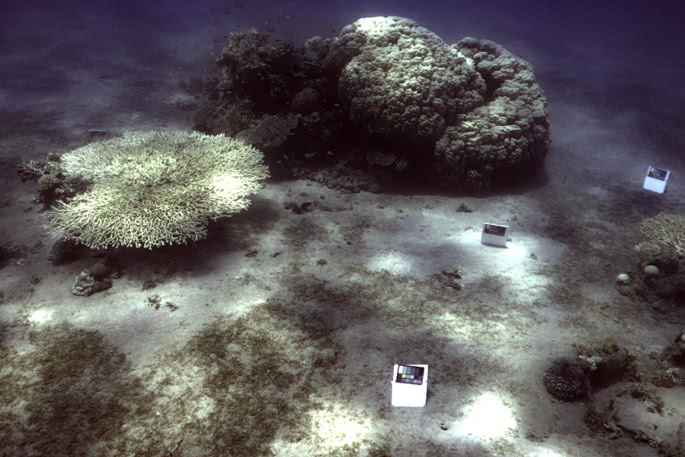} \\
\rotatebox{90}{\hspace{+0.15cm}Ancuti \etal\cite{ancutiTIP2018}} &
\includegraphics[width=0.22\linewidth]{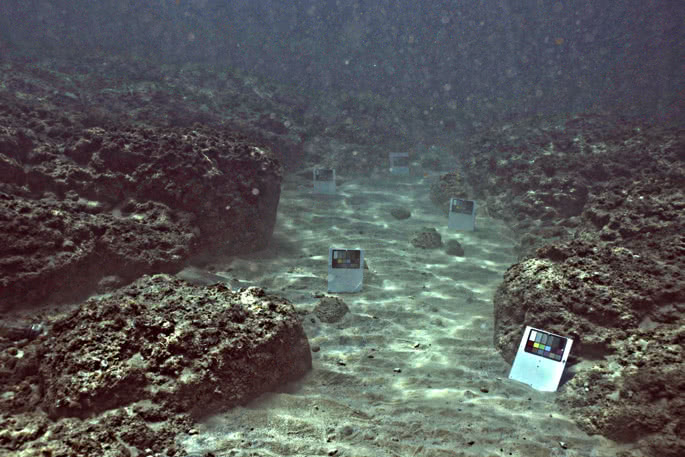} &
\includegraphics[width=0.22\linewidth]{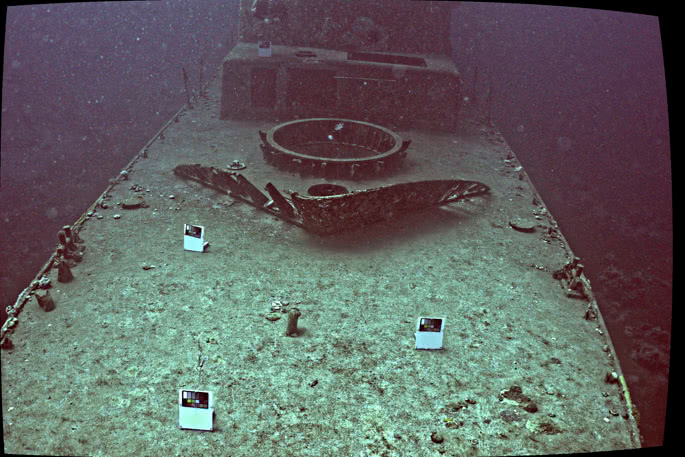} &
\includegraphics[width=0.22\linewidth]{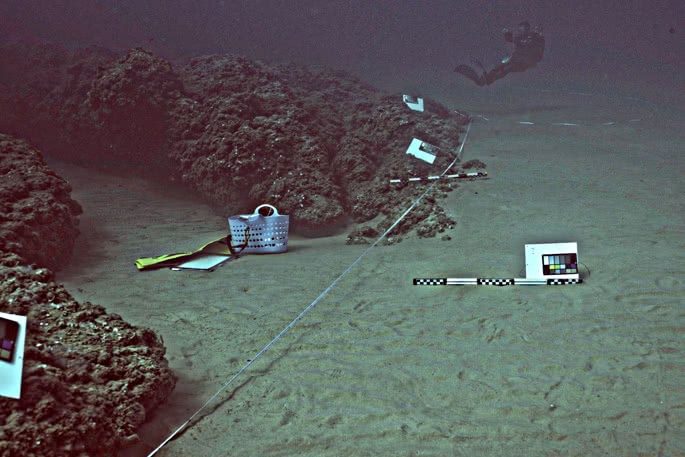} &
\includegraphics[width=0.22\linewidth]{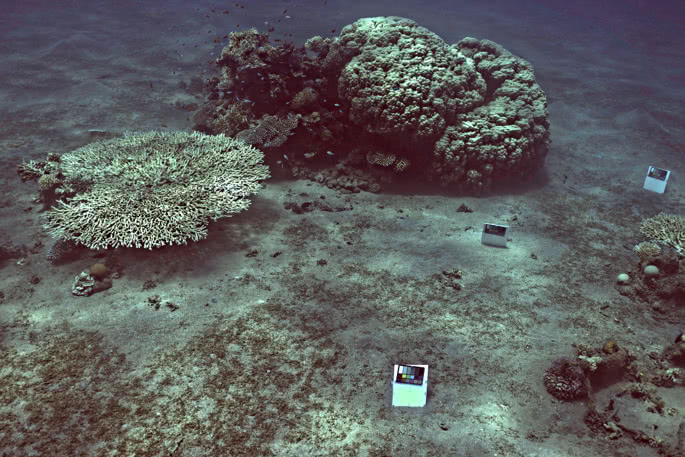} \\
\rotatebox{90}{\hspace{-0.1cm} Emberton \etal\cite{Emberton2017}} &
\includegraphics[width=0.22\linewidth]{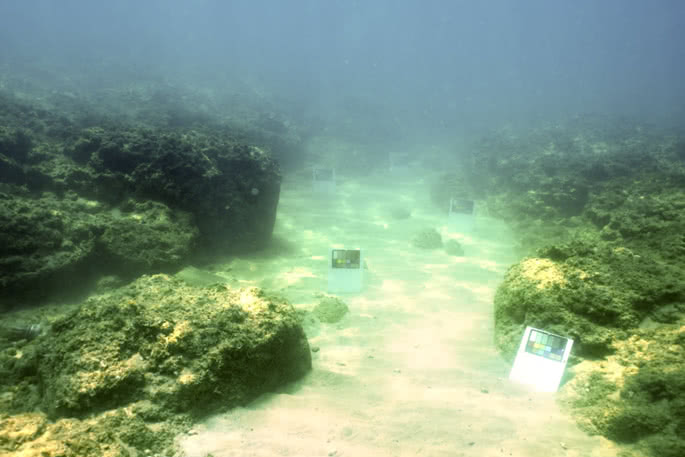} &
\includegraphics[width=0.22\linewidth]{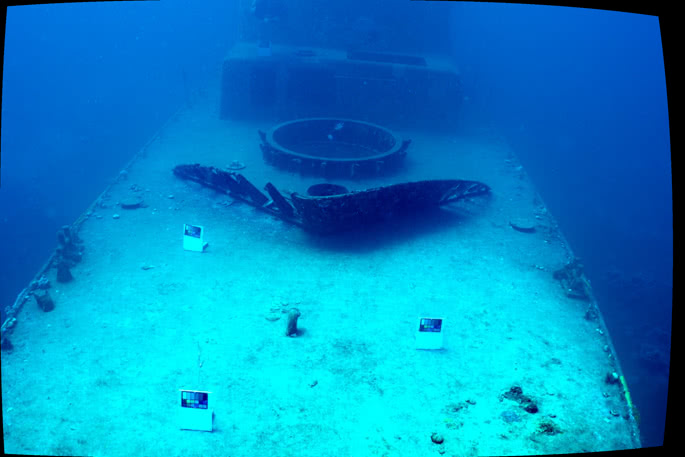} &
\includegraphics[width=0.22\linewidth]{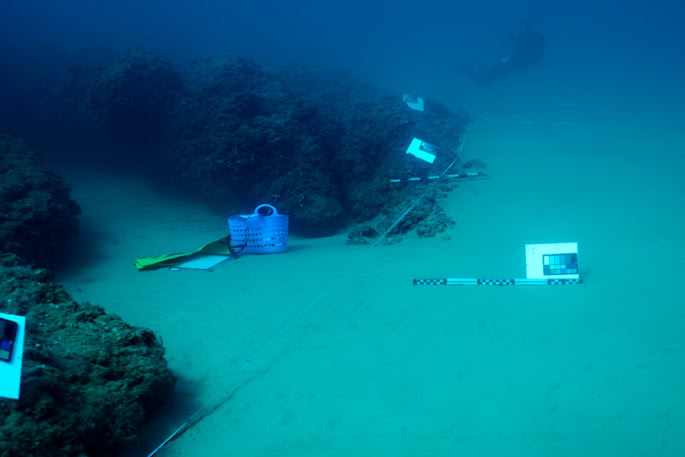} &
\includegraphics[width=0.22\linewidth]{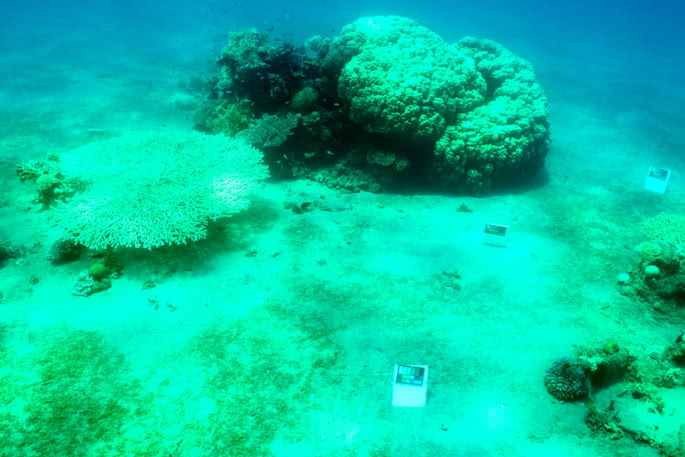} \\
\rotatebox{90}{\hspace{0.9cm} Ours} &
\includegraphics[width=0.22\linewidth]{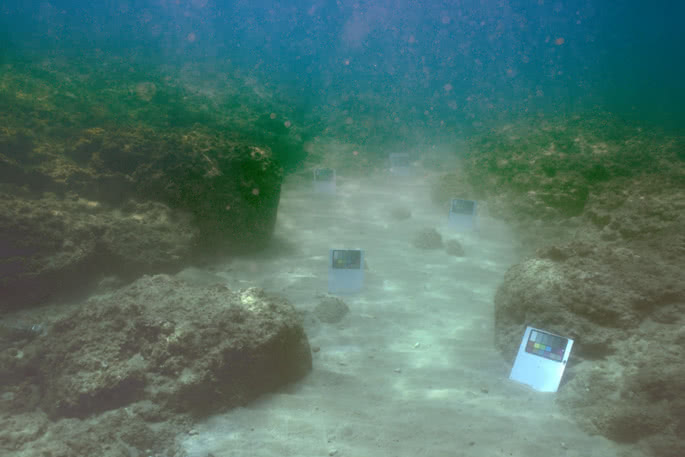} &
\includegraphics[width=0.22\linewidth]{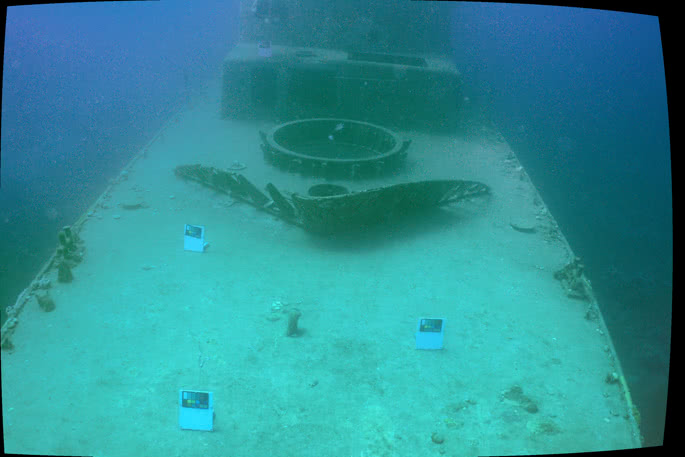} &
\includegraphics[width=0.22\linewidth]{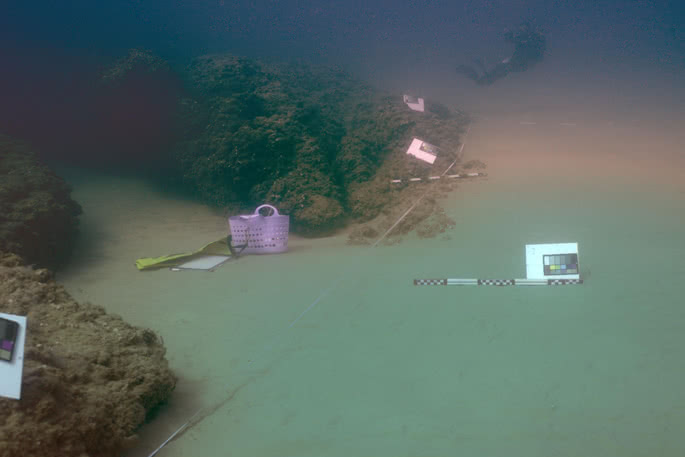} &
\includegraphics[width=0.22\linewidth]{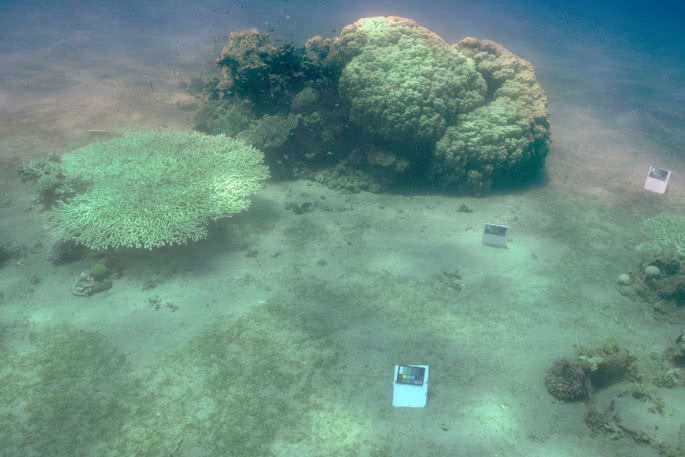} \\
\end{tabular} \vspace{-0.2cm}
\caption[Underwater image enhancement: comparison of single-image underwater color restoration algorithms]{Comparison of single-image underwater enhancement techniques.}
\label{fig:OurDataset1}
\end{figure*}

\begin{figure*}[b]   
\centering \small\addtolength{\tabcolsep}{-4pt}
\begin{tabular}{ccccc}
\rotatebox{90}{\hspace{0.8cm} Input} &
\begin{tikzpicture}
\draw (0, 0) node[inner sep=0] {\includegraphics[width=0.23\linewidth]{RGT_3008_input_rgb.jpg}};
\draw (-0.108\linewidth, +1.18cm) node {\makebox[0pt][l]{\scriptsize\textbf{\color{white} R3008}}};
\end{tikzpicture} &
\begin{tikzpicture}
\draw (0, 0) node[inner sep=0] {\includegraphics[width=0.23\linewidth]{RGT_4376_input_rgb.jpg}};
\draw (-0.108\linewidth, +1.18cm) node {\makebox[0pt][l]{\scriptsize\textbf{\color{white} R4376}}};
\end{tikzpicture} &
\begin{tikzpicture}
\draw (0, 0) node[inner sep=0] {\includegraphics[width=0.23\linewidth]{RGT_5478_input_rgb.jpg}};
\draw (-0.108\linewidth, +1.18cm) node {\makebox[0pt][l]{\scriptsize\textbf{\color{white} R5478}}};
\end{tikzpicture} &
\begin{tikzpicture}
\draw (0, 0) node[inner sep=0] {\includegraphics[width=0.23\linewidth]{RGT_4491_input_rgb.jpg}};
\draw (-0.108\linewidth, +1.18cm) node {\makebox[0pt][l]{\scriptsize\textbf{\color{white} R4491}}};
\end{tikzpicture}  \vspace{-0.0cm} \\  \vspace{+0.0cm}
\rotatebox{90}{\hspace{0.6cm} Distance} &
\includegraphics[width=0.23\linewidth]{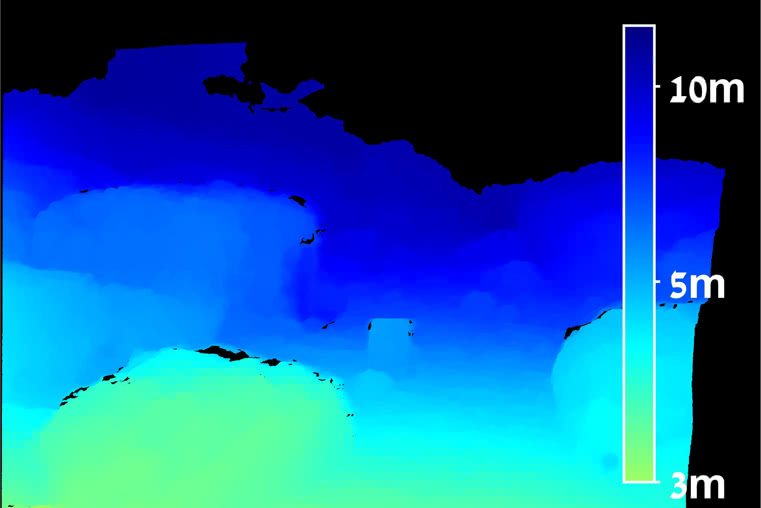} &
\includegraphics[width=0.23\linewidth]{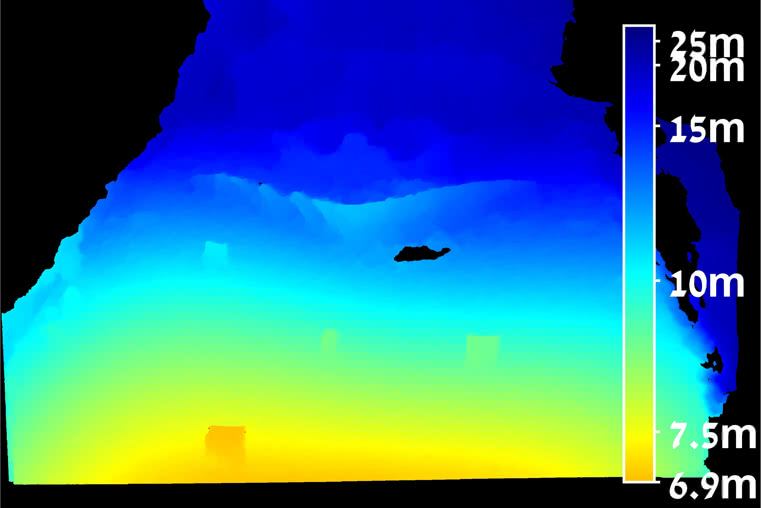} &
\includegraphics[width=0.23\linewidth]{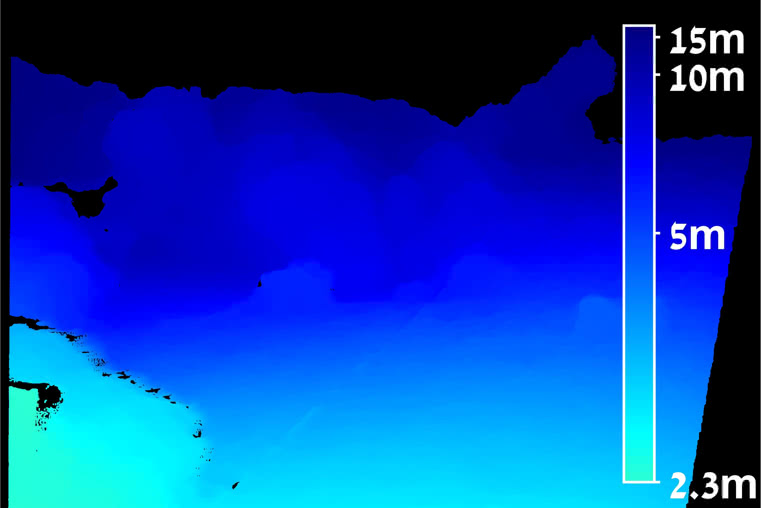} &
\includegraphics[width=0.23\linewidth]{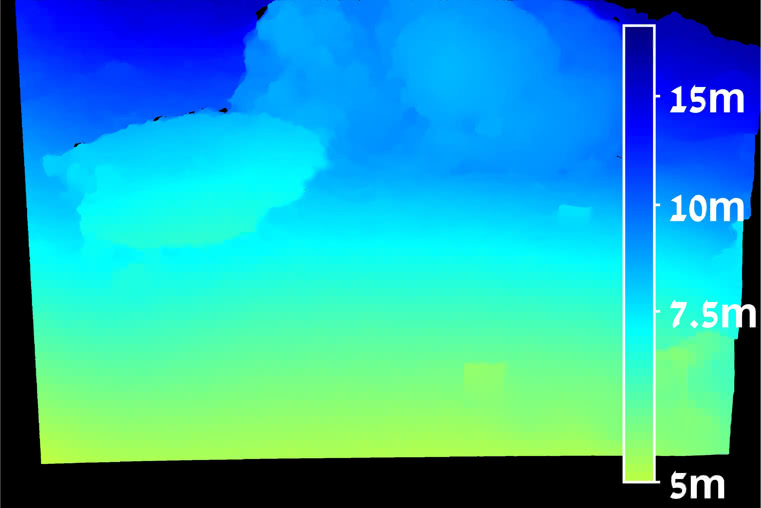} \vspace{-0.0cm} \\  \vspace{+0.0cm}
\rotatebox{90}{\hspace{+0.2cm} Drews \etal\cite{drews2013transmission}} &
\includegraphics[width=0.23\linewidth]{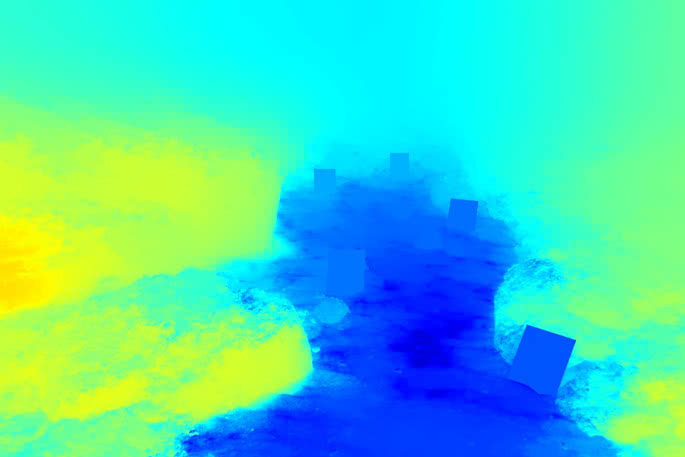} &
\includegraphics[width=0.23\linewidth]{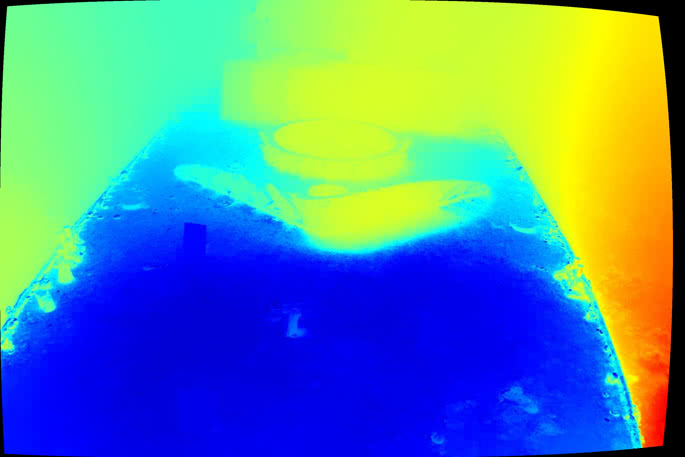} &
\includegraphics[width=0.23\linewidth]{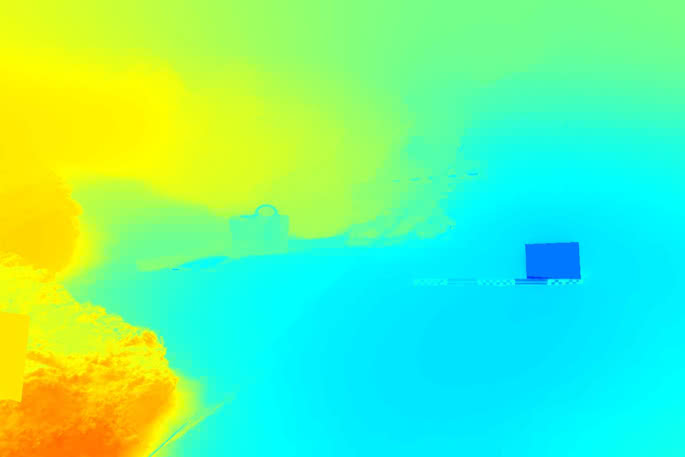} &
\includegraphics[width=0.23\linewidth]{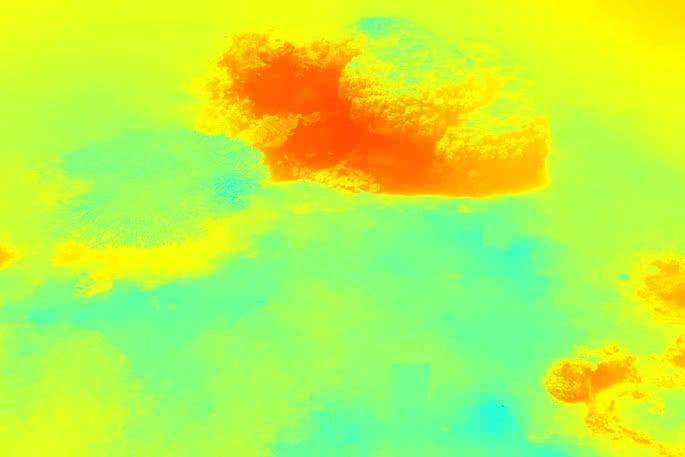} \vspace{-0.05cm} \\  \vspace{+0.1cm}
& $\rho = -0.30$  &  $\rho = -0.73$  &  $\rho = -0.28$  &  $\rho = -0.45$ \\ \vspace{+0.1cm}
\rotatebox{90}{\hspace{+0.3cm} Peng \etal\cite{blurrinessICIP2015}} &
\includegraphics[width=0.23\linewidth]{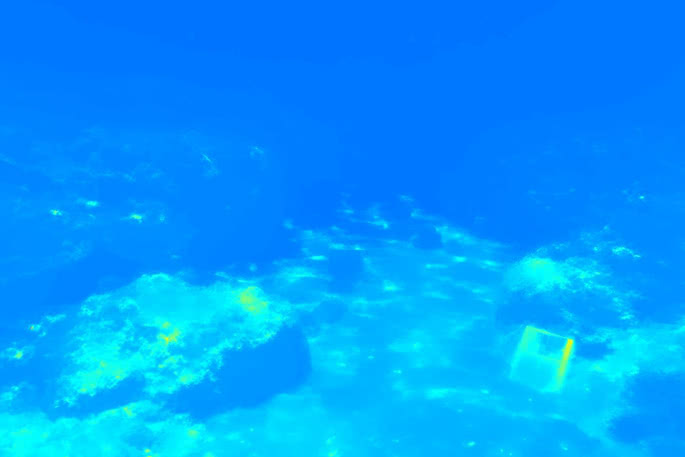} &
\includegraphics[width=0.23\linewidth]{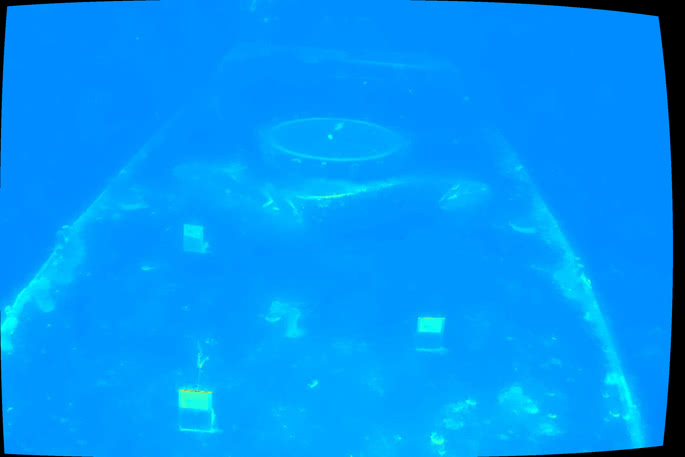} &
\includegraphics[width=0.23\linewidth]{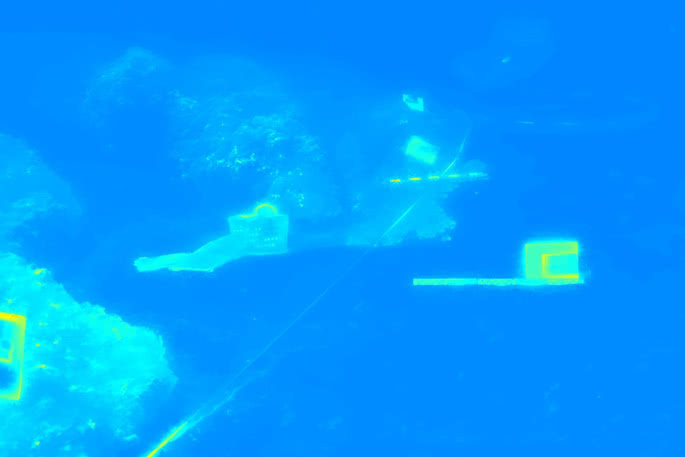} &
\includegraphics[width=0.23\linewidth]{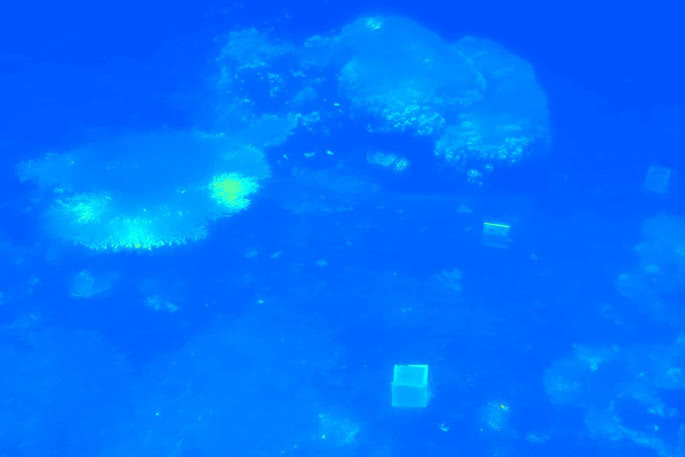} \vspace{-0.15cm} \\  \vspace{+0.1cm}
& $\rho = 0.72$  &  $\rho = 0.44$  &  $\rho = 0.17$  &  $\rho = 0.05$ \\
\rotatebox{90}{\hspace{+0.1cm} Ancuti \etal\cite{ancutiICIP2017}} &
\includegraphics[width=0.23\linewidth]{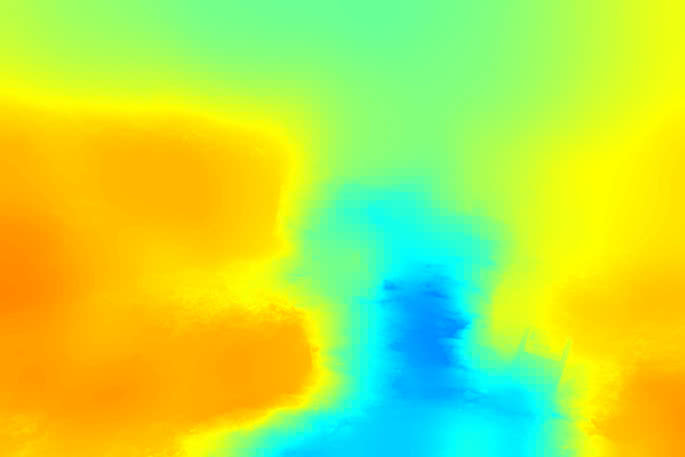} &
\includegraphics[width=0.23\linewidth]{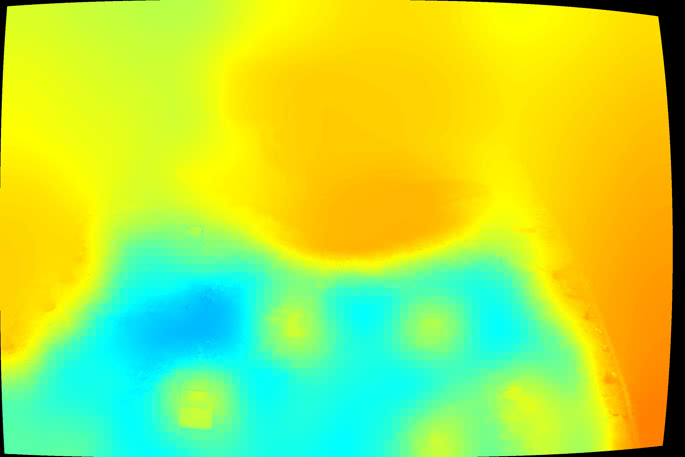} &
\includegraphics[width=0.23\linewidth]{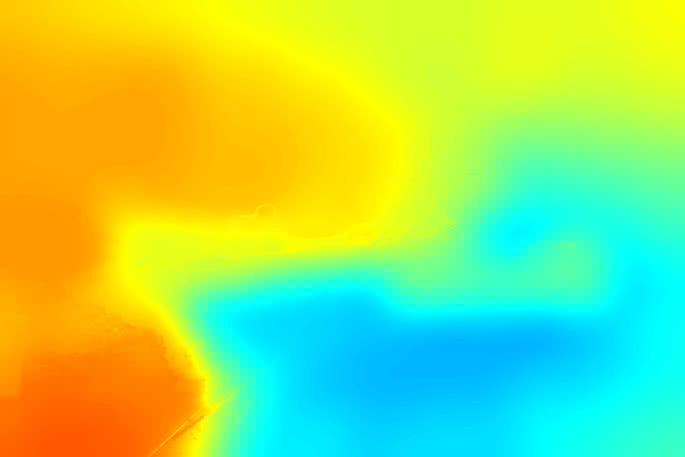} &
\includegraphics[width=0.23\linewidth]{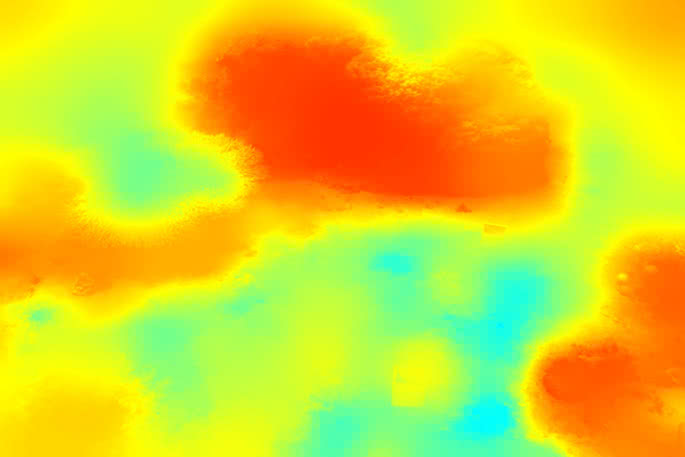}  \vspace{-0.05cm} \\  \vspace{+0.1cm}
& $\rho = -0.15$  &  $\rho = -0.58$  &  $\rho = -0.40$  &  $\rho = -0.34$ \\
\rotatebox{90}{\hspace{-0.05cm} Emberton \etal\cite{Emberton2017}} &
\includegraphics[width=0.23\linewidth]{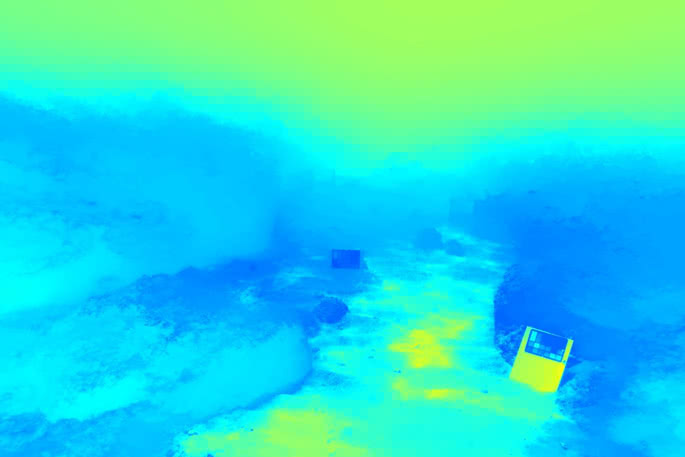} &
\includegraphics[width=0.23\linewidth]{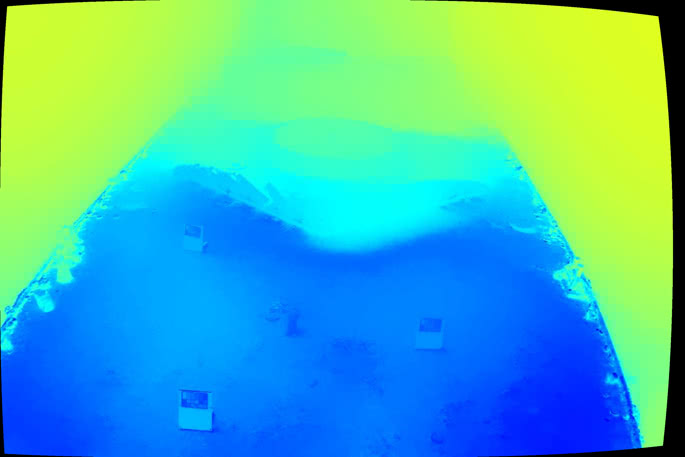} &
\includegraphics[width=0.23\linewidth]{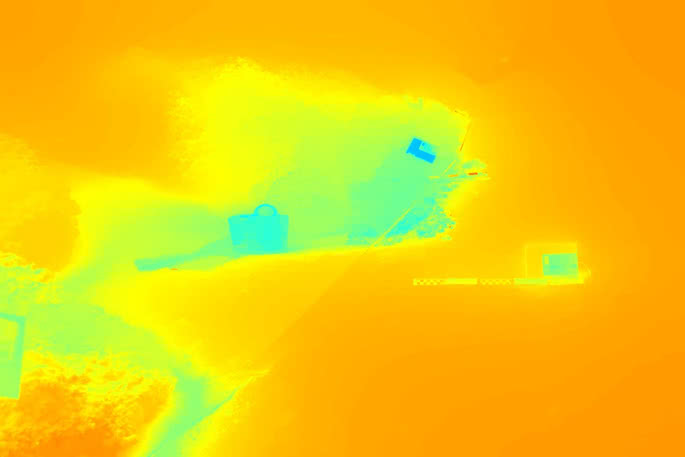} &
\includegraphics[width=0.23\linewidth]{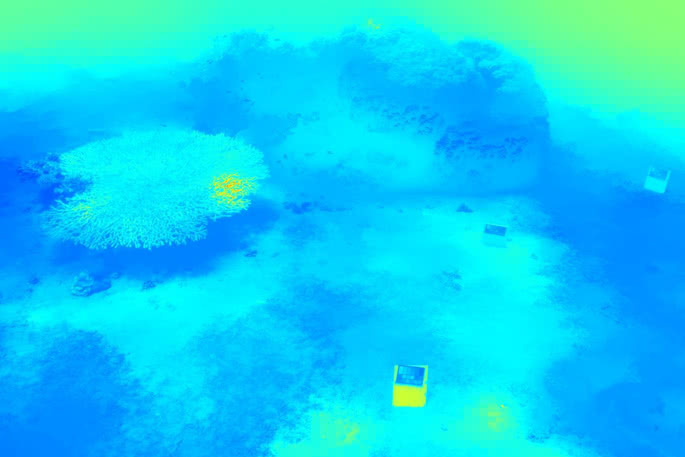} \vspace{-0.05cm} \\  \vspace{+0.1cm}
& $\rho = -0.05$  &  $\rho = -0.61$  &  $\rho = 0.10$  &  $\rho = -0.53$ \\
\rotatebox{90}{\hspace{0.9cm} Ours} &
\includegraphics[width=0.23\linewidth]{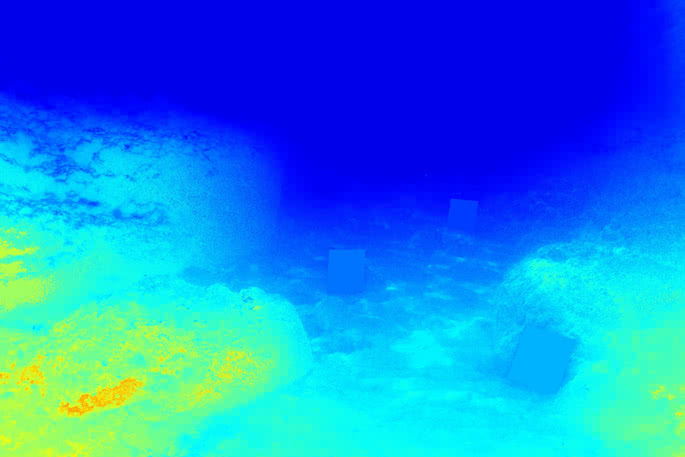} &
\includegraphics[width=0.23\linewidth]{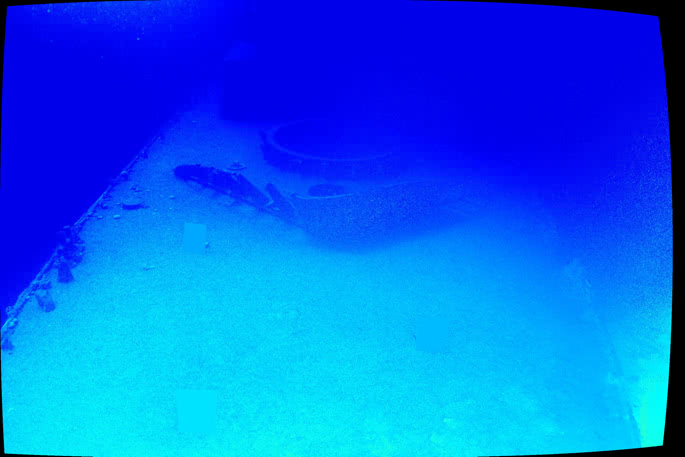} &
\includegraphics[width=0.23\linewidth]{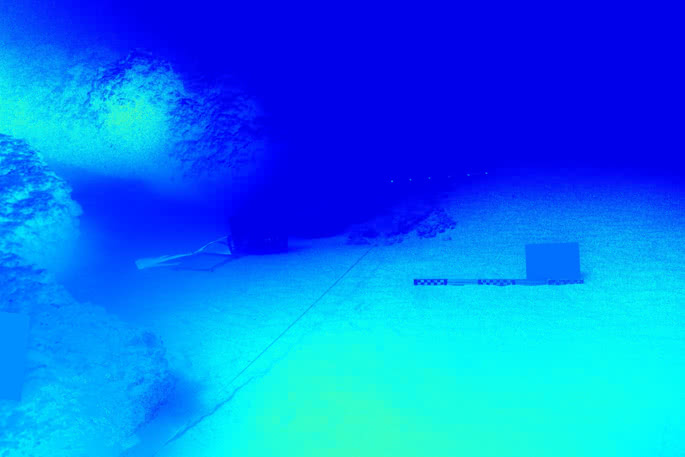} &
\includegraphics[width=0.23\linewidth]{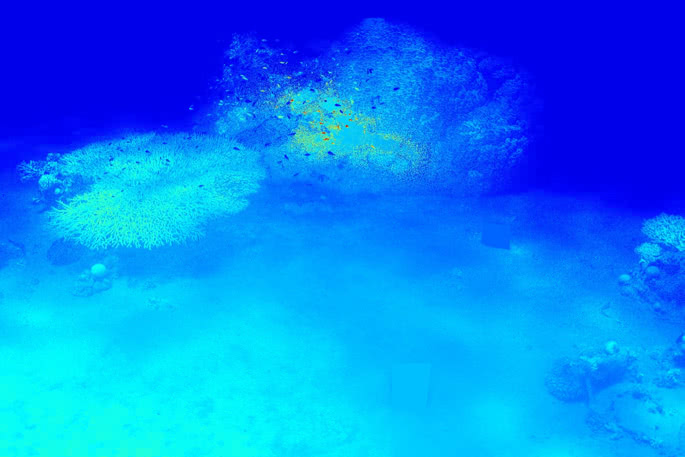} \vspace{-0.05cm} \\  \vspace{+0.1cm}
& $\rho = 0.93$  &  $\rho = 0.75$  &  $\rho = 0.66$  &  $\rho = 0.80$ \\
\end{tabular} \vspace{-0.1cm}
\caption[Underwater image enhancement: true distances and estimated transmission maps]{True distances and estimated transmission maps for images shown in Fig.~\ref{fig:OurDataset1}, shown only for methods that estimate a transmission map. The quality of the transmission is measured by the Pearson correlation coefficient $\rho$, which is calculated between the negative logarithm of the estimated transmission and the true distance.}
\label{fig:OurDatasetTrans1}
\end{figure*}

\begin{figure*}[b]   
\small\addtolength{\tabcolsep}{-3pt}
\begin{tabular}{lllll}
\rotatebox{90}{\hspace{0.8cm} Input} &
\begin{tikzpicture}
\draw (0, 0) node[inner sep=0] {\includegraphics[width=0.22\linewidth,trim=0 10 0 70, clip]{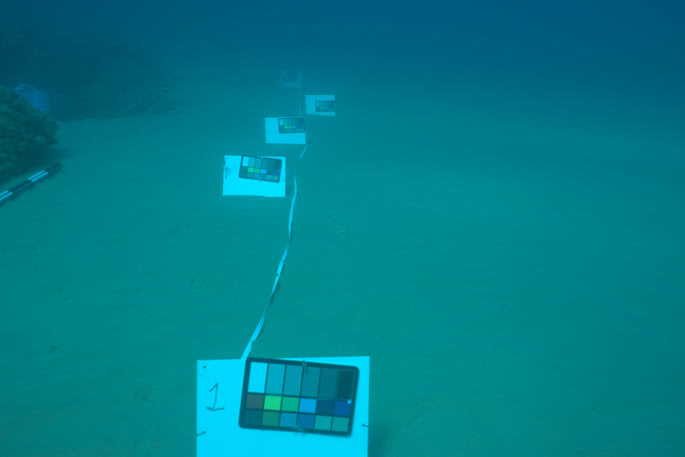}};
\draw (-0.103\linewidth, +0.95cm) node {\makebox[0pt][l]{\scriptsize\textbf{\color{white} R5450}}};
\end{tikzpicture} &
\begin{tikzpicture}
\draw (0, 0) node[inner sep=0] {\includegraphics[width=0.22\linewidth,trim=0 20 0 60, clip]{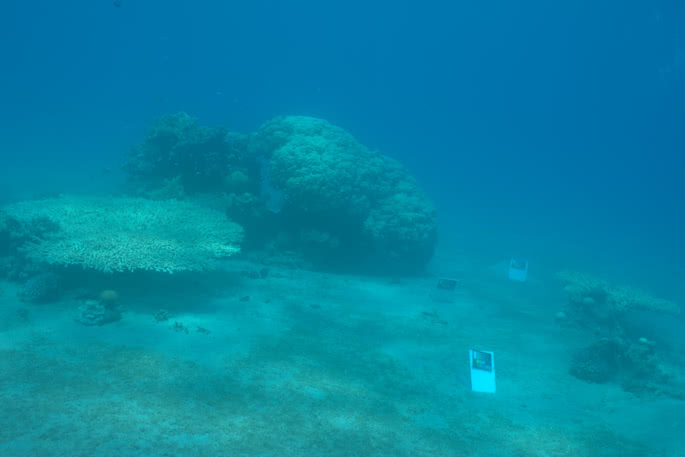}};
\draw (-0.103\linewidth, +0.95cm) node {\makebox[0pt][l]{\scriptsize\textbf{\color{white} R4485}}};
\end{tikzpicture} &
\begin{tikzpicture}
\draw (0, 0) node[inner sep=0] {\includegraphics[width=0.22\linewidth,trim=0 20 0 60, clip]{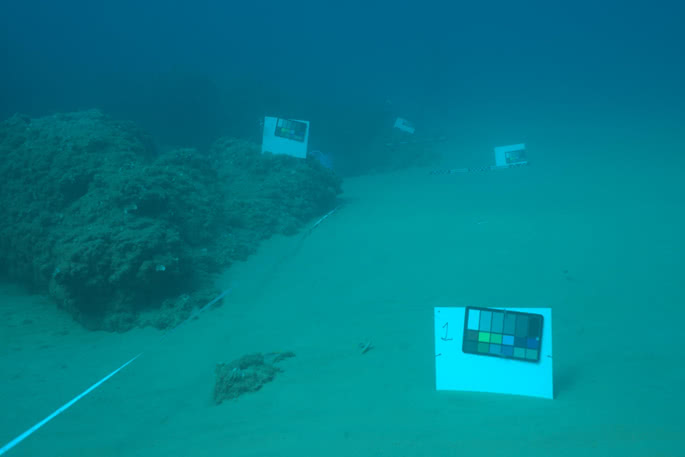}};
\draw (-0.103\linewidth, +0.95cm) node {\makebox[0pt][l]{\scriptsize\textbf{\color{white} R5469}}};
\end{tikzpicture} &
\begin{tikzpicture}
\draw (0, 0) node[inner sep=0] {\includegraphics[width=0.22\linewidth,trim=0 20 0 60, clip]{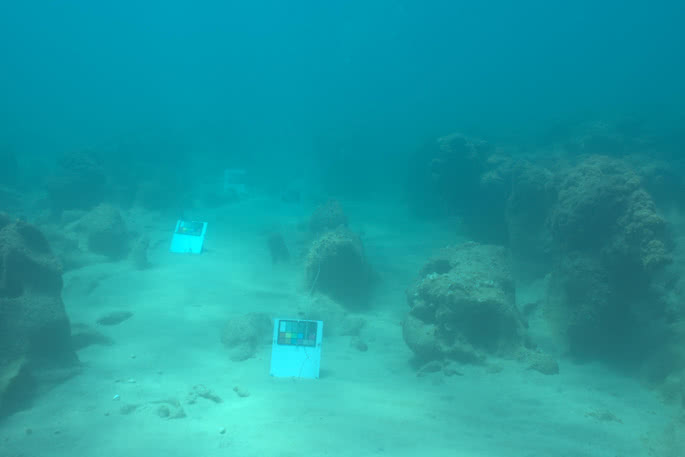}};
\draw (-0.103\linewidth, +0.95cm) node {\makebox[0pt][l]{\scriptsize\textbf{\color{white} R3204}}};
\end{tikzpicture} \\
\rotatebox{90}{\hspace{0.6cm} Contrast} &
\includegraphics[width=0.22\linewidth,trim=0 10 0 70, clip]{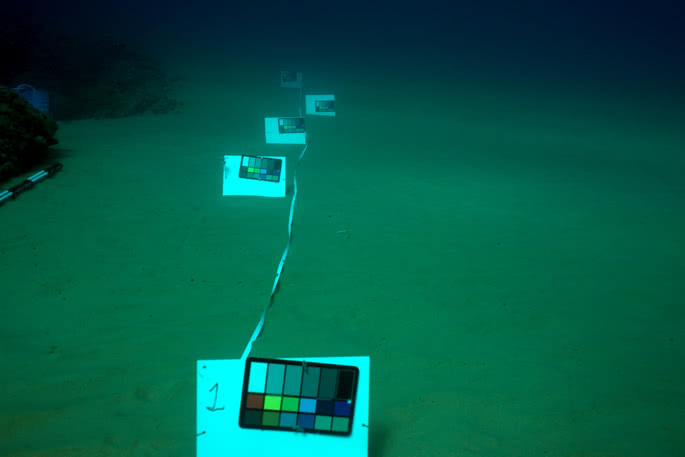} &
\includegraphics[width=0.22\linewidth,trim=0 20 0 60, clip]{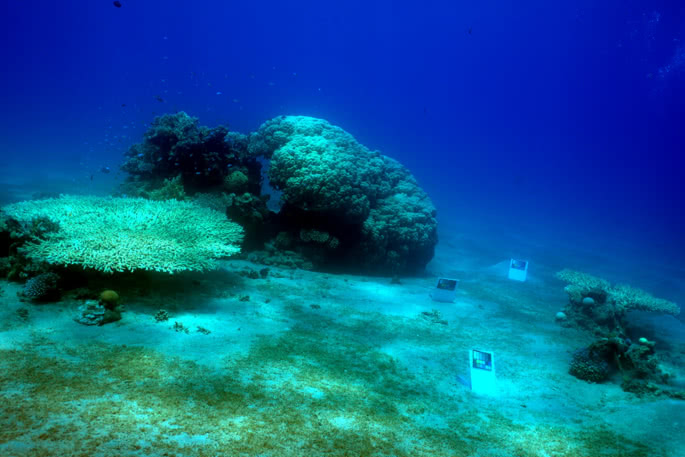} &
\includegraphics[width=0.22\linewidth,trim=0 20 0 60, clip]{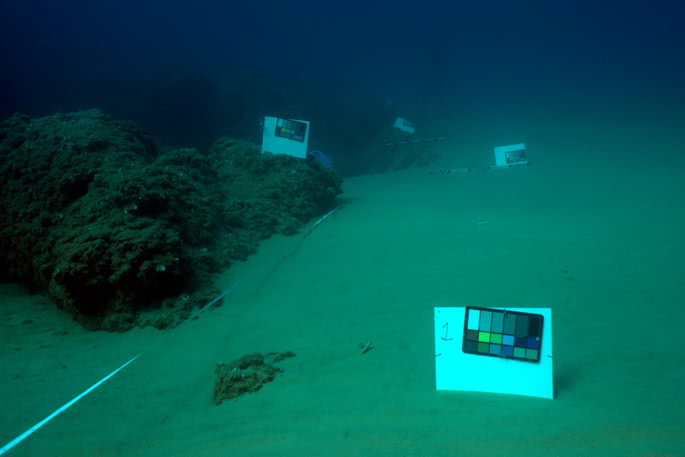} &
\includegraphics[width=0.22\linewidth,trim=0 20 0 60, clip]{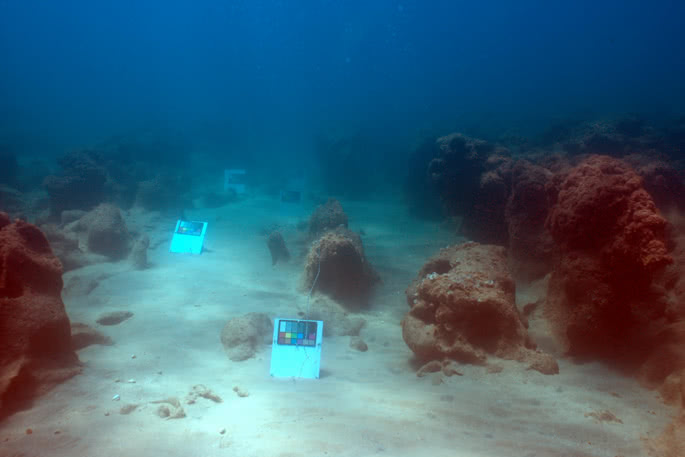} \\
\rotatebox{90}{\hspace{0.1cm} Drews \etal\cite{drews2013transmission}} &
\includegraphics[width=0.22\linewidth]{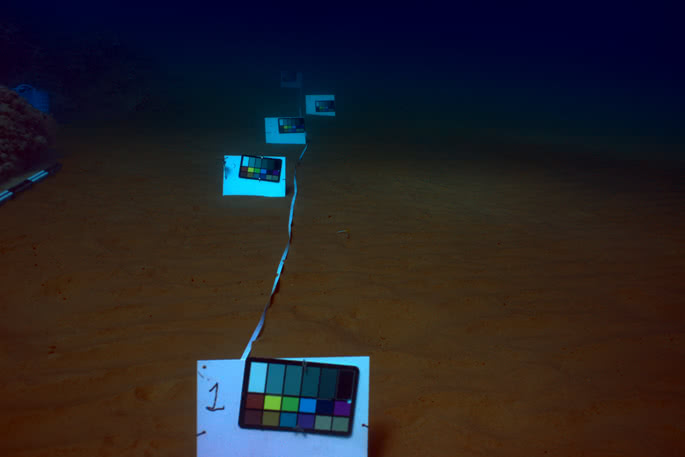} &
\includegraphics[width=0.22\linewidth]{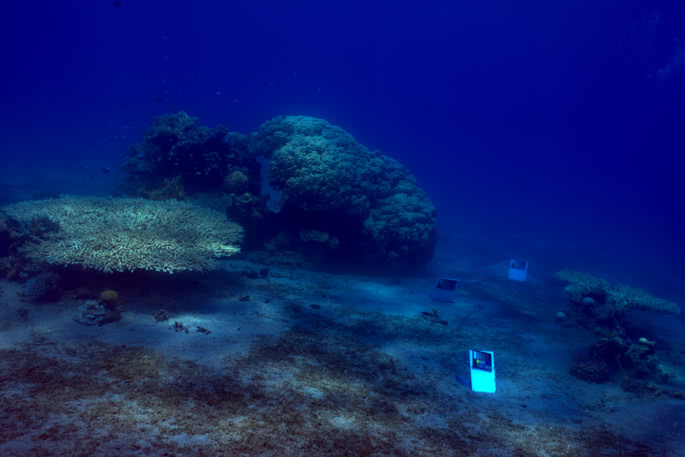} &
\includegraphics[width=0.22\linewidth]{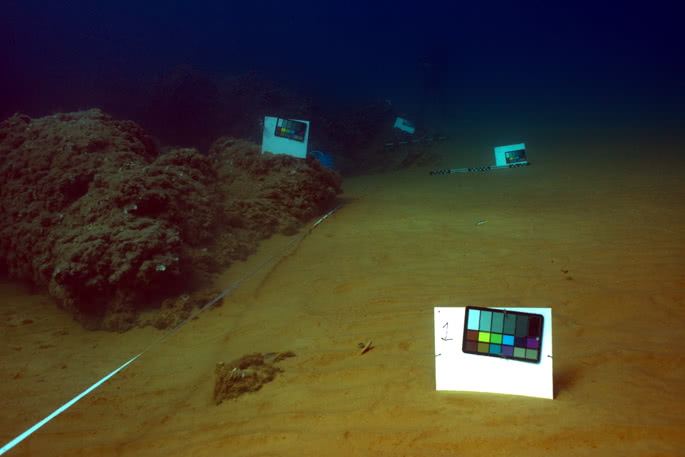} &
\includegraphics[width=0.22\linewidth]{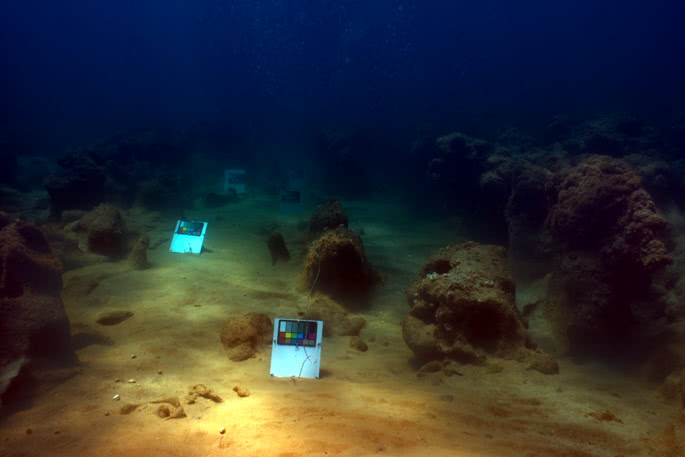} \\
\rotatebox{90}{\hspace{0.2cm} Peng \etal\cite{blurrinessICIP2015}} &
\includegraphics[width=0.22\linewidth]{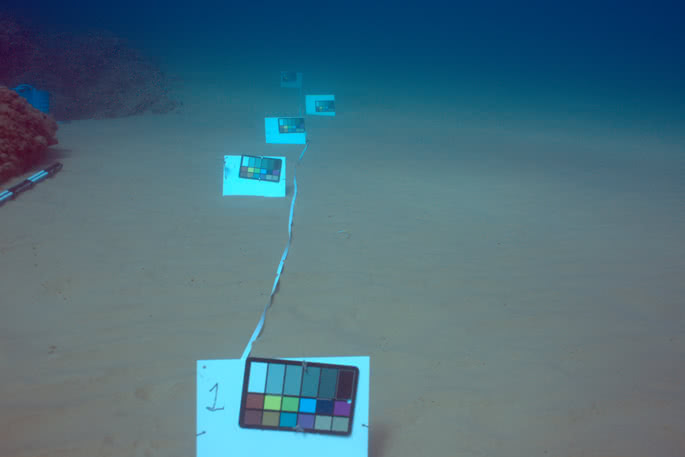} &
\includegraphics[width=0.22\linewidth]{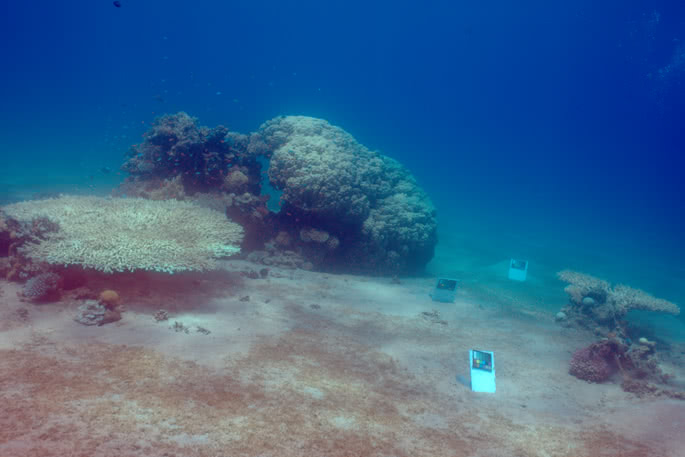} &
\includegraphics[width=0.22\linewidth]{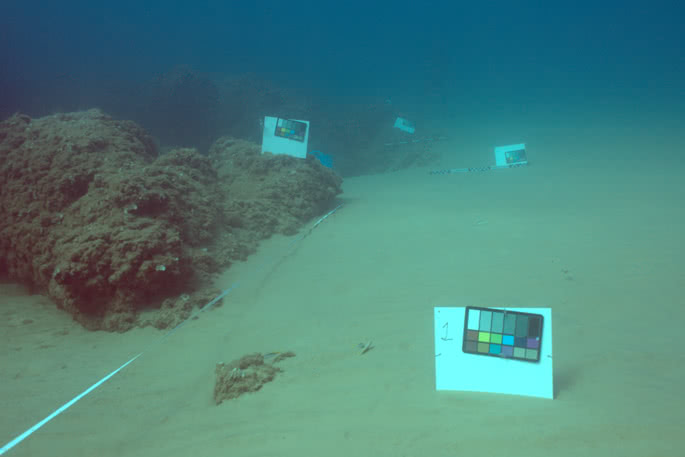} &
\includegraphics[width=0.22\linewidth]{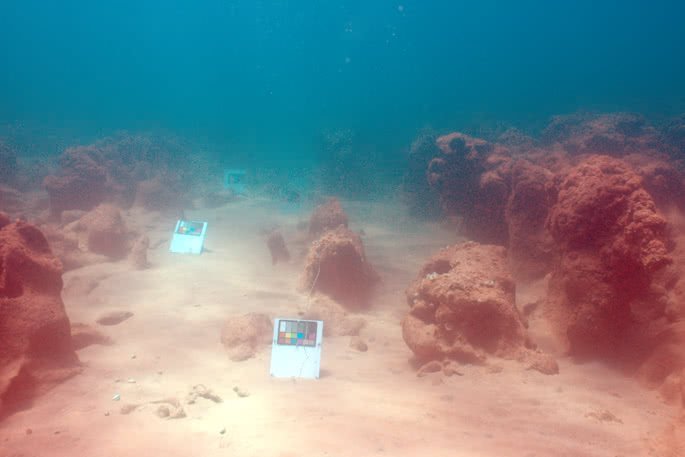} \\
\rotatebox{90}{\hspace{0.1cm} Ancuti \etal\cite{ancutiICPR2016}} &
\includegraphics[width=0.22\linewidth]{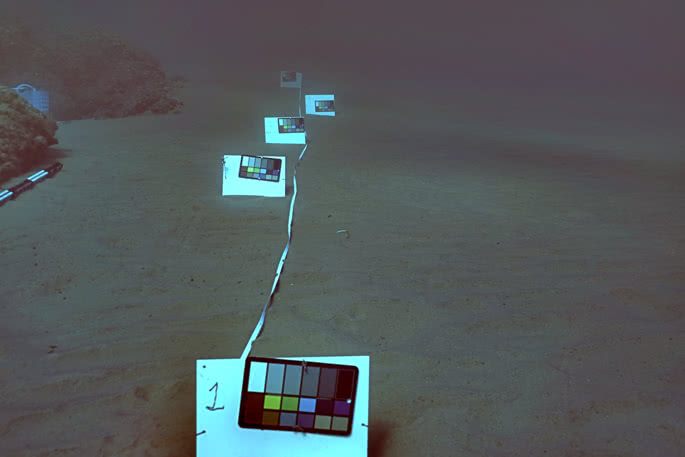} &
\includegraphics[width=0.22\linewidth]{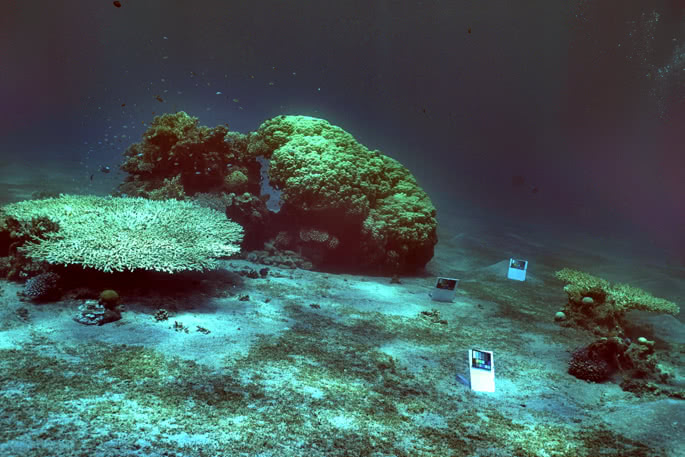} &
\includegraphics[width=0.22\linewidth]{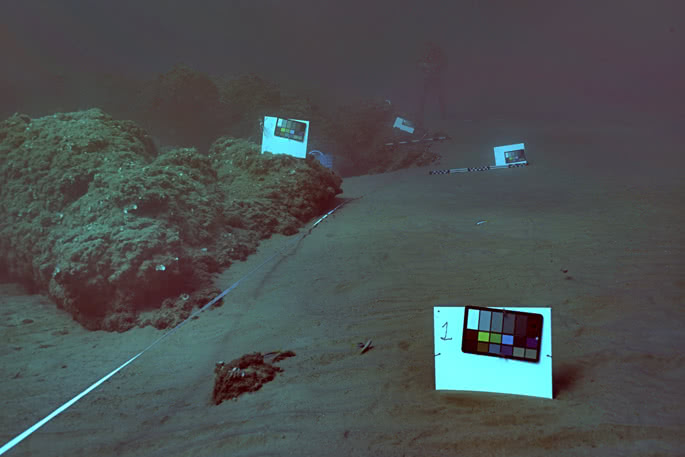} &
\includegraphics[width=0.22\linewidth]{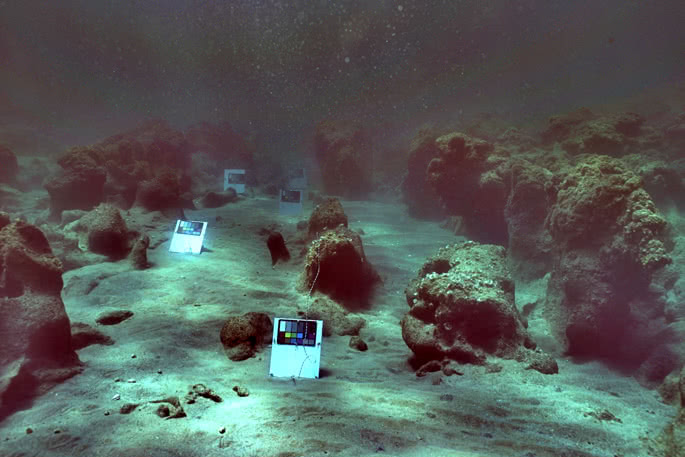} \\
\rotatebox{90}{\hspace{0.1cm} Ancuti \etal\cite{ancutiICIP2017}} &
\includegraphics[width=0.22\linewidth]{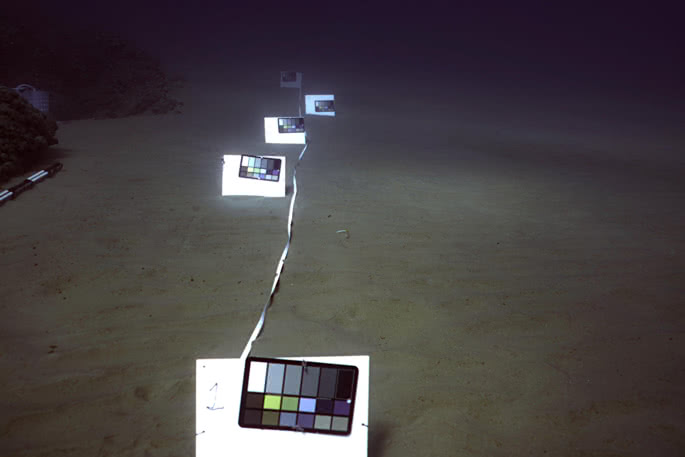} &
\includegraphics[width=0.22\linewidth]{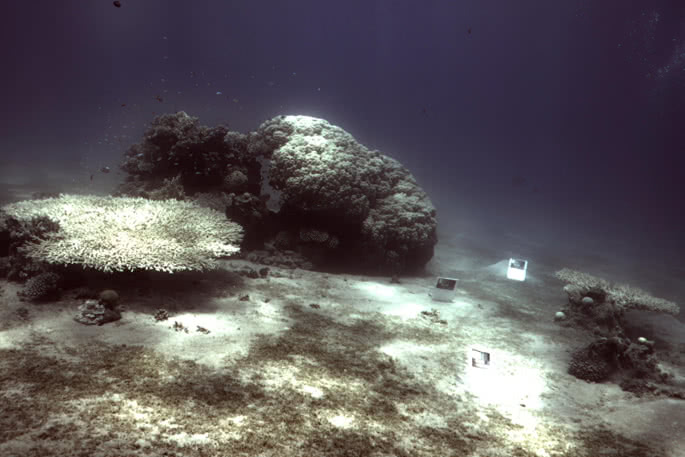} &
\includegraphics[width=0.22\linewidth]{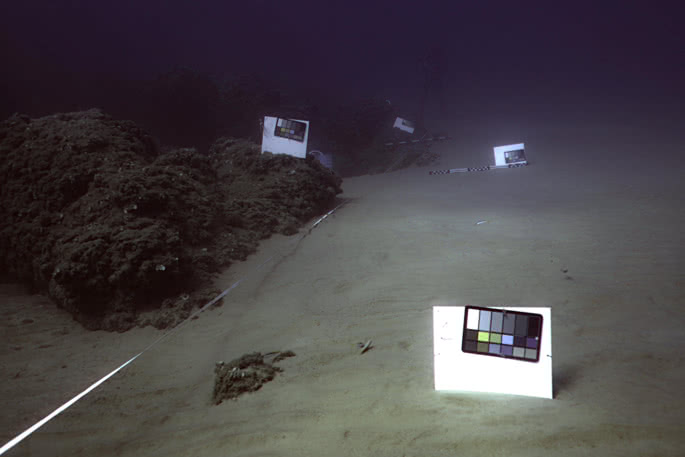} &
\includegraphics[width=0.22\linewidth]{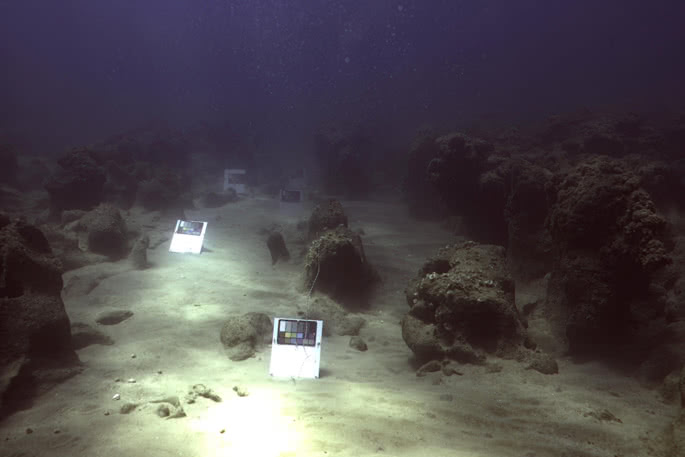} \\
\rotatebox{90}{\hspace{0.15cm}Ancuti \etal\cite{ancutiTIP2018}} &
\includegraphics[width=0.22\linewidth]{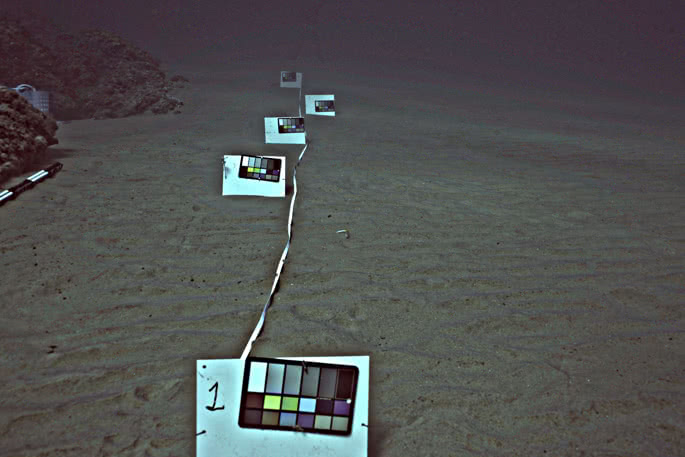} &
\includegraphics[width=0.22\linewidth]{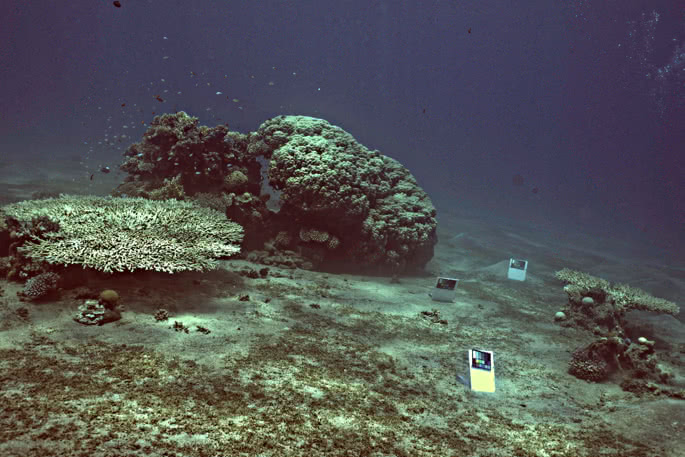} &
\includegraphics[width=0.22\linewidth]{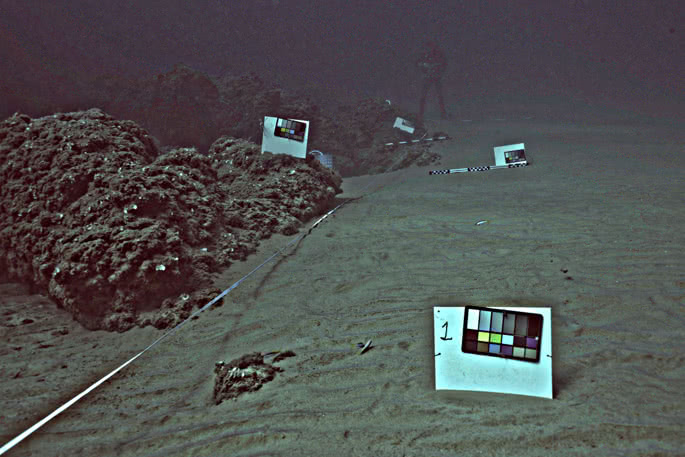} &
\includegraphics[width=0.22\linewidth]{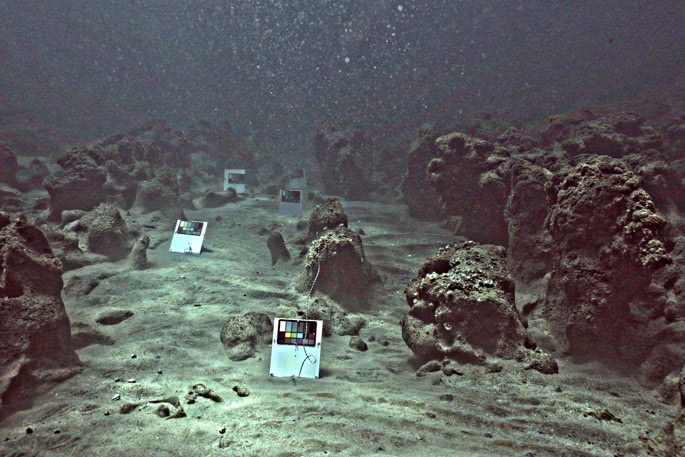} \\
\rotatebox{90}{\hspace{-0.1cm} Emberton \etal\cite{Emberton2017}} &
\includegraphics[width=0.22\linewidth]{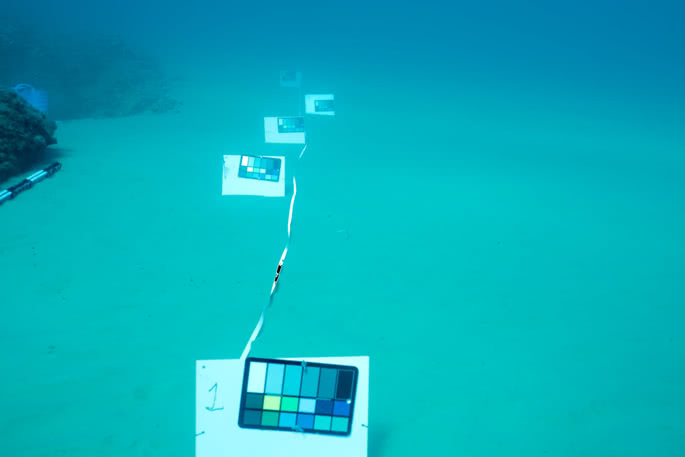} &
\includegraphics[width=0.22\linewidth]{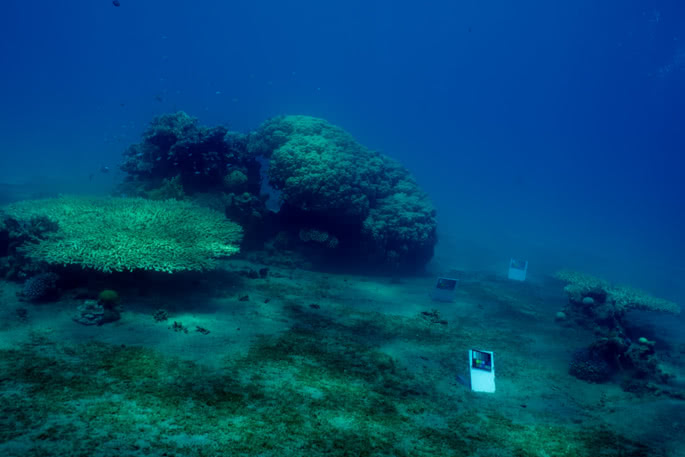} &
\includegraphics[width=0.22\linewidth]{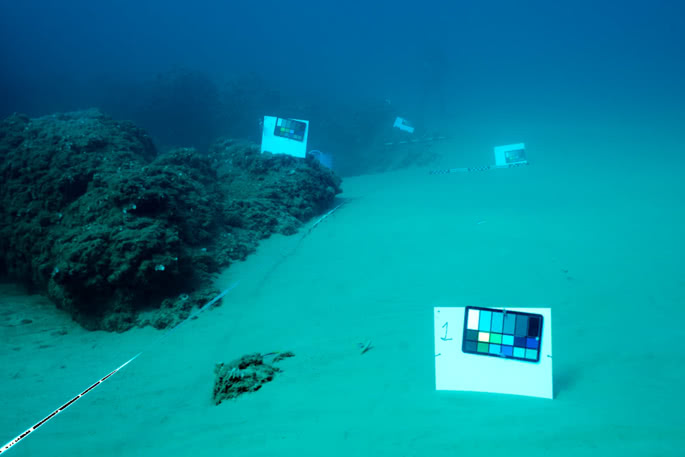} &
\includegraphics[width=0.22\linewidth]{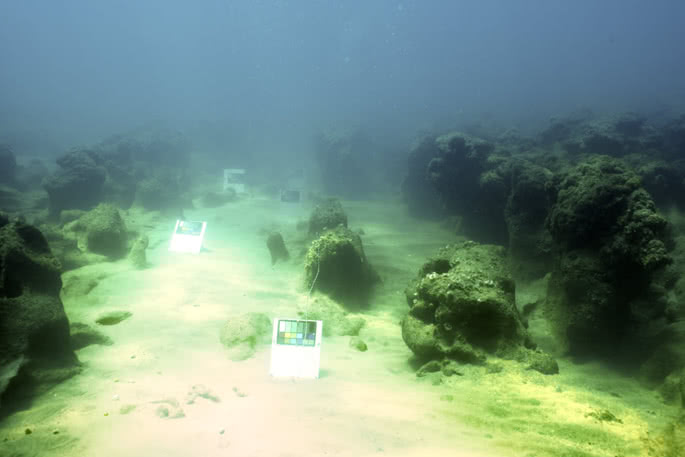} \\
\rotatebox{90}{\hspace{0.9cm} Ours} &
\includegraphics[width=0.22\linewidth]{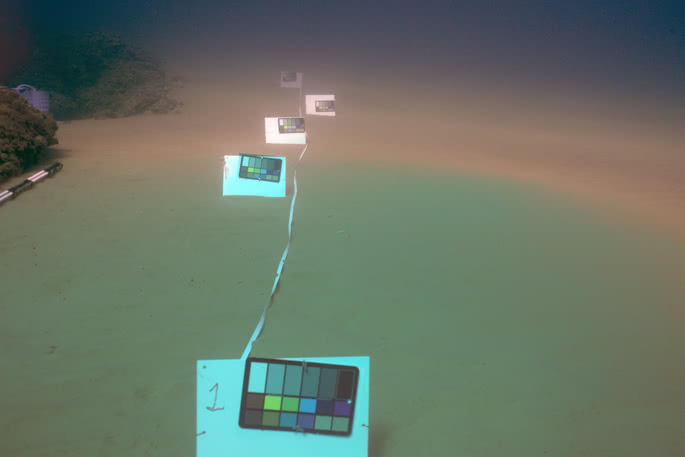} &
\includegraphics[width=0.22\linewidth]{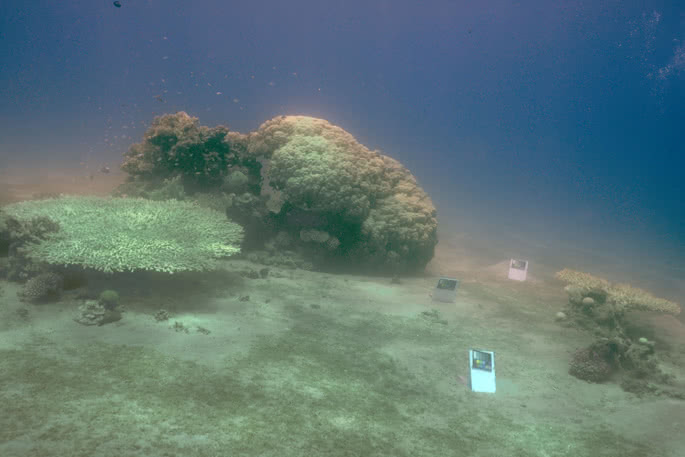} &
\includegraphics[width=0.22\linewidth]{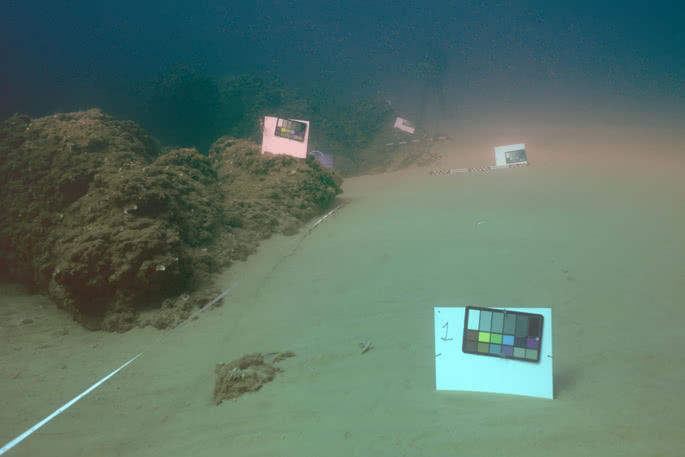} &
\includegraphics[width=0.22\linewidth]{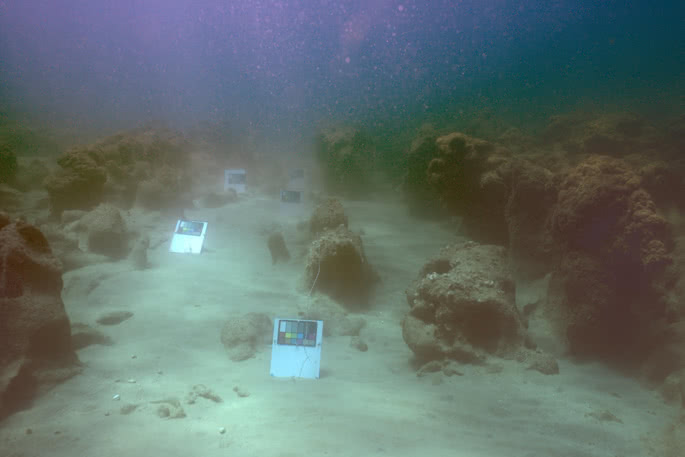} \\
\end{tabular}\vspace{-0.2cm}
\caption[Underwater image enhancement: comparison of single-image underwater color restoration algorithms (cont.)]{Comparison of single-image underwater enhancement techniques.}
\label{fig:OurDataset2}
\end{figure*}

\begin{figure*}[b]   
\centering \small\addtolength{\tabcolsep}{-4pt}
\begin{tabular}{ccccc}
\rotatebox{90}{\hspace{0.8cm} Input} &
\begin{tikzpicture}
\draw (0, 0) node[inner sep=0] {\includegraphics[width=0.23\linewidth]{RGT_5450_input_rgb.jpg}};
\draw (-0.108\linewidth, +1.18cm) node {\makebox[0pt][l]{\scriptsize\textbf{\color{white} R5450}}};
\end{tikzpicture} &
\begin{tikzpicture}
\draw (0, 0) node[inner sep=0] {\includegraphics[width=0.23\linewidth]{RGT_4485_input_rgb.jpg}};
\draw (-0.108\linewidth, +1.18cm) node {\makebox[0pt][l]{\scriptsize\textbf{\color{white} R4485}}};
\end{tikzpicture} &
\begin{tikzpicture}
\draw (0, 0) node[inner sep=0] {\includegraphics[width=0.23\linewidth]{RGT_5469_input_rgb.jpg}};
\draw (-0.108\linewidth, +1.18cm) node {\makebox[0pt][l]{\scriptsize\textbf{\color{white} R5469}}};
\end{tikzpicture} &
\begin{tikzpicture}
\draw (0, 0) node[inner sep=0] {\includegraphics[width=0.23\linewidth]{RGT_3204_input_rgb.jpg}};
\draw (-0.108\linewidth, +1.18cm) node {\makebox[0pt][l]{\scriptsize\textbf{\color{white} R3204}}};
\end{tikzpicture}  \vspace{-0.0cm} \\  \vspace{+0.0cm}

\rotatebox{90}{\hspace{0.6cm} Distance} &
\includegraphics[width=0.23\linewidth]{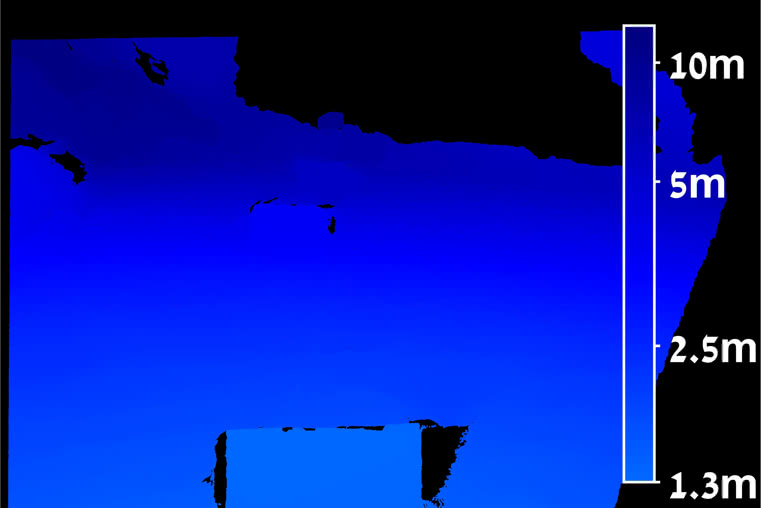} &
\includegraphics[width=0.23\linewidth]{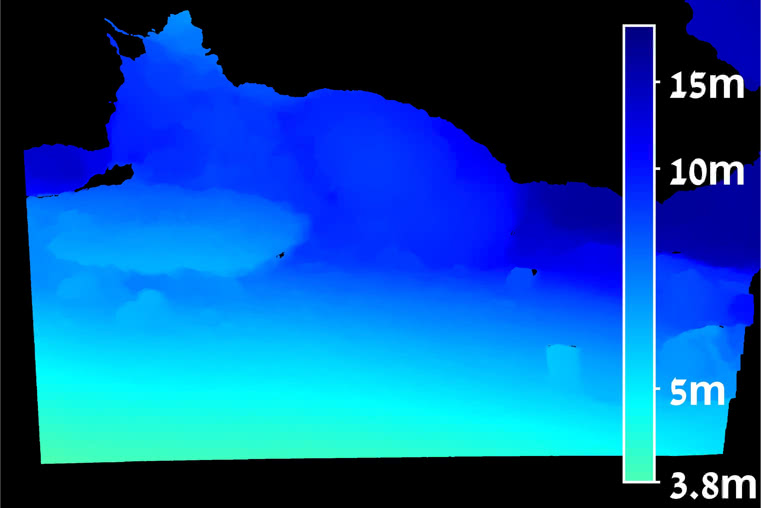} &
\includegraphics[width=0.23\linewidth]{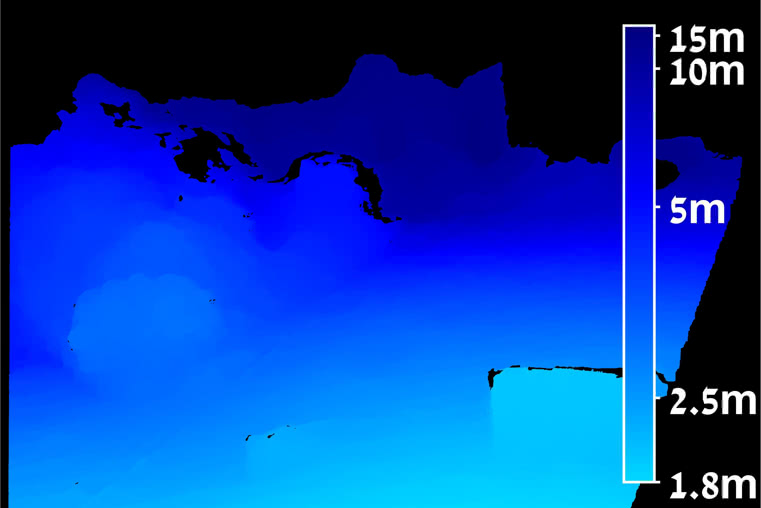} &
\includegraphics[width=0.23\linewidth]{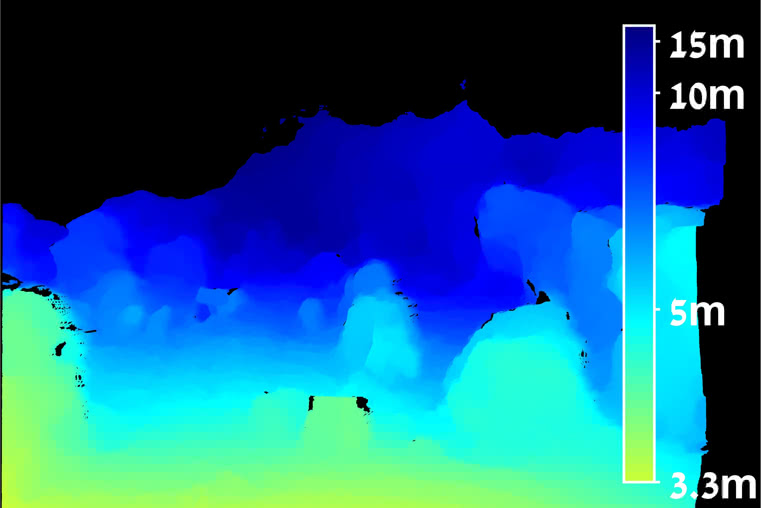} \vspace{-0.0cm} \\  \vspace{+0.0cm}
\rotatebox{90}{\hspace{+0.2cm} Drews \etal\cite{drews2013transmission}} &
\includegraphics[width=0.23\linewidth]{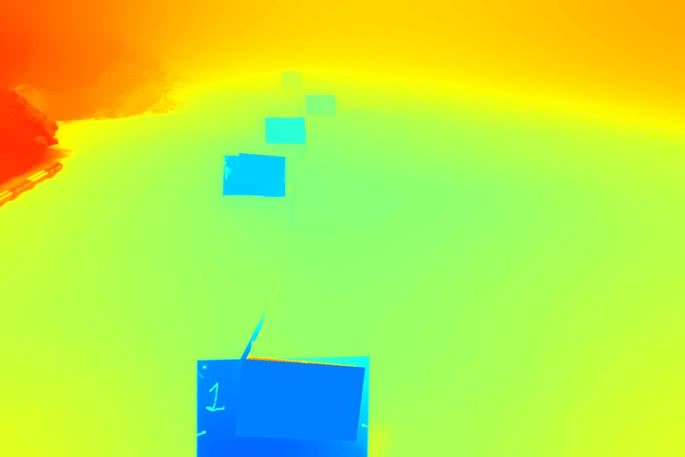} &
\includegraphics[width=0.23\linewidth]{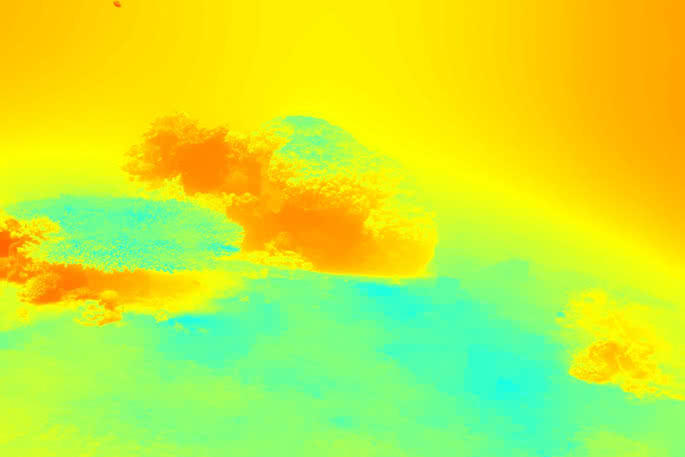} &
\includegraphics[width=0.23\linewidth]{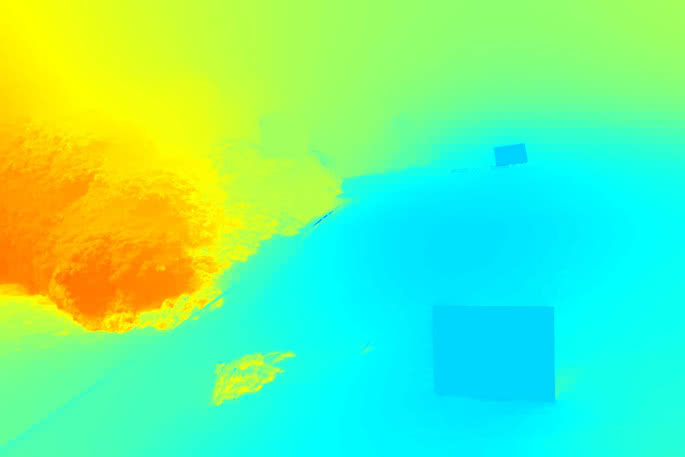} &
\includegraphics[width=0.23\linewidth]{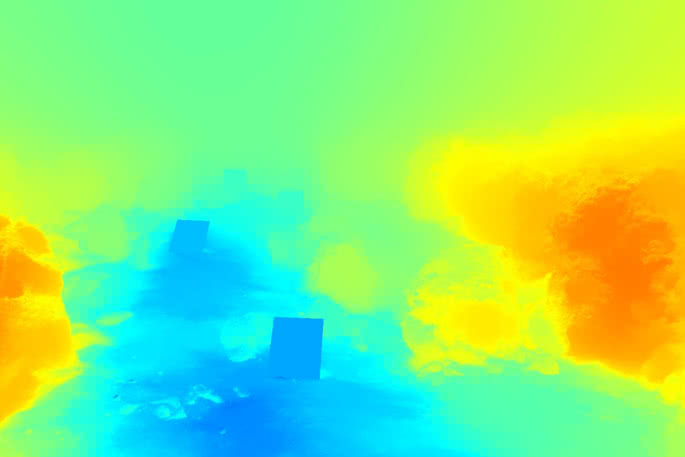} \vspace{-0.1cm} \\  \vspace{+0.1cm}
& $\rho = -0.40$  &  $\rho = -0.61$  &  $\rho = -0.41$  &  $\rho = -0.33$ \\ \vspace{+0.1cm}
\rotatebox{90}{\hspace{+0.3cm} Peng \etal\cite{blurrinessICIP2015}} &
\includegraphics[width=0.23\linewidth]{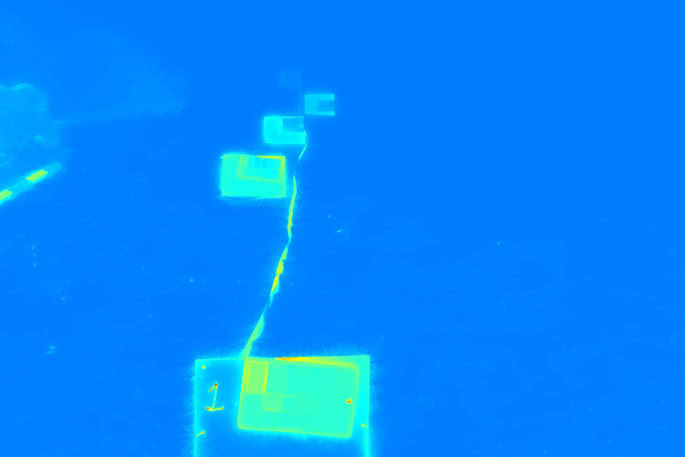} &
\includegraphics[width=0.23\linewidth]{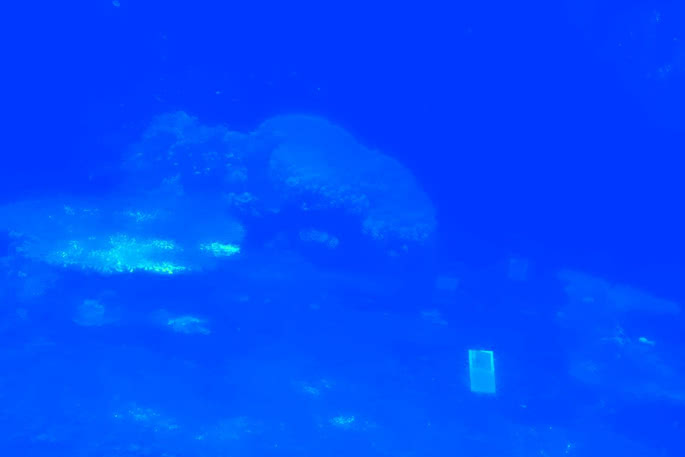} &
\includegraphics[width=0.23\linewidth]{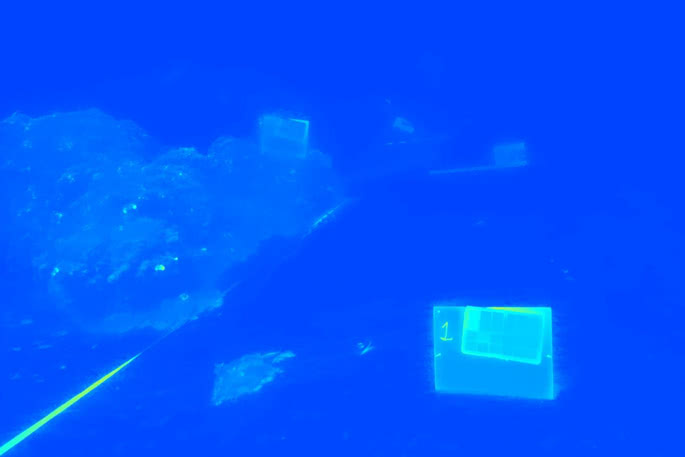} &
\includegraphics[width=0.23\linewidth]{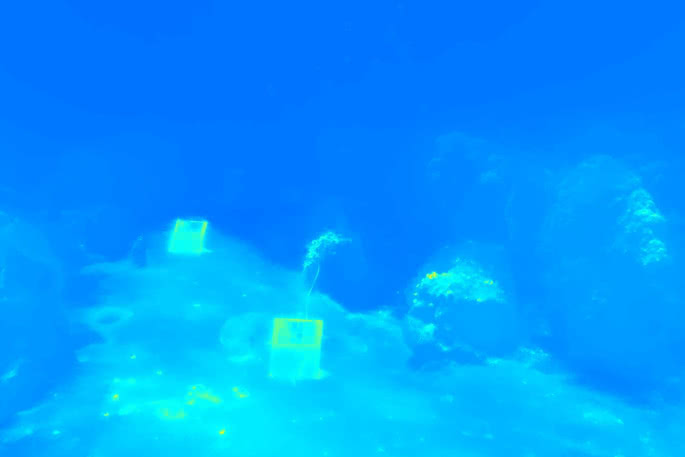} \vspace{-0.2cm} \\  \vspace{+0.1cm}
& $\rho = 0.15$  &  $\rho = 0.44$  &  $\rho = 0.13$  &  $\rho = 0.68$ \\
\rotatebox{90}{\hspace{0.1cm} Ancuti \etal\cite{ancutiICIP2017}} &
\includegraphics[width=0.23\linewidth]{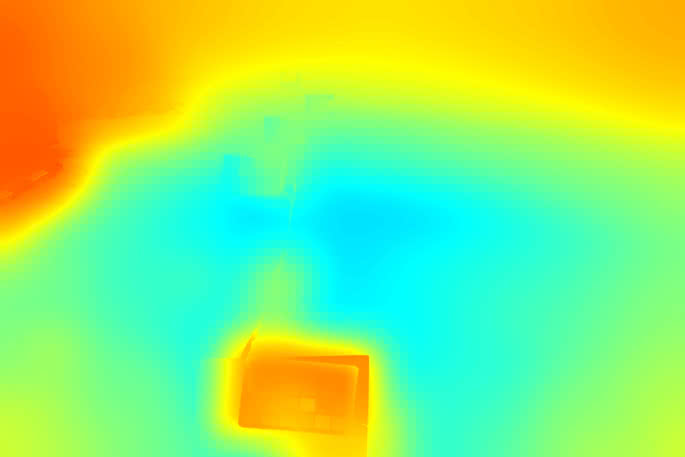} &
\includegraphics[width=0.23\linewidth]{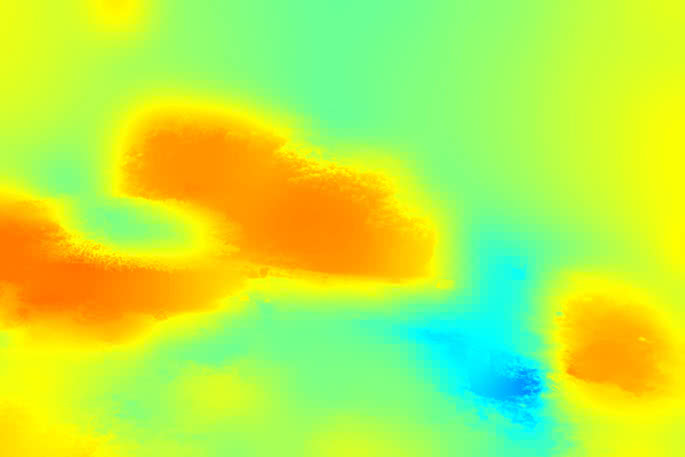} &
\includegraphics[width=0.23\linewidth]{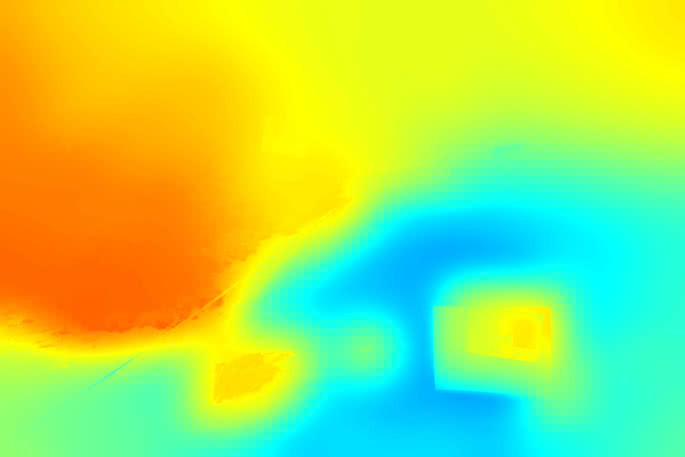} &
\includegraphics[width=0.23\linewidth]{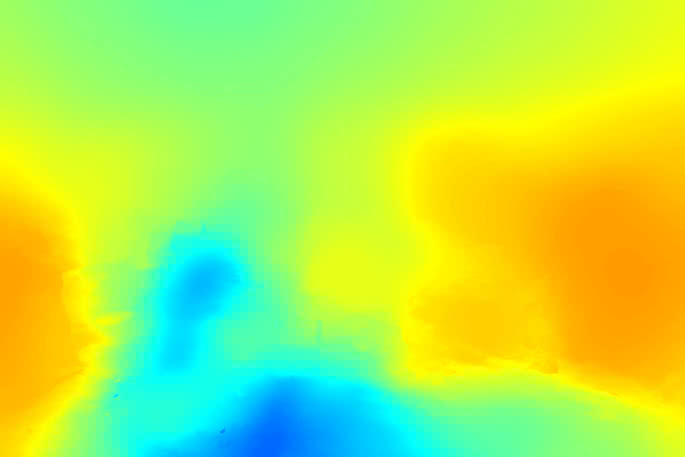}  \vspace{-0.1cm} \\  \vspace{+0.1cm}
& $\rho = -0.29$  &  $\rho = -0.12$  &  $\rho = -0.41$  &  $\rho = -0.28$ \\
\rotatebox{90}{\hspace{-0.705cm} Emberton \etal\cite{Emberton2017}} &
\includegraphics[width=0.23\linewidth]{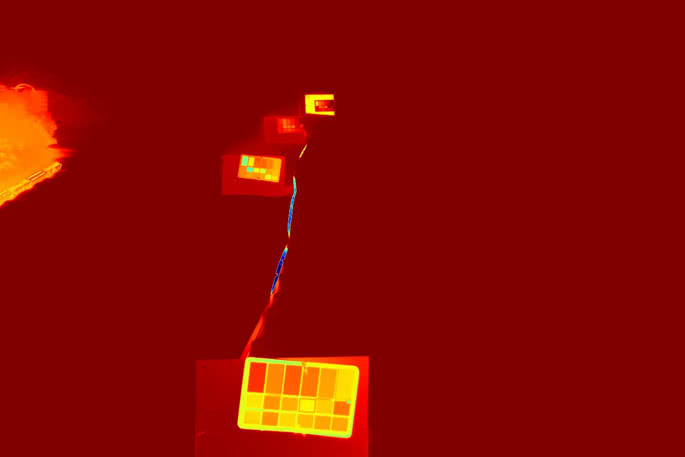} &
\includegraphics[width=0.23\linewidth]{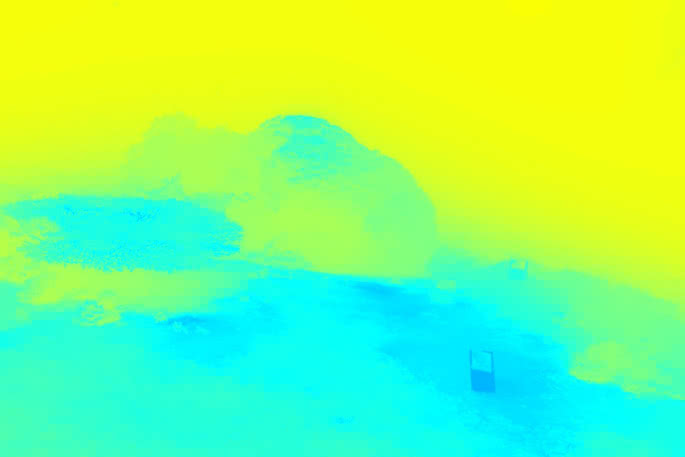} &
\includegraphics[width=0.23\linewidth]{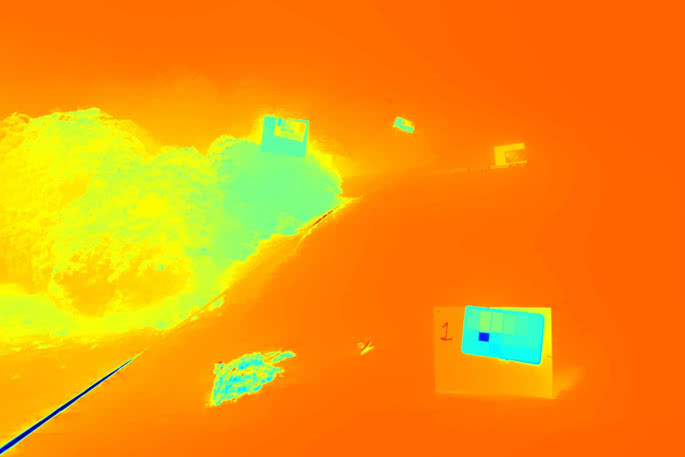} &
\includegraphics[width=0.23\linewidth]{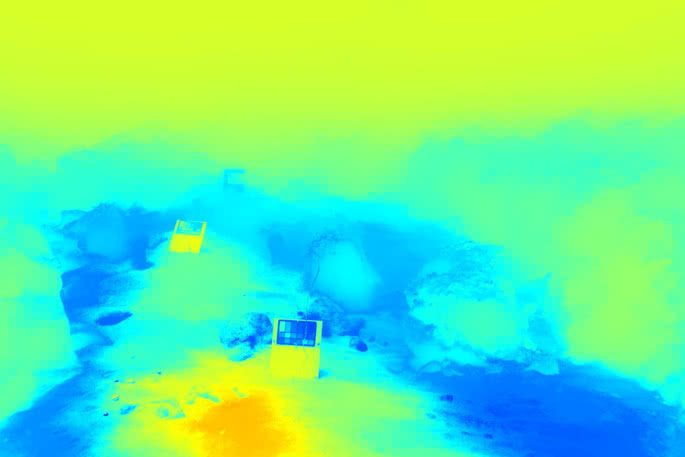} \vspace{-0.15cm} \\  \vspace{+0.1cm}
& $\rho = -0.06$  &  $\rho = -0.74$  &  $\rho = -0.07$  &  $\rho = -0.13$ \\
\rotatebox{90}{\hspace{0.9cm} Ours} &
\includegraphics[width=0.23\linewidth]{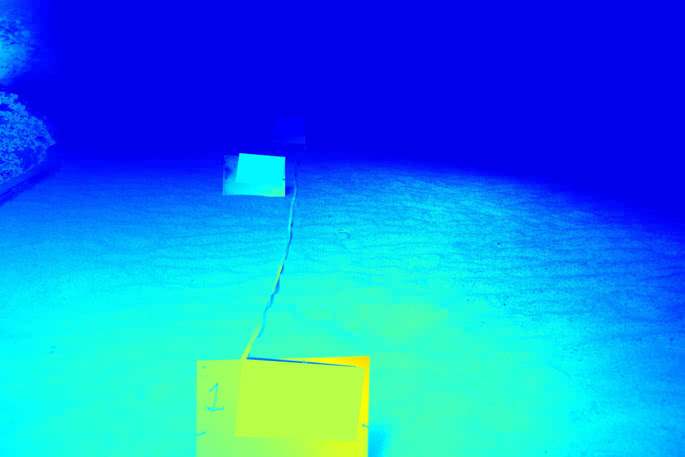} &
\includegraphics[width=0.23\linewidth]{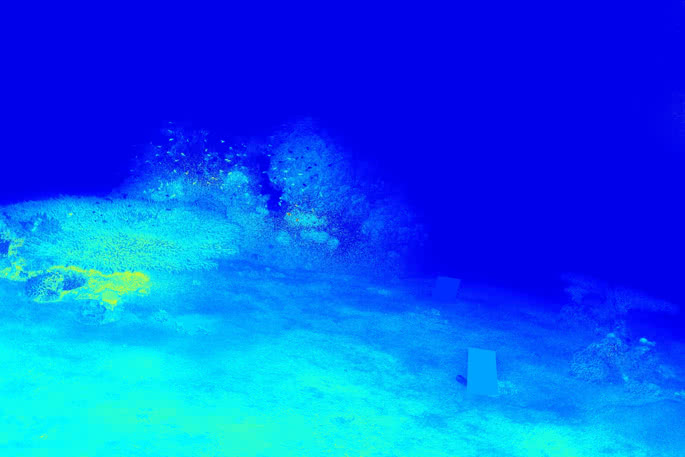} &
\includegraphics[width=0.23\linewidth]{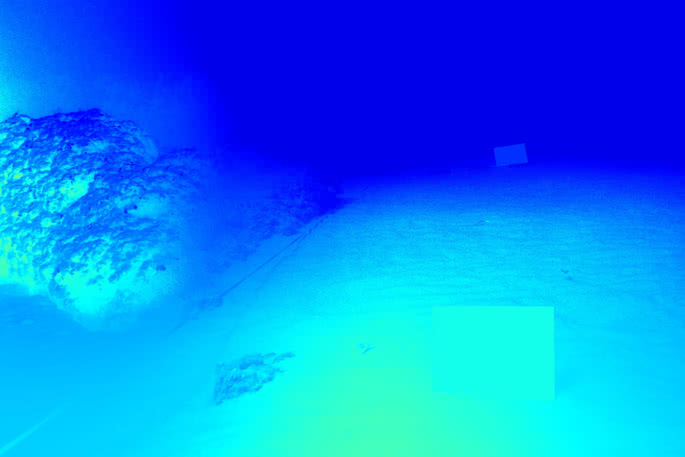} &
\includegraphics[width=0.23\linewidth]{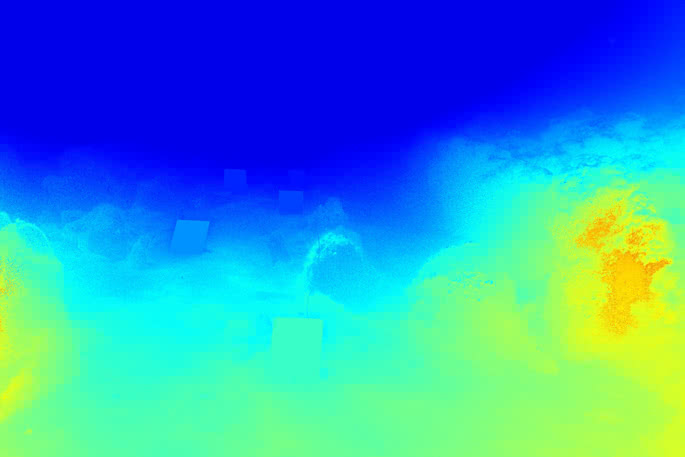} \vspace{-0.1cm} \\  \vspace{+0.1cm}
& $\rho = 0.87$  &  $\rho = 0.73$  &  $\rho = 0.90$  &  $\rho = 0.89$ \\
\end{tabular} \vspace{-0.1cm}
\caption[Underwater image enhancement: true distances and estimated transmission maps (cont.)]{True distances and estimated transmission maps for images shown in Fig.~\ref{fig:OurDataset2}, shown only for methods that estimate a transmission map. The quality of the transmission is measured by the Pearson correlation coefficient $\rho$, which is calculated between the negative logarithm of the estimated transmission and the true distance.}
\label{fig:OurDatasetTrans2}
\end{figure*}

Figs.~\ref{fig:OurDatasetTrans1} and~\ref{fig:OurDatasetTrans2} depict the ground-truth distances and the transmission maps for the images shown in Figs.~\ref{fig:OurDataset1} and~\ref{fig:OurDataset2}, respectively.
The true distance is shown in meters and is color-mapped, where black indicates missing distance values due to occlusions and unmatched pixels between the images. For example, the images shown here are from the right camera, and therefore many distance values are missing on the right border, since these regions were not visible in the left camera.
The transmission maps are shown only for the algorithms that estimate them for the color correction~\cite{ancutiICIP2017, drews2013transmission, Emberton2017, blurrinessICIP2015} and ours. These transmission maps are estimated from a single image and are evaluated quantitatively based on the true distance map of each scene.
The transmission is a function of both the distance $z$ and the unknown attenuation coefficient $\beta_c$: $t_c = e^{-\beta_cz}$. Therefore, we measure the correlation between the true distance $z_{GT}$ and the negative logarithm of the estimated transmission ($-log(t_c)=\beta_cz$). The Pearson correlation coefficient ${ \rho = \frac{\text{cov}(z_{GT}, -log(t))}{\sigma_{z_{GT}} \sigma_{log(t)}} }$ is noted below each transmission map, and shows that the proposed method estimates a much more accurate transmission. In addition, the data is summarized in Table~\ref{table:PearsonTrans}.
Note that the images were undistorted to correct the lens distortion during the stereo calculation, and this is taken into account in the calculation of $\rho$.

The correlation coefficient has a value between $-1$ and $+1$, where $+1$ is a perfect correlation, indicating a good transmission estimation, while lower values indicate that the 3D structure of the scene was not estimated correctly. A negative value implies either a wrong estimation of the veiling light or that the used prior is not valid for that scene.

\clearpage

\clearpage

\begin{table*}[!th]\footnotesize \addtolength{\tabcolsep}{-1pt}
\centering
\caption[Quantitative evaluation of transmission accuracy by underwater image enhancement methods]{Each column shows the Pearson correlation coefficient $\rho$ between the estimated transmission maps and the ground truth distance for a different image, as shown in Figs.~\ref{fig:OurDatasetTrans1},~\ref{fig:OurDatasetTrans2}, and~\ref{fig:WrongWaterType}. Since the distance is calculated in meters, we calculate the correlation between the distance and $-log(t)$.\vspace{-0.1cm}}
\label{table:PearsonTrans}
\scalebox{1.0}{
\begin{tabular}{|l|c|c|c|c|c|c|c|c|c|}
\hline
Image                                   & R3008     & R4376    & R5478    & R4491    & R5450    & R4485    & R5469    & R3204    & R3158 \\
\hline
Drews \etal\cite{drews2013transmission} & -0.28     & -0.71    & -0.40    & -0.45    & -0.46    & -0.59    & -0.43    & -0.30    & -0.26 \\
Peng \etal\cite{blurrinessICIP2015}     & 0.72      &  0.43    &  0.16    &  0.06    &  0.15    &  0.44    &  0.13    &  0.68    & 0.63 \\
Ancuti \etal\cite{ancutiICIP2017}       &  0.01     &  0.12    &  0.13    &  0.06    &  0.39    &  0.06    &  0.04    &  0.11    & 0.20\\
Emberton \etal\cite{Emberton2017}       & -0.04     & -0.87    &  0.11    & -0.54    & -0.04    & -0.74    & -0.05    & -0.12    & 0.06\\
Ours                                    &\tbf{0.93} &\tbf{0.75}&\tbf{0.66}&\tbf{0.80}&\tbf{0.87}&\tbf{0.73}&\tbf{0.90}&\tbf{0.89}&\tbf{0.81} \\
\hline
\end{tabular}
}\vspace{-0.3cm}
\end{table*}

\begin{figure*}[!hb]   
\centering \small\addtolength{\tabcolsep}{-3pt}
\begin{tabular}{cc}
\begin{tikzpicture}
\draw (0, 0) node[inner sep=0] {\includegraphics[height=3.8cm]{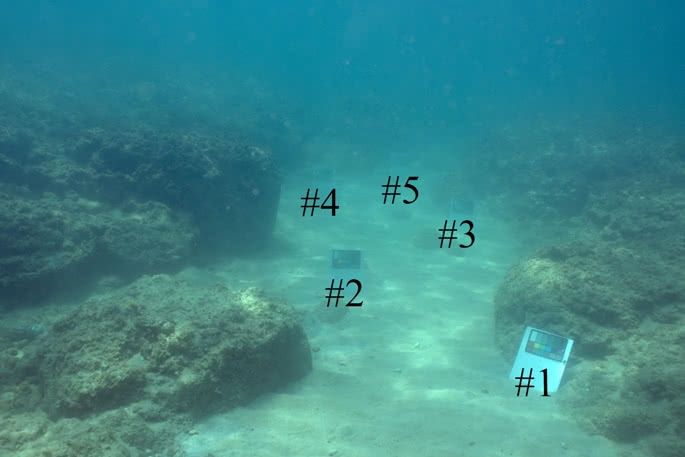}};
\draw (-2.7cm, +1.68cm) node {\makebox[0pt][l]{\scriptsize\textbf{\color{white} R3008}}};
\end{tikzpicture}  &
\includegraphics[height=3.8cm]{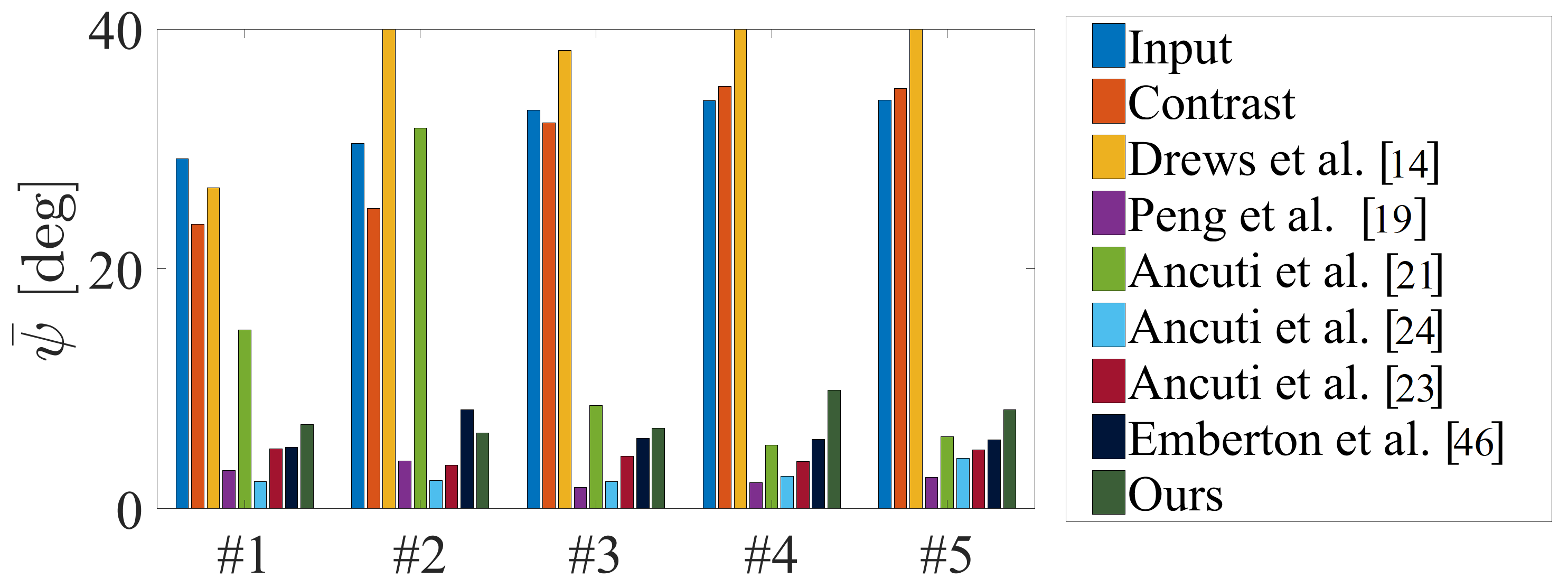} \\
\begin{tikzpicture}
\draw (0, 0) node[inner sep=0] {\includegraphics[height=3.8cm]{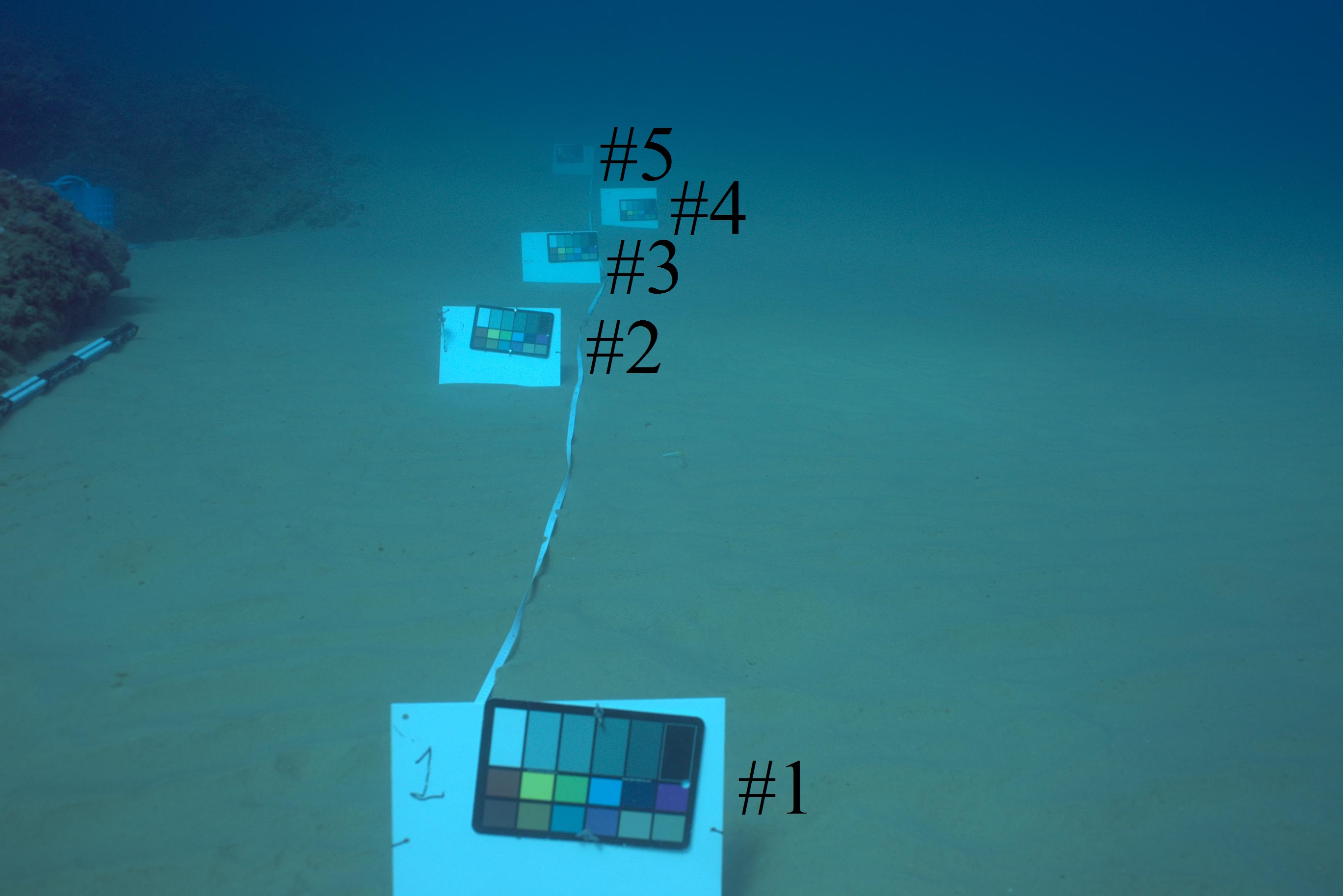}};
\draw (-2.7cm, +1.68cm) node {\makebox[0pt][l]{\scriptsize\textbf{\color{white} R5450}}};
\end{tikzpicture}  &
\includegraphics[height=3.8cm]{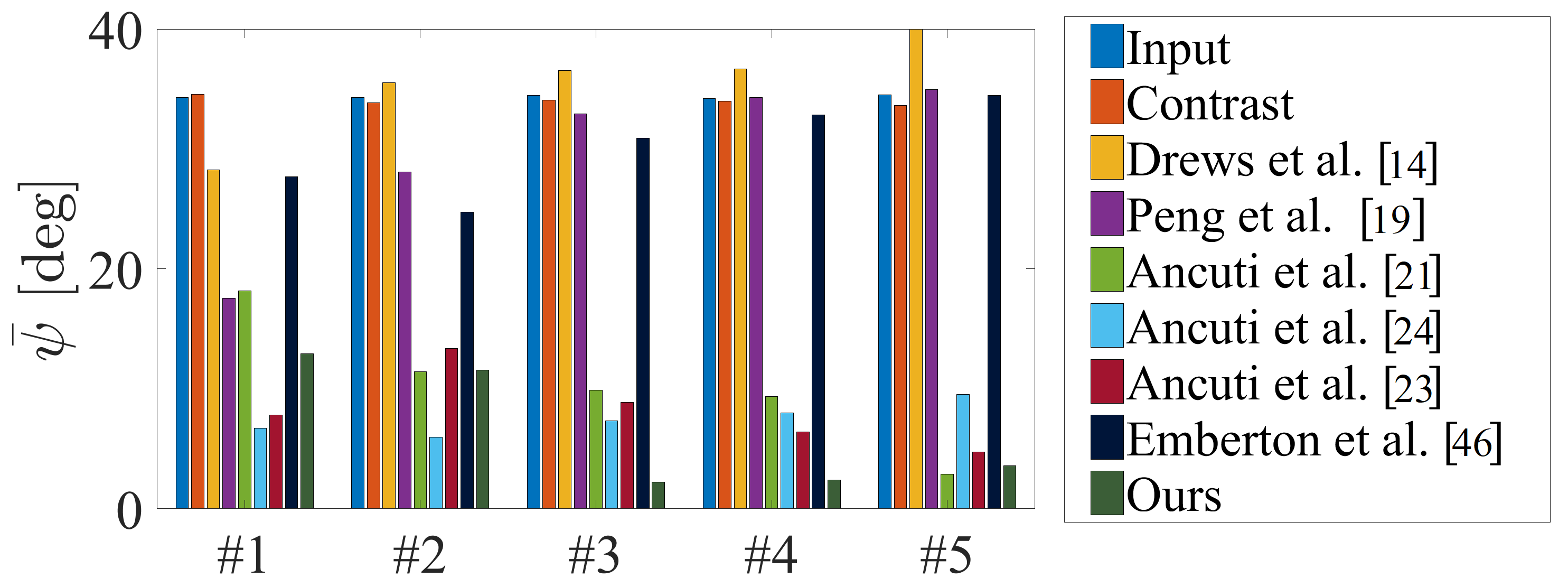} \\
\end{tabular} \vspace{-0.1cm}
\caption[Underwater image enhancement: quantitative evaluation of color reproduction using color charts]{ Average angular reproduction error (in degrees) of gray-scale patches, measured for different color restoration techniques (shown in Figs.~\ref{fig:OurDataset1} and \ref{fig:OurDataset2}). The color charts are numbered in increasing order, starting with the closest to the camera, and are annotated on the images on the left column. The right column shows the average angular reproduction error $\bar{\psi}$ for each chart and each method. Lower is better.\vspace{-0.2cm}}
\label{fig:OurDatasetColorEval}
\end{figure*}

Some of the methods, \eg Emberton \etal\cite{Emberton2017}, do not aim for a physically valid transmission, as they set its value in veiling-light regions to be intentionally high in order to suppress noise enhancement. However, we still expect to see a correlation between the true distances and the transmission, since in veiling light regions the true distance are often missing. This is a caused by the lack of texture in those regions, making it difficult to find matches between the left and right images and therefore impossible to triangulate.
As Table~\ref{table:PearsonTrans} shows, our method outperforms other algorithms in terms of transmission accuracy.

\subsection{Color Restoration - Quantitative Evaluation}

All of the scenes in Figs.~\ref{fig:OurDataset1} and~\ref{fig:OurDataset2} contain color charts at different distances from the camera.
Fig.~\ref{fig:OurDatasetColorEval} shows a quantitative evaluation of the color restoration accuracy for two different images out of the entire set, while Table~\ref{table:ColorChartsErrors} lists the values for all of the images. The input images are shown on the left of Fig.~\ref{fig:OurDatasetColorEval}, with the color charts marked. On the right, the bars depict the average reproduction angular error $\bar{\psi}$ (unit: degrees)~\cite{Finlayson2014ReproductionAngularError} in RGB space between the gray-scale patches and a pure gray color, for each chart, image, and method. For a given chart, the average angular error is defined as:
\begin{equation}
\bar{\psi} = \frac{1}{6} \cdot \sum_{\vect{x}_i} \cos ^{-1} \left( \frac{\vect{I}(\vect{x}) \cdot \left(1, 1, 1 \right)}{  \lVert \vect{I}(\vect{x}) \rVert \cdot \sqrt{3} } \right) ~ ~,
\end{equation}
where $\vect{x_i}$ mark the coordinates of the grayscale patches in the image plane (there are $6$ of those).
Lower angles indicate a more accurate color restoration. By calculating the angle we eliminate the influence of the global illumination on the brightness of each color patch and are able to calculate a robust measure on the neutral patches.
The charts' order is by increasing distance from the camera, \ie, chart $\#1$ is closest to the camera.
By comparing the input image to a na\"{\i}ve contrast stretch (the two leftmost bars for each chart), we see that a global operation cannot compensate for the distance-dependent degradation: while the contrast stretch angle (orange bar) is often lower than the input angle (the leftmost blue bar) for the closest chart ($\#1$), this difference often shrinks for farther charts, demonstrating that a distance-dependent correction is required. The top methods for color restoration are ours and \cite{ancutiICIP2017}, as evident from Table~\ref{table:ColorChartsErrors}.

\begin{table}[tb]\addtolength{\tabcolsep}{-3.5pt}
\centering
\caption[Quantitative evaluation of color restoration by underwater image enhancement methods]{The average reproduction angular error $\bar{\psi}$ in RGB space between the gray-scale patches and a pure gray color, for all each chart in each image, and all methods, including the input and global contrast stretch (labeled Cont.). Lower is better.\vspace{-0.2cm}}
\label{table:ColorChartsErrors}
\footnotesize
\begin{tabular}{|c|c|c|c|c|c|c|c|c|c|}
\hline
Method & Input & Cont. & \cite{drews2013transmission} & \cite{blurrinessICIP2015} & \cite{ancutiICPR2016} &  \cite{ancutiICIP2017} &  \cite{ancutiTIP2018} & \cite{Emberton2017} & Ours \\
\hline
R3008 \#1 & 29.18 & 23.73 & 26.74 & 3.19 & 14.89 & 2.25 & 5.00 & 5.14 & 7.05 \\
 R3008 \#2 & 30.48 & 25.06 & 43.74 & 3.98 & 31.75 & 2.35 & 3.63 & 8.25 & 6.32 \\
 R3008 \#3 & 33.24 & 32.18 & 38.20 & 1.80 & 8.62 & 2.27 & 4.39 & 5.88 & 6.71 \\
 R3008 \#4 & 34.03 & 35.23 & 40.44 & 2.18 & 5.30 & 2.71 & 3.96 & 5.80 & 9.88 \\
 R3008 \#5 & 34.08 & 35.05 & 40.05 & 2.62 & 6.01 & 4.21 & 4.91 & 5.76 & 8.27 \\
 R3204 \#1 & 27.43 & 19.27 & 12.81 & 1.73 & 23.14 & 2.79 & 13.00 & 6.64 & 7.58 \\
 R3204 \#2 & 34.10 & 34.32 & 32.03 & 7.95 & 13.07 & 3.04 & 4.24 & 5.08 & 7.91 \\
 R3204 \#3 & 34.11 & 34.76 & 35.92 & 17.36 & 6.05 & 4.91 & 6.14 & 7.20 & 9.82 \\
 R3204 \#4 & 34.21 & 34.71 & 35.66 & 24.66 & 5.76 & 4.76 & 3.21 & 7.29 & 9.12 \\
 R3204 \#5 & 34.21 & 34.84 & 37.30 & 28.01 & 4.17 & 5.83 & 4.88 & 7.62 & 13.94 \\
 R4376 \#1 & 35.17 & 34.80 & 37.00 & 30.29 & 31.19 & 6.58 & 17.05 & 39.10 & 25.28 \\
 R4376 \#2 & 35.21 & 34.64 & 40.27 & 29.65 & 33.69 & 6.85 & 19.87 & 37.73 & 23.80 \\
 R4376 \#3 & 35.13 & 34.70 & 40.86 & 31.26 & 18.61 & 5.55 & 16.62 & 37.04 & 23.10 \\
 R4376 \#4 & 35.96 & 39.01 & 39.33 & 28.61 & 5.22 & 11.25 & 8.47 & 38.07 & 14.04 \\
 R4485 \#1 & 34.37 & 34.76 & 40.39 & 22.08 & 15.69 & 1.74 & 13.89 & 36.46 & 6.36 \\
 R4485 \#2 & 34.34 & 35.30 & 46.21 & 33.87 & 15.38 & 4.18 & 9.02 & 37.91 & 3.48 \\
 R4485 \#3 & 34.54 & 36.32 & 42.41 & 34.71 & 12.43 & 3.12 & 5.91 & 35.91 & 1.72 \\
 R4491 \#1 & 34.37 & 35.19 & 41.76 & 7.75 & 21.85 & 6.30 & 13.45 & 33.42 & 10.14 \\
 R4491 \#2 & 34.24 & 34.68 & 45.78 & 6.64 & 10.98 & 5.70 & 10.59 & 34.26 & 7.51 \\
 R4491 \#3 & 34.29 & 36.14 & 44.35 & 33.92 & 12.02 & 9.11 & 7.54 & 34.11 & 4.39 \\
 R5450 \#1 & 34.31 & 34.56 & 28.25 & 17.55 & 18.18 & 6.73 & 7.81 & 27.69 & 12.93 \\
 R5450 \#2 & 34.29 & 33.86 & 35.53 & 28.09 & 11.44 & 5.97 & 13.37 & 24.74 & 11.58 \\
 R5450 \#3 & 34.45 & 34.06 & 36.56 & 32.92 & 9.89 & 7.33 & 8.90 & 30.89 & 2.23 \\
 R5450 \#4 & 34.21 & 33.99 & 36.66 & 34.31 & 9.37 & 7.98 & 6.42 & 32.85 & 2.38 \\
 R5450 \#5 & 34.50 & 33.65 & 40.29 & 34.96 & 2.87 & 9.54 & 4.72 & 34.48 & 3.59 \\
 R5469 \#1 & 34.28 & 34.09 & 20.11 & 15.85 & 20.50 & 4.92 & 10.17 & 30.29 & 7.01 \\
 R5469 \#2 & 34.36 & 34.39 & 20.67 & 19.52 & 9.45 & 6.26 & 13.82 & 34.82 & 5.20 \\
 R5469 \#3 & 34.38 & 33.98 & 27.36 & 23.26 & 10.95 & 7.38 & 7.47 & 34.38 & 1.57 \\
 R5469 \#4 & 34.42 & 34.15 & 29.80 & 24.99 & 5.93 & 7.05 & 3.09 & 34.75 & 2.63 \\
 R5478 \#1 & 34.34 & 34.07 & 23.51 & 33.74 & 25.20 & 4.80 & 12.11 & 31.35 & 4.20 \\
 R5478 \#2 & 34.42 & 33.95 & 27.30 & 34.25 & 15.67 & 6.28 & 5.53 & 35.28 & 3.63 \\
 R5478 \#3 & 34.50 & 34.23 & 32.78 & 34.43 & 12.01 & 9.22 & 5.27 & 35.49 & 2.83 \\
 \hline
 Average & 33.91 & 33.55 & 35.00 & 21.44 & 13.98 & \tbf{5.59} & 8.58 & 25.49 & \tbf{\textit{8.32}} \\
 \hline
\end{tabular}
\end{table}


\vspace{+0.8cm}
\subsection{Water type effect}\vspace{-0.1cm}

\begin{figure}[tb]   
\centering \small\addtolength{\tabcolsep}{-5pt}
\begin{tabular}{ccc}
\begin{tikzpicture}
\draw (0, 0) node[inner sep=0] {\includegraphics[width=0.32\linewidth]{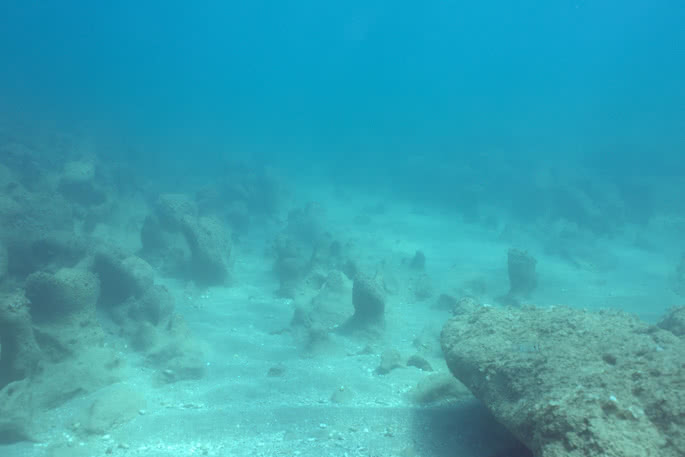}};
\draw (-0.15\linewidth, +0.8cm) node {\makebox[0pt][l]{\scriptsize\textbf{\color{white} R3158}}};
\end{tikzpicture} &
\includegraphics[width=0.32\linewidth]{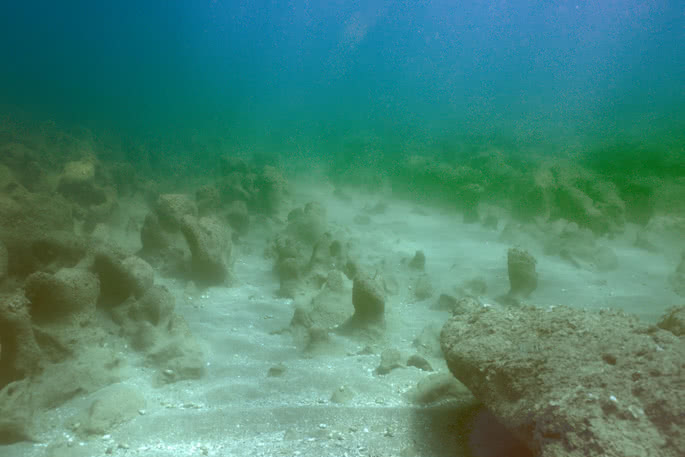} &
\includegraphics[width=0.32\linewidth]{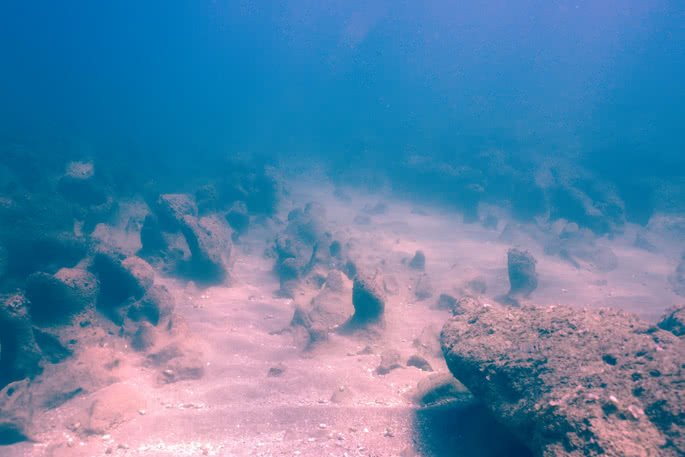} \\
Input Image & Our Result & Water Type I \\
\includegraphics[width=0.32\linewidth]{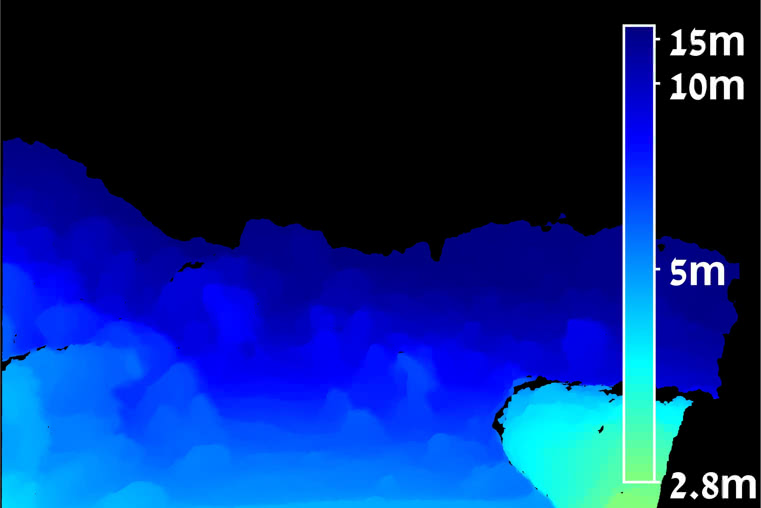} &
\includegraphics[width=0.32\linewidth]{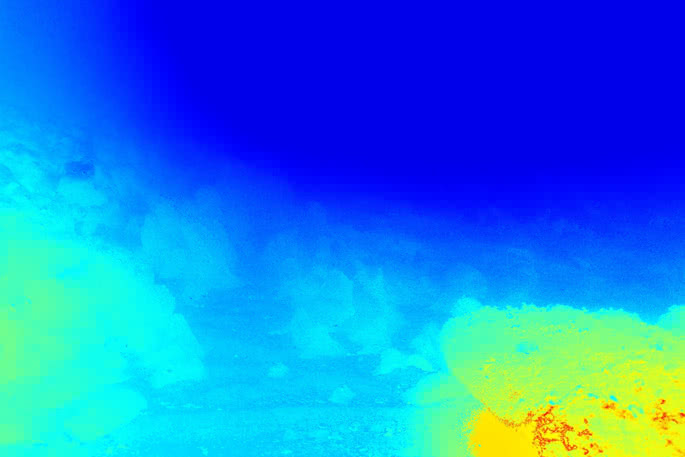} &
\includegraphics[width=0.32\linewidth]{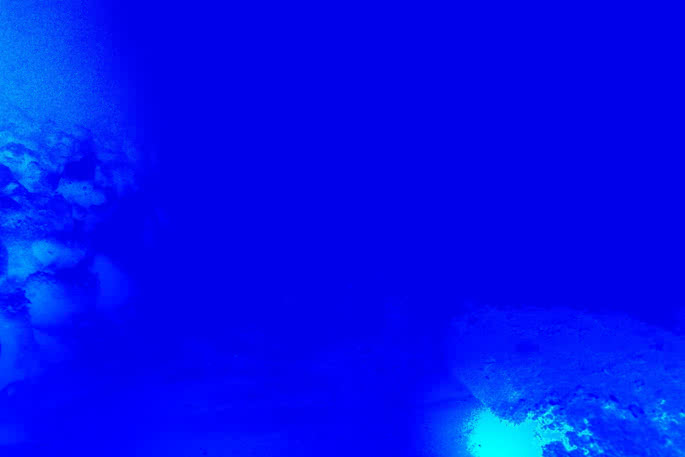} \\
True Distances & $\rho = 0.81$ & $\rho = 0.41$ \vspace{0.2cm} \\

\includegraphics[width=0.32\linewidth]{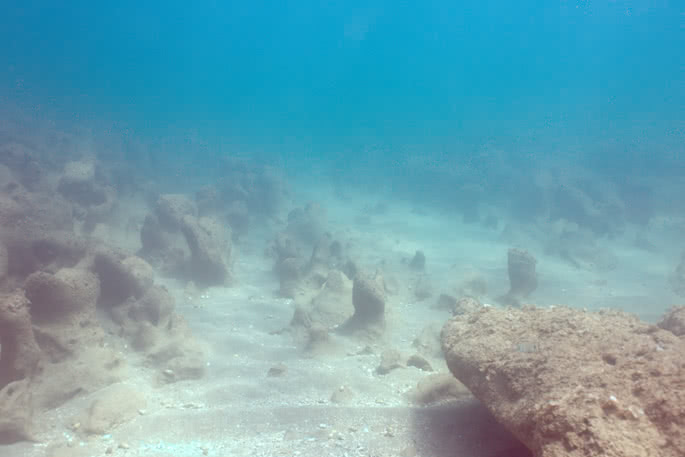} &
\includegraphics[width=0.32\linewidth]{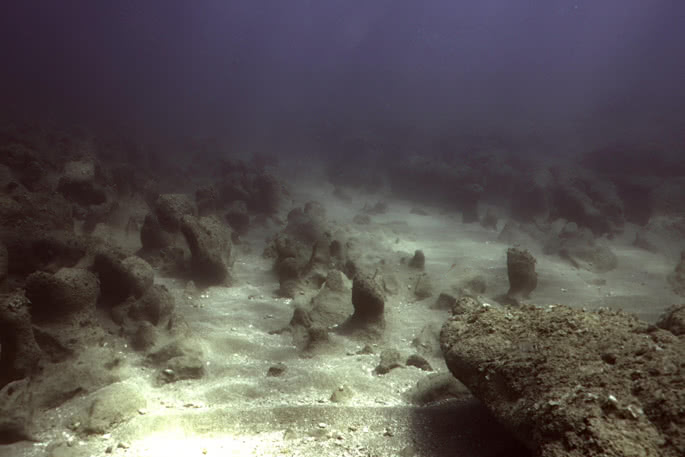} &
\includegraphics[width=0.32\linewidth]{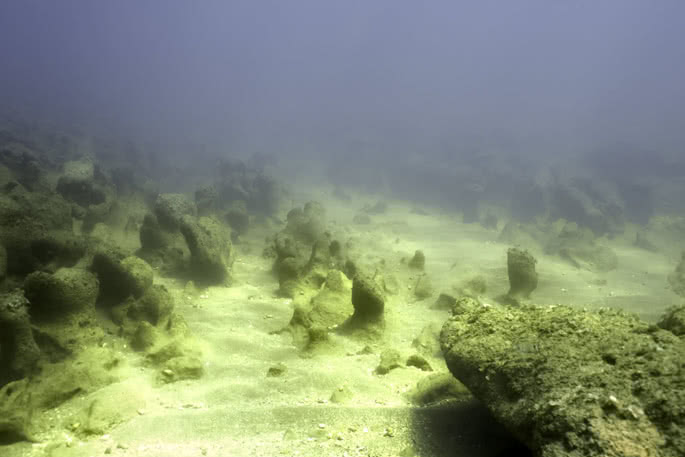} \\
Peng \etal\cite{blurrinessICIP2015} & Ancuti \etal\cite{ancutiICIP2017} & Emberton \etal\cite{Emberton2017} \\
\includegraphics[width=0.32\linewidth]{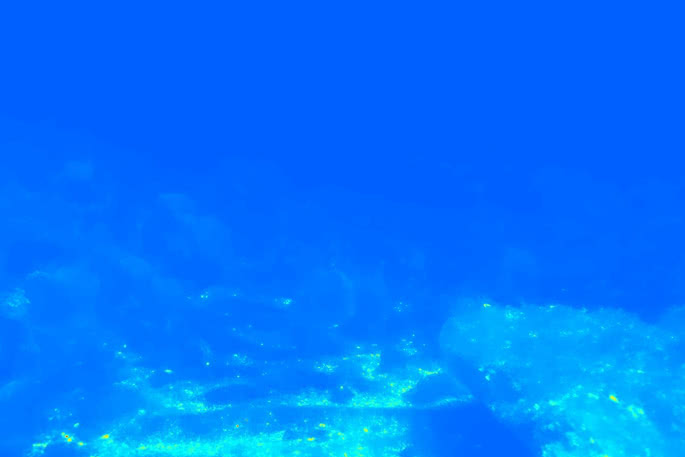} &
\includegraphics[width=0.32\linewidth]{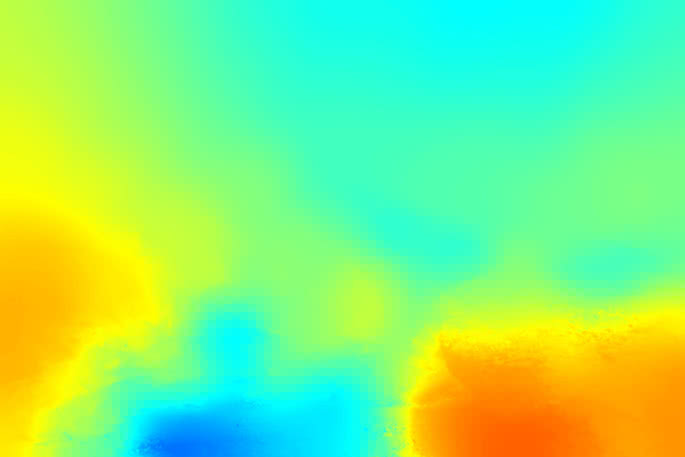} &
\includegraphics[width=0.32\linewidth]{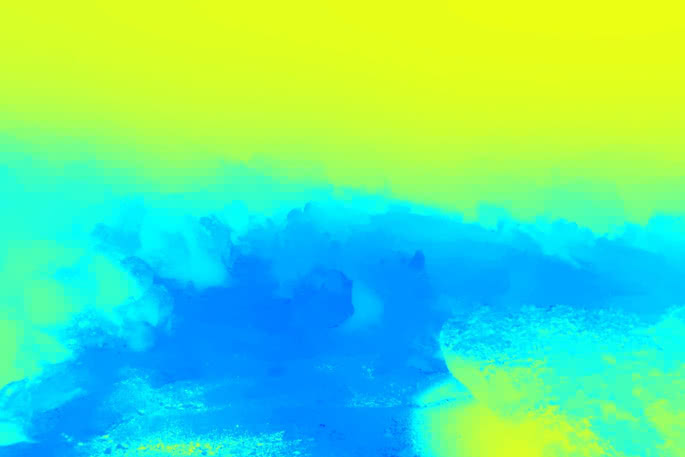} \\
$\rho = 0.64$ & $\rho = 0.17$ & $\rho = 0.05$ \\
\end{tabular} \vspace{-0.1cm}
\caption[Underwater image enhancement: effect of using a wrong water type]{Top two rows: An image taken in murky waters is enhanced both by automatically estimating the water type (center column) and by assuming a specific water type which is wrong (right column). Bottom two rows: the enhanced output of different methods for comparison.}
\label{fig:WrongWaterType}
\end{figure}

Finally, we wish to demonstrate the importance of correctly estimating the water type. Fig.~\ref{fig:WrongWaterType} shows on the leftmost column (top row) an image taken in murky waters at the Mediterranean Sea. The center column shows our output, which automatically estimated the water type to be C5, whereas the rightmost column shows the output assuming the image was taken in open ocean waters (type I). As evident, the color restoration with the wrong attenuation ratios leads to distorted colors, and the transmission estimation is much less accurate compared to the selected water type.

\section{Conclusions}

We expanded the haze-lines model to handle wavelength-dependent attenuation. Specifically, we recover underwater scenes by taking into account various water bodies that were classified by oceanographers according to their optical properties.  We showed that recovering transmission for each color channel separately adds just two global parameters to the problem. By considering a comprehensive physical model we were able to reconstruct scenes with a complex 3D structure and accurately correct the colors of objects that are farther away.

Unlike terrestrial images, the imaging model in Eq.~\ref {eq:BasicModelUW} is over-simplified~\cite{DeryaCVPR17}. This makes underwater scenes more difficult to simulate despite recent efforts in this direction~\cite{blasinski2017uiss}. As a result, the evaluation of underwater image enhancement methods is more difficult. This can also explain the relative few learning-based techniques in this field. We collected a new \insitu dataset of underwater images with ground truth and hope that our stereo dataset with color charts will advance the field of underwater image enhancement, since it has many real-world implications for oceanic research.

\ifCLASSOPTIONcompsoc
  \section*{Acknowledgments}
  TT was supported by the The Leona M. and Harry B. Helmsley Charitable Trust and The Maurice Hatter Foundation.
  DB was supported by The Mediterranean Sea Research Center of Israel and by Apple Graduate Fellowship.
  This research was supported in part by ISF grant 1917/15.
\else
  \section*{Acknowledgment}
\fi

\ifCLASSOPTIONcaptionsoff
  \newpage
\fi



\bibliographystyle{IEEEtran}
\bibliography{IEEEabrv,main}
\end{document}